\documentclass[lettersize,journal]{IEEEtran}
\usepackage{amsmath,amsfonts}
\usepackage{algorithmic}
\usepackage{algorithm}
\usepackage{array}
\usepackage[caption=false,font=normalsize,labelfont=sf,textfont=sf]{subfig}
\usepackage{textcomp}
\usepackage{stfloats}
\usepackage{url}
\usepackage{verbatim}
\usepackage{graphicx}
\usepackage{cite}
\usepackage{ulem}
\usepackage{url}

\usepackage{scalerel}
\usepackage{color}
\usepackage{tikz}
\usetikzlibrary{svg.path}
\definecolor{orcidlogocol}{HTML}{A6CE39}
\tikzset{
  orcidlogo/.pic={
    \fill[orcidlogocol] svg{M256,128c0,70.7-57.3,128-128,128C57.3,256,0,198.7,0,128C0,57.3,57.3,0,128,0C198.7,0,256,57.3,256,128z};
    \fill[white] svg{M86.3,186.2H70.9V79.1h15.4v48.4V186.2z}
                 svg{M108.9,79.1h41.6c39.6,0,57,28.3,57,53.6c0,27.5-21.5,53.6-56.8,53.6h-41.8V79.1z M124.3,172.4h24.5c34.9,0,42.9-26.5,42.9-39.7c0-21.5-13.7-39.7-43.7-39.7h-23.7V172.4z}
                 svg{M88.7,56.8c0,5.5-4.5,10.1-10.1,10.1c-5.6,0-10.1-4.6-10.1-10.1c0-5.6,4.5-10.1,10.1-10.1C84.2,46.7,88.7,51.3,88.7,56.8z};
  }
}
\newcommand\orcidicon[1]{\href{https://orcid.org/#1}{\mbox{\scalerel*{
\begin{tikzpicture}[yscale=-1,transform shape]
\pic{orcidlogo};
\end{tikzpicture}
}{|}}}}
\usepackage{hyperref}
\hypersetup{
    colorlinks=true,
    linkcolor=blue,
    filecolor=black,      
    urlcolor=magenta,
    citecolor=purple
}
\usepackage{tikz-network}
\usepackage{xcolor}        
\usepackage{multirow}
\usepackage{ulem}
\usepackage{amsmath}
\usepackage{colortbl}
\usepackage{makecell, booktabs}
\usepackage{amssymb}
\usepackage{lineno}
\usepackage[table]{xcolor}
\newcommand\eg{\textit{e.g.}}

\newcommand{\settablefont}{\fontsize{7.5}{10.5}\selectfont}

\usepackage{xcolor}  
\usepackage{multirow}
\usepackage{amsmath}
\usepackage{colortbl}
\usepackage{makecell}
\usepackage{amssymb}
\usepackage[table]{xcolor}
\usepackage{xspace}
\newcommand{\ourmethod}{{\fontfamily{lmtt}\selectfont \textbf{DSRS}}\xspace}
\newcommand{\ournet}{{\fontfamily{lmtt}\selectfont \textbf{DSRS-Net}}\xspace}
\definecolor{bestcolor}{HTML}{DCDCDC}
\definecolor{secondcolor}{HTML}{E8E8E8}
\definecolor{thirdcolor}{HTML}{F5F5F5}
\newcommand{\bestcell}[1]{\cellcolor{bestcolor}\textbf{#1}}
\newcommand{\secondcell}[1]{\cellcolor{secondcolor}#1}
\newcommand{\thirdcell}[1]{\cellcolor{thirdcolor}#1}
\usepackage{capt-of} 
\definecolor{pageblue}{RGB}{78,140,196}
\definecolor{darkgreen}{RGB}{34,139,34}
\usepackage{mathtools}
\usepackage{pifont}
\newcommand{\cmark}{\textcolor{darkgreen}{\ding{51}}}
\newcommand{\xmark}{ {\ding{55}}}

\title{
\raisebox{-0.25\height}{%
    \includegraphics[height=1.2cm]{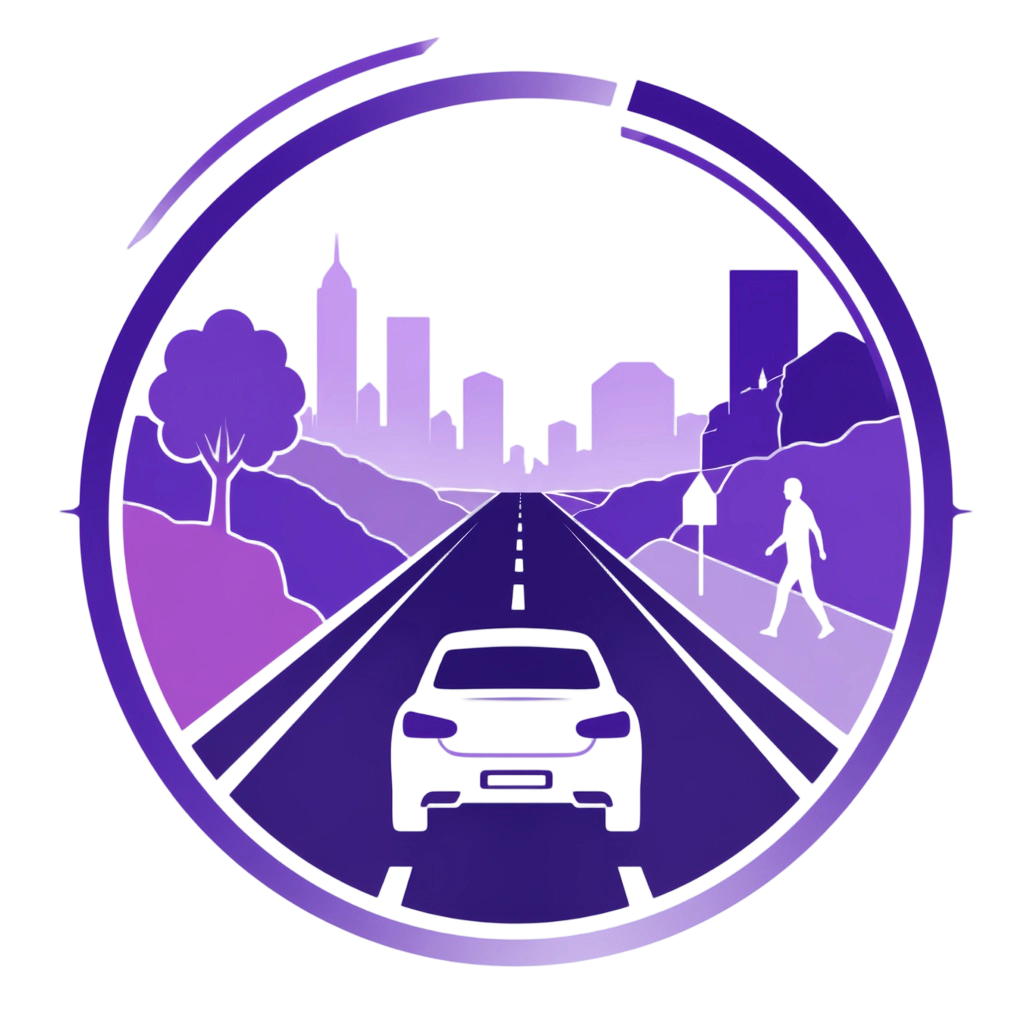}%
}\hspace{0.3em}%
\ourmethod: Teaching MLLMs to Perceive, Reason, and Segment \\in Complex Driving Scenes
}

\author{
Sicen Guo$^{\orcidicon{0009-0000-8079-8056}\,}$, 
Zhengfei Song$^{\orcidicon{0009-0005-6923-7254}\,}$, 
Kuanhsun Chen$^{\orcidicon{0009-0000-0888-7580}\,}$,
Xu Huang$^{\orcidicon{0009-0002-9993-7612}\,}$, \\ 
Ping Zhong$^{\orcidicon{0000-0003-3393-8874}\,}$,
Wei Ye$^{\orcidicon{0000-0002-3784-7788}\,}$,
Shaoyi~Du$^{\orcidicon{0000-0002-7092-0596}\,}$,
Rui Fan$^{\orcidicon{0000-0003-2593-6596}\,}$,~\IEEEmembership{{Senior} Member,~IEEE}
\thanks{
This research was supported by the National Natural Science Foundation of China under Grant 62473288, the National Key Laboratory of Human-Machine Hybrid Augmented Intelligence, Xi'an Jiaotong University (No. HMHAI-202406), the Fundamental Research Funds for the Central Universities, NIO University Programme (NIO UP), and the Xiaomi Young Talents Program. The research leading to these results has also received partial funding from the European Commission - European Union (under HORIZON EUROPE (HORIZON Research and Innovation Actions) under grant agreement 101093003 (TEMA) HORIZON-CL4-2022-DATA-01-01). Views and opinions expressed are, however, those of the authors only and do not necessarily reflect those of the European Union - European Commission. Neither the European Commission nor the European Union can be held responsible for them. (\emph{Corresponding author: Rui Fan})}
\thanks{Sicen Guo, Zhengfei Song, Kuanhsun Chen, and Xu Huang are with the College of Electronic \& Information Engineering, Tongji University, Shanghai 201804, China (e-mail: \{guosicen, 2151094, 2350487, hxnb\}@tongji.edu.cn).}
\thanks{Ping Zhong is with the Department of Computer Science and Technology, Central South University. (email: ping.zhong@csu.edu.cn).}
\thanks{Wei Ye is with the College of Electronics and Information Engineering, Tongji University, Shanghai 201804, China. He is also with the Shanghai Innovation Institute, Shanghai 200231, China (e-mail: yew@tongji.edu.cn).}
\thanks{Shaoyi Du is with the National Key Laboratory of Human-Machine Hybrid Augmented Intelligence, the National Engineering Research Center for Visual Information and Applications, and the Institute of Artificial Intelligence and Robotics, Xi'an Jiaotong University, Xi’an, Shaanxi 710049, China (e-mail: dushaoyi@xjtu.edu.cn).}
\thanks{Rui Fan is with the College of Electronic and Information Engineering, Shanghai Institute of Intelligent Science and Technology, Shanghai Research Institute for Intelligent Autonomous Systems, Shanghai Key Laboratory of Intelligent Autonomous Systems, State Key Laboratory of Autonomous Intelligent Unmanned Systems, and Frontiers Science Center for Intelligent Autonomous Systems (Ministry of Education), Tongji University, Shanghai 201804, China, as well as with the National Key Laboratory of Human-Machine Hybrid Augmented Intelligence, Xi'an Jiaotong University, Xi'An, Shaanxi 710049, China (e-mail: rui.fan@ieee.org).
}
}

\begin{document}


\maketitle
\normalem

\begin{figure}[t!]
\centering
\includegraphics[width=0.473\textwidth]{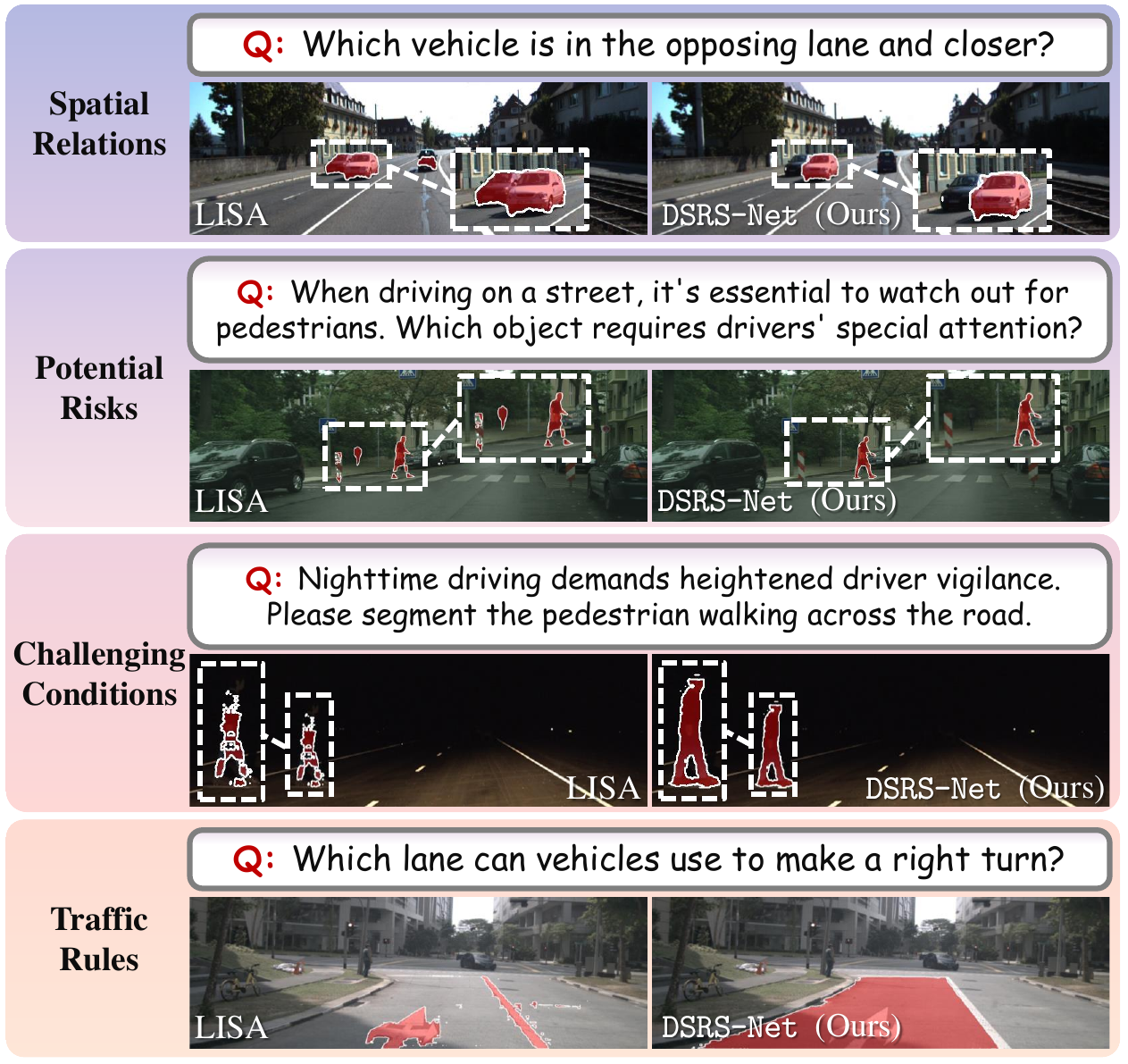}
\centering
\caption
{
Comparison of the segmentation masks by LISA \cite{lai2024lisa} and our \ournet, facing the challenges in the driving scene reasoning segmentation task.
}
\vspace{-0.5em}
\label{fig.page1}
\end{figure} 

\begin{abstract}
Existing reasoning segmentation frameworks typically employ multimodal large language models (MLLMs) with broad world knowledge to extract semantics in a human-like manner and generate corresponding segmentation masks. Nevertheless, these frameworks lack the necessary scene-specific awareness. This paper introduces driving scene reasoning segmentation (\ourmethod), a scene-specific task that requires models to interpret implicit instructions, infer spatial relations, detect potential risks, handle challenging conditions, and comprehend underlying traffic rules, ultimately generating accurate segmentation masks. Meanwhile, we establish a benchmark dataset to facilitate future research on the \ourmethod task. To pioneer this task, we propose \ournet, a novel architecture with advanced reasoning segmentation capability for automated driving. Specifically, \ournet seamlessly infuses geometric features for deeper reasoning segmentation, without compromising the original multimodal fusion mechanism. This strategy effectively mitigates the modality representation mismatch in MLLMs caused by RGB-only pretraining, while enhancing the integration of geometric, contextual, and linguistic features. Extensive experiments demonstrate the superior performance of \ournet in both driving scene reasoning segmentation and other language-driven segmentation tasks across various public datasets. 
The source code and dataset will be released upon publication.
\end{abstract}


\section{Introduction}
\label{sec:intro}

The ultimate ambition of automated driving is to develop vehicles whose scene-specific awareness and decision-making capabilities rival and ultimately surpass those of human drivers \cite{11193653}, enabling safe and efficient interaction with vehicles of varying automation levels and other road users in mixed traffic environments. Such coexistence poses profound challenges in terms of safety, efficiency, and interaction dynamics. Although current advances have made remarkable strides in fundamental perception tasks such as semantic segmentation, their scene understanding of complex mixed traffic remains superficial, lacking deeper contextual reasoning.
Today's vehicles can recognize basic elements, but they often struggle to reason about real-world mixed traffic scenarios involving shared road usage, implicit intentions, and interactive behaviors \cite{wang2025traffic,jia2023think}.
This underscores a fundamental gap between low-level perceptual recognition and the high-level scene reasoning required for automated vehicles \cite{jiang2025survey}.

\begin{figure*}[t!]
\centering
\includegraphics[width=0.985\textwidth]{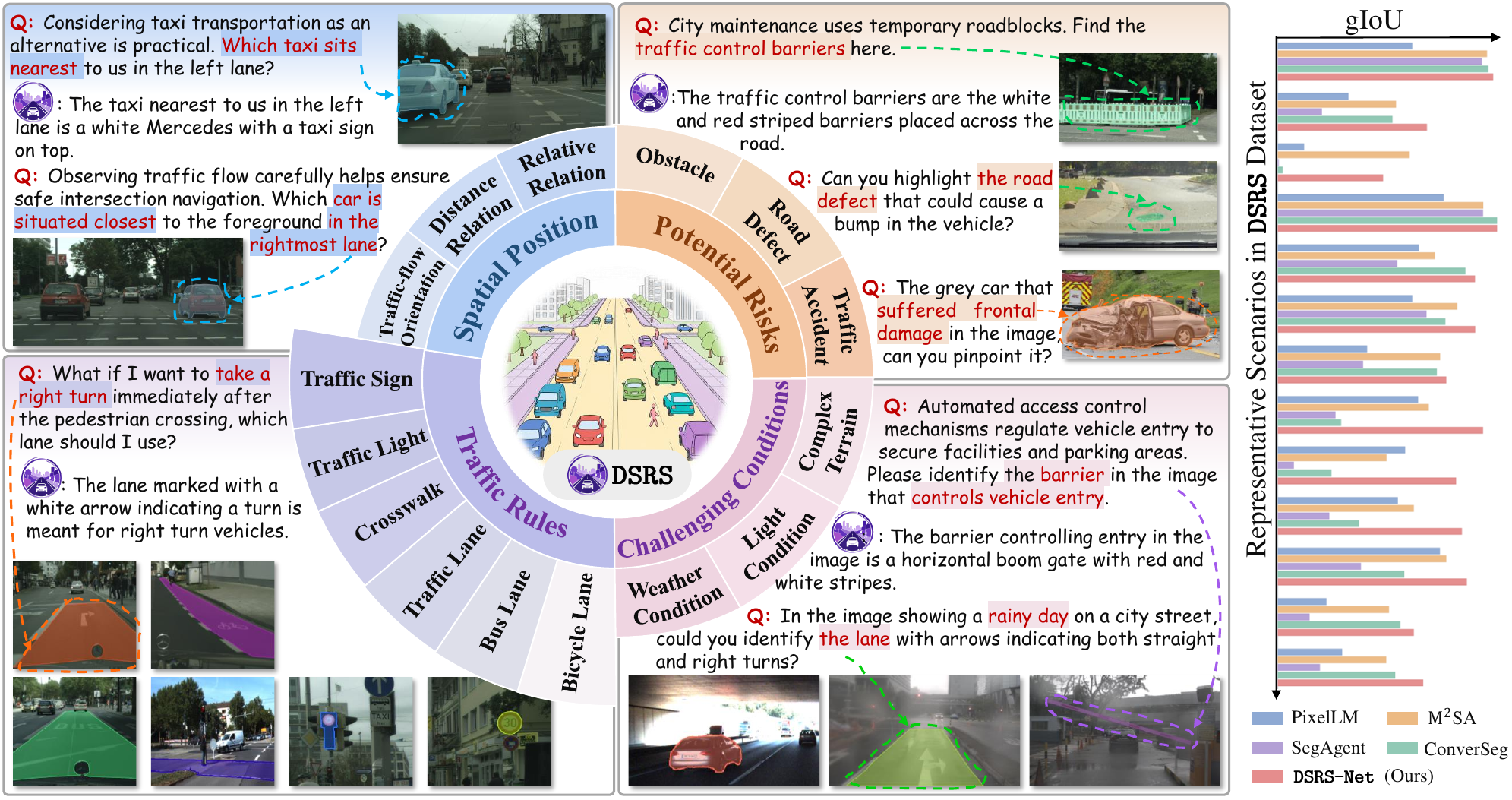}
\centering
\caption
{ Overview of the proposed \ourmethod benchmark and \ournet. The left part illustrates representative driving-scene reasoning segmentation scenarios. The right part reports quantitative comparisons on the \ourmethod dataset.
}
\vspace{-0.5em}
\label{fig.page2}
\end{figure*} 

To bridge this gap, we advocate for a paradigm shift away from conventional, purely perception tasks. Instead of merely supervising models to recognize what they see, we should empower them to reason in a human-like, instruction-driven manner \cite{lai2024lisa}. The development of multimodal large language models (MLLMs) offers a compelling glimpse into this future, which reframes visual understanding as an interactive, language-driven process \cite{xia2024gsva}. Nevertheless, despite the remarkable progress of these MLLMs \cite{yang2023improved} in general vision-language benchmarks, the field of reasoning segmentation in complex mixed traffic scenes remains largely underexplored. Traditional reasoning segmentation tasks have largely focused on broad representations of world knowledge and exhibit limited scene-specific awareness in driving scenes \cite{11048675}. As illustrated in Fig. \ref{fig.page1}, LISA \cite{lai2024lisa} lacks the capacity for driving scene reasoning and thus cannot generate correct segmentation outputs. 

As illustrated in Fig. \ref{fig.page2}, this limitation motivates us to introduce a scene-specific, multimodal perception task for automated vehicle systems: \textbf{Driving Scene Reasoning Segmentation} (\ourmethod). Since automated vehicles operate in mixed traffic environments involving dynamic interactions with multiple road users, the \ourmethod task integrates six core capabilities: (1) interpreting abstract, often implicit, human instructions, (2) comprehending traffic rules, \eg, functions of different lanes, traffic lights, and traffic signs, (3) inferring spatial relations, (4) detecting potential risks in complex driving scenes, including traffic accidents and both positive and negative obstacles, (5) handling challenging conditions, \eg, complex terrain, extreme light and weather conditions, and (6) generating the corresponding segmentation masks.

Endowing neural networks with scene-specific reasoning capabilities requires unleashing the power of MLLMs \cite{liu2023visual}. A commonly adopted architecture \cite{shen2025reasoning} maps RGB images and text instructions into a shared linguistic-contextual space \cite{liu2023visual}, followed by a segmentation head \cite{kirillov2023segment} to generate pixel-level predictions. However, this coarse-grained strategy of simply connecting modules often proves insufficient for the fine-grained reasoning required in complex mixed traffic scenes \cite{ranasinghe2024learning}. Accurately inferring spatial relations, such as occlusion, direction, and distance, cannot rely solely on RGB images \cite{cheng2024spatialrgpt}. Incorporating geometric clues, such as relative depth maps, offers a more reliable option. Moreover, in low-visibility or texture-less environments, and when object-level properties such as size and shape information are required, the incorporation of geometric clues can substantially enhance the model's reasoning capabilities \cite{fan2020sne-roadseg}.

While integrating geometric clues is a logical next step, existing approaches are superficial and predominantly limited to text generation tasks \cite{cai2024spatialbot, cheng2024spatialrgpt, yang2025spatialllm,feng2025spatialintelligence}. These approaches often simply concatenate depth maps and RGB images as inputs to MLLMs for superficial geometric understanding. However, their neglect of the inherent modality gap that large language models (LLMs) are typically pretrained with RGB visual inputs, prevents them from effectively utilizing such geometric features \cite{nadeem2025maskadapt, wang2025spatialclip}. 
Moreover, the mismatch between geometric and contextual representations may introduce noise into the original alignment between visual and linguistic features in the MLLM, ultimately compromising the reasoning performance.
This leads us to ask: can a model truly enhance its reasoning capabilities without explicitly aligning geometric features with contextual semantics?

To answer this question and pioneer the \ourmethod task, we propose the \textbf{driving scene reasoning segmentation network} (\ournet), a novel framework that seamlessly infuses geometric features into MLLM for deeper scene-specific reasoning. Its core is our proposed depth-guided hierarchical feature interaction (DHFI) strategy. 
By projecting geometric priors into the hierarchical vision features, the DHFI strategy enforces a contextually aligned integration of appearance-based perception and geometric structure.
For instance, \ournet learns to continuously correlate the concept of a ``car'' not merely through visual texture cues, but through its geometric structure and spatial positioning within the scene. This capability enables the model to holistically comprehend and infer scene semantics across color, texture, and geometric domains. In a nutshell, our contributions are threefold:

\begin{itemize}
    \item We introduce the \ourmethod task, a scene-specific extension of reasoning segmentation that explicitly evaluates traffic-rule understanding, spatial reasoning, hazard awareness, and perception under challenging driving conditions. To catalyze research in this direction, we release a benchmark dataset comprising over 1,500 image-instruction-mask samples tailored for automated vehicles.
    \item We propose the \ournet, which establishes a pioneering pathway for integrating geometric priors into an MLLM-based segmentation model. 
    \item We propose the DHFI strategy, a specialized mechanism that effectively projects geometric priors into the hierarchical vision feature space.
\end{itemize}


\section{Related Work}
\label{sec:related_work}

\subsection{Language-Driven Segmentation Tasks}
\label{sec:res}
Image segmentation has evolved from predefined pixel-wise prediction toward more intuitive and interactive paradigms \cite{liu2025refersam}. Referring expression segmentation \cite{hu2016segmentation} enables object localization through natural language but largely depends on explicit visual attributes, limiting its ability to interpret implicit human intent in complex real-world scenarios \cite{shen2025reasoning}. Reasoning segmentation \cite{lai2024lisa} addresses this limitation by requiring multi-step inference and external knowledge. Representative methods such as LISA \cite{lai2024lisa} decode segmentation masks from MLLM-generated embeddings, while GSVA \cite{xia2024gsva} further handles cases where the referred object is absent. Nevertheless, existing models \cite{jang2025mmr, lan2026text4seg++, liu2026unipixel} mainly rely on broad, general-purpose knowledge and remain limited in scenarios requiring fine-grained, scene-specific understanding. Therefore, we introduce the scene-specific \ourmethod task and benchmark, which require models to interpret implicit instructions in complex driving scenes, supporting safer and more intelligent automated driving.

\subsection{MLLM-Based Models for Automated Driving}
\label{sec:related_mllm}


Integrating language prompts into driving tasks provides a promising avenue for improving reasoning and generalization in automated driving systems \cite{11098671}. 
For instance, DiLU \cite{wen2023dilu} learns from prior driving experiences, while DriveMLM \cite{wang2023drivemlm} and LMDrive \cite{shao2024lmdrive} integrate multimodal inputs, such as images and point clouds, to generate control signals from language instructions. However, these methods mainly address high-level decisions and lack the pixel-level localization required for safety-critical perception. This gap is also evident in existing benchmarks. nuScenes \cite{caesar2020nuscenes} and BDD100K \cite{yu2020bdd100k} benchmarks lack reasoning-oriented question–answer annotations, whereas DriveLM-Data \cite{sima2024drivelm} and Impromptu VLA \cite{chi2025impromptu} benchmarks lack pixel-level labels for fine-grained scene understanding. To this end, we develop the \ourmethod benchmark.

\section{Methodology}
\label{sec:methodology}

\begin{figure*}[t!]
\centering
\includegraphics[width=0.99\textwidth]{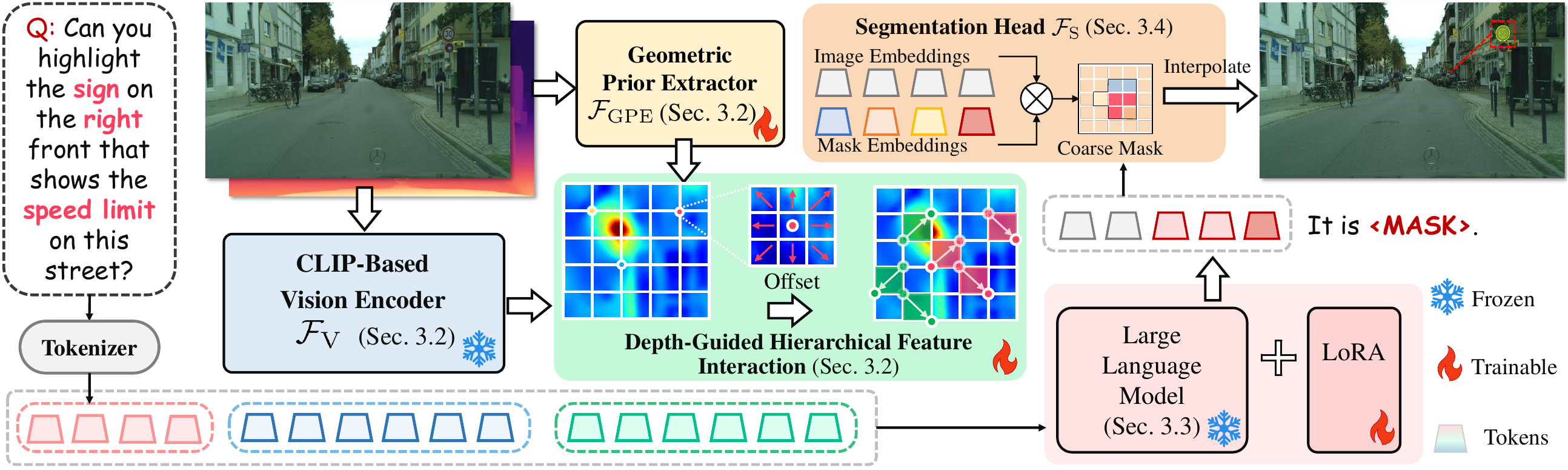}
\centering
\caption
{
An illustration of our proposed \ournet. Given the RGB image, depth map, and text query, \ournet generates text output and segmentation masks. We use the lightweight LoRA strategy \cite{hu2022lora} for efficient fine-tuning. 
}
\label{fig.framework}
\end{figure*} 

\subsection{Overall Workflow}
\label{sec:overview}
As illustrated in Fig. \ref{fig.framework}, in the CLIP-based vision encoder branch, the RGB image $\mathbf{x}_{\mathrm{img}} \in \mathbb{R}^{H\times W \times 3}$ is first divided into non-overlapping $14 \times 14$ patches. These patches are flattened and projected into $D$-dimensional tokens, which are then passed through the vision encoder $\mathcal{F}_{\mathrm{V}}$ for further contextualized representation learning. Simultaneously, the corresponding depth map $\mathbf{x}_{\mathrm{geo}} \in \mathbb{R}^{ H\times W}$ passes through the GPE module $\mathcal{F}_{\mathrm{GPE}}$ to generate the geometric prior $\boldsymbol{F}_0^S$ for subsequent feature interactions. Specifically, we evenly split the CLIP-based vision encoder into four blocks. At the $i$-th stage, the geometric prior $\boldsymbol{F}_i^S$ interacts with the vision feature $\boldsymbol{F}_i^V$ via the DHFI module, thereby improving the model's overall comprehension and contextual reasoning ability. Subsequently, two separate projectors are employed to map the vision and geometric features into their corresponding visual tokens $\boldsymbol{T}^V$ and $\boldsymbol{T}^S$. Then, the downstream LLM $\mathcal{F}_{\mathrm{LLM}}$ processes the fused tokens and generates the textual response $\boldsymbol{Y}_\text{txt}$ auto-regressively. The generation of binary segmentation masks is indicated by the presence of designated \texttt{<MASK>} tokens within $\boldsymbol{Y}_{\text{txt}}$.
The hidden representation associated with image tokens is extracted from the LLM output and yields the image embeddings ${\boldsymbol{H}}_{\text{img}}$. Finally, ${\boldsymbol{H}}_\text{img}$ is fed to the segmentation head $\mathcal{F}_{\mathrm{S}}$ to predict the ultimate binary segmentation mask.

\subsection{Multimodal Vision Encoder}
\label{sec:encoder}

\subsubsection{CLIP-Based Vision Encoder}
\label{sec:clip_encoder}
We leverage a CLIP-based vision encoder $\mathcal{F}_{\mathrm{V}}$ to extract contextual visual features. Given an input RGB image $\mathbf{x}_{\mathrm{img}}$, the encoder first extracts the hierarchical vision features $\boldsymbol{F}_i^V \in \mathbb{R}^{\frac{H W}{14^2} \times D}$. Subsequently, a projection layer maps the extracted features into the visual token $\boldsymbol{T}^V  \in \mathbb{R}^{\frac{H W}{14^2} \times C }$, aligning them with the input embedding space of the LLM.

\subsubsection{Geometric Prior Extractor}
\label{sec:spe}
Recent studies have demonstrated that embedding convolutional mechanisms into Transformer-based architectures can significantly improve the model's capacity for fine-grained scene inference and precise boundary modeling \cite{gu2023convformer, yuan2021incorporating}. Building on this idea, we propose the GPE module $\mathcal{F}_{\mathrm{GPE}}$, which utilizes ResNet blocks to extract structured geometric priors $\boldsymbol{F}_0^S \in \mathbb{R}^{\frac{HW}{14^2} \times D}$ from the relative depth map $\mathbf{x}_{\mathrm{geo}}$.

\subsubsection{Depth-Guided Hierarchical Feature Interaction}
\label{sec:fusion}

Most existing MLLMs are pretrained predominantly on RGB images, restricting their capacity to comprehend unfamiliar geometric modalities such as depth maps \cite{wang2025generative, han2025controlthinker}. Consequently, the straightforward concatenation of depth and RGB images \cite{cai2024spatialbot,cheng2024spatialrgpt} has shown limited effectiveness in enhancing reasoning capabilities, and may even disrupt the original visual-language alignment established within MLLMs. To address these limitations, we design the DHFI strategy to explicitly project geometric priors into the visual-semantic space, thereby enhancing reasoning capabilities in MLLMs. Such an integration allows the model to capitalize on geometric priors without learning unfamiliar input modalities from scratch, thus improving efficiency and generalization.

\begin{figure}[t!]
\centering
\includegraphics[width=0.485\textwidth]{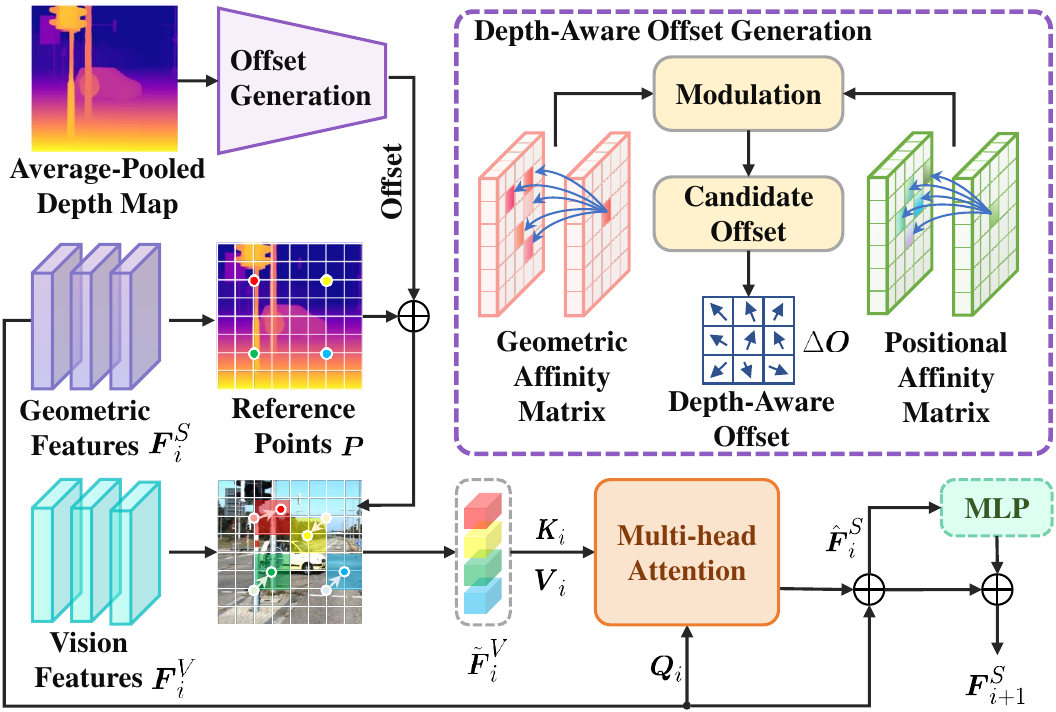}
\centering
\caption
{
An illustration of our proposed depth-aware deformable attention module.
}
\vspace{-0.5em}
\label{fig.framework1}
\end{figure}

The DHFI strategy comprises four stacked depth-aware deformable attention ($\text{DA}^2$) blocks, which preserve the hierarchical structure of the vision pipeline and progressively align semantic and geometric representations. As illustrated in Fig. \ref{fig.framework1}, in each $\text{DA}^2$ module, given the input geometric prior $\boldsymbol{F}_i^S$, a uniform grid of reference points $\boldsymbol{P} \in \mathbb{R}^{\frac{H}{14} \times \frac{W}{14} \times 2}$ is generated. These points are initialized as linearly 2D coordinates ranging from $(0,0)$ to $\left(\frac{H}{14}-1, \frac{W}{14}-1\right)$, forming a regular grid of size $\frac{H}{14} \times \frac{W}{14}$. These coordinates are subsequently normalized to the range $[0, 1]$ along both axes, where $(0, 0)$ corresponds to the top-left corner and $(1, 1)$ to the bottom-right corner of the grid.
Existing deformable attention frameworks \cite{xia2022vision} often rely on a multi-layer perceptron (MLP) to learn sampling offsets from scratch. 
While MLPs are theoretically capable of modeling complex functions, their unconstrained learning schemes often demand large-scale training data to achieve reliable generalization. To this end, we decompose the offset generation process into a two-stage process:

\begin{figure*}[t!]
\centering
\includegraphics[width=0.99\textwidth]{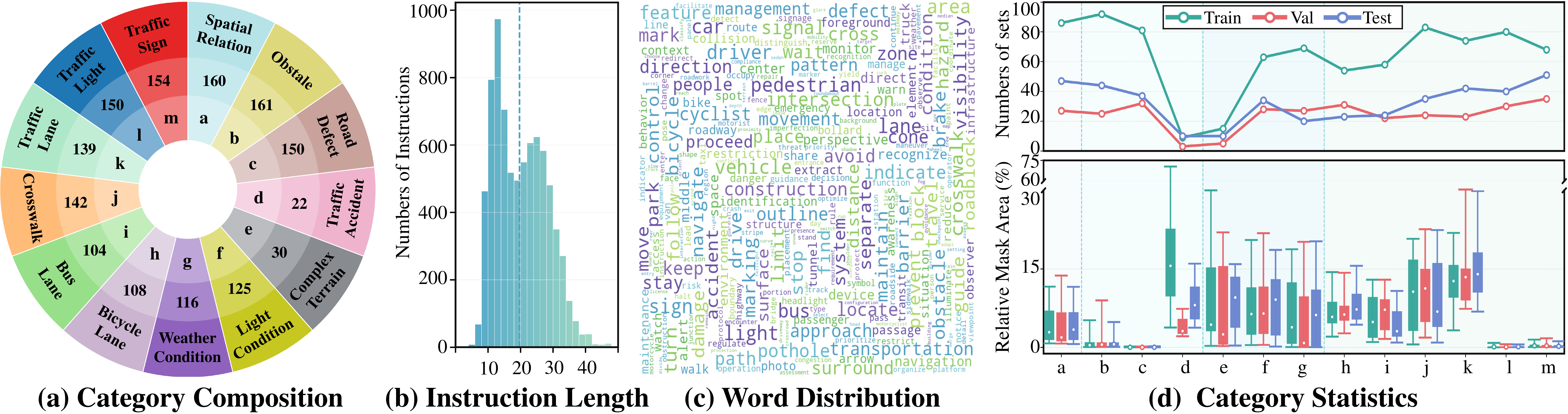}
\centering
\caption
{
A comprehensive analysis of our proposed \ourmethod dataset: (a) category composition; (b) distribution of instruction lengths, with the dashed line indicating the mean length; (c) frequency-weighted word cloud; and (d) category-wise sample counts (top) and relative mask-scale distributions, with category indices corresponding to those in (a).
}
\label{fig.DSRS_main}
\end{figure*} 

In the first stage, the depth-aware offset generation module leverages the average-pooled depth map along with positional priors to produce a set of modulated geometric-aware offset candidates. Specifically, we first calculate the geometric affinity matrix $\boldsymbol{D} \in \mathbb{R}^{\frac{H W}{14^2} \times \frac{H W}{14^2}}$ by calculating pairwise relative depth differences between each depth patch, expressed as follows:
\begin{equation}
\boldsymbol{D}_{i j, i^{\prime} j^{\prime}}=d_{i j}-d_{i^{\prime} j^{\prime}},
\end{equation}
where each element $\boldsymbol{D}_{i j, i^{\prime} j^{\prime}}$ represents the depth distance between the patches at locations $(i, j)$ and $\left(i^{\prime}, j^{\prime}\right)$.
Since the offsets depend on both geometric structure and positional layout, we further compute a positional affinity matrix ${\boldsymbol{L}} \in \mathbb{R}^{\frac{H W}{14^2} \times \frac{H W}{14^2} \times 2}$ among patch-wise 2D coordinates to enhance location-sensitive sampling, expressed as follows:
\begin{equation}
{\boldsymbol{l}}_{i j, i^{\prime} j^{\prime}}={\boldsymbol{p}}_{i j}-{\boldsymbol{p}}_{i^{\prime} j^{\prime}},
\end{equation}
where ${\boldsymbol{l}}_{i j, i^{\prime} j^{\prime}}\in \mathbb{R}^{2}$ represents the relative 2D positional distance vector in ${\boldsymbol{L}}$ and ${\boldsymbol{p}}_{i j}$ represents absolute 2D position at locations $(i, j)$. Accordingly, the modulated candidate offsets are generated by applying an MLP to the channel-wise concatenation of the affinity matrices $\boldsymbol{D}$ and ${\boldsymbol{L}}$. Such an explicitly encoding geometric relations process can capture explicit object-level dependencies, facilitating more transparent decision-making over purely data-driven offset prediction.

In the second stage, the $\text{DA}^2$ employs an MLP to select a subset of candidate offsets $\Delta{\boldsymbol{O}}$. This selection strategy filters out unstable offset candidates, reducing noise sensitivity and stabilizing downstream attention computation. 

Subsequently, the vision features $\boldsymbol{F}_i^V$ are sampled at the locations of the deformed points $\boldsymbol{P}+\Delta\boldsymbol{O}$, generating the deformed vision features $\tilde{\boldsymbol{F}}_i$.
To realize cross-modal alignment, $\text{DA}^2$ employs a multi-head cross-attention mechanism \cite{cheng2022masked} conditioned on learned offsets, followed by an MLP for nonlinear feature integration. This process can be formulated as follows:
\begin{equation}
\boldsymbol{Q}_i=\boldsymbol{F}_i^S W_q,
\boldsymbol{K}_i=\tilde{\boldsymbol{F}}_i^V W_k,
\boldsymbol{V}_i=\tilde{\boldsymbol{F}}_i^V W_v,
\end{equation}
\begin{equation}
\hat{\boldsymbol{F}}_{i}^S=\boldsymbol{F}_{i}^S+\operatorname{MHA}\left(\boldsymbol{Q}_i,\boldsymbol{K}_i,\boldsymbol{V}_i\right),
\end{equation}
\begin{equation}
\boldsymbol{F}_{i+1}^S=\hat{\boldsymbol{F}}_{i}^S+\operatorname{MLP}\left(\operatorname{LN}\left(\hat{\boldsymbol{F}}_{i}^S\right)\right), 
\end{equation}
where ``$\operatorname{MHA}(\cdot)$'' denotes the multi-head attention operation \cite{cheng2022masked}, $W_q, W_k$, and $W_v$ are learnable projection matrices for the query, key, and value, and ``$\operatorname{LN}(\cdot)$'' represents the layer normalization operation \cite{lei2016layer}. To ensure compatibility with downstream LLM, the geometric feature $\boldsymbol{F}_{4}^S$ is mapped into the semantic token space using a projection layer, resulting in the visual token $\boldsymbol{T}^S$.

\begin{table*}[t]
\centering
\setlength{\tabcolsep}{4.8mm}
\settablefont
\begin{tabular}{ll|cc|cc|cc|cc}
\toprule[1pt]

\multirow{2}{*}{Method} 
& \multirow{2}{*}{MLLM} 
& \multicolumn{4}{c|}{Zero-shot (zs)} 
& \multicolumn{4}{c}{Fine-tuned (ft)} \\

\cline{3-10}

& 
& \multicolumn{2}{c|}{Val} 
& \multicolumn{2}{c|}{Test}
& \multicolumn{2}{c|}{Val}
& \multicolumn{2}{c}{Test} \\

& 
& gIoU & cIoU
& gIoU & cIoU
& gIoU & cIoU
& gIoU & cIoU \\

\hline
\hline

ReLA \cite{liu2023gres}
& -
&13.1&9.1&17.1&12.3&43.6&50.8&44.4&53.3\\
X-Decoder \cite{zou2023generalized}
& -
&24.1&14.3&26.3&15.1&56.5&48.8&51.4&46.2\\
SEEM \cite{zou2023segment}
& -
&23.9&16.2&29.7&14.9&42.1&34.1&42.4&33.2\\
Grounded-SAM \cite{liu2024grounding}
& -
&6.6&26.2&10.0&28.6&20.4&32.8&17.0&31.8\\
LISA \cite{lai2024lisa}
& LLaVAv1 7B&22.7&20.3&30.5&28.3&55.7&52.5&56.8&55.1\\
GSVA \cite{xia2024gsva}
& LLaVAv1 7B&26.1&19.4&31.1&26.3&   {59.7}&60.8&59.9&61.0\\
PixelLM \cite{ren2024pixellm}
& LLaVAv1 7B&25.0&16.5&31.3&22.3&39.3&53.8&40.1&53.8\\
M$^2$SA \cite{jang2025mmr}
& LLaVAv1 7B&22.2&22.7&30.2&29.0&54.6&54.7&57.5&61.1\\
SegAgent \cite{zhu2025segagent}
& Qwen-VL 7B&23.6&11.4&31.2&16.9&24.7&12.3&30.3&17.7\\
Seg-R1 \cite{you2025seg}
& Qwen2.5-VL 7B
&   {43.4}&29.4&  {47.9}&35.1&50.8&39.8&51.8&40.8\\
ConverSeg-Net \cite{sahoo2026conversational}
& Qwen2.5-VL 3B
&18.6&23.4&28.4&34.0&43.2&45.9&47.9&47.9\\
\midrule
\rowcolor[HTML]{DCDCDC}
\textbf{\ournet}
& LLaVA1.5 7B&37.4&35.0&47.5&50.2&56.0&70.3
&61.1&77.8\\
\rowcolor[HTML]{DCDCDC}
\textbf{\ournet}
& InternVL2.5 8B&\textbf{45.8}&\textbf{40.5}&\textbf{54.9}&\textbf{53.1}&\textbf{67.2}&\textbf{76.7}&\textbf{71.2}&\textbf{79.7}\\
\bottomrule[1pt]
\end{tabular}
\caption{
Comparative results on driving scene segmentation. 
`zs' denotes zero-shot evaluation, and `ft' indicates fine-tuning on the \ourmethod dataset. In addition, the \ourmethod dataset has no overlap with the initial training dataset. 
}
\label{tb.dsrs}
\end{table*}

\subsection{Large Language Model}
\label{sec:llm}
Receiving both the visual representations $\boldsymbol{T}^V$ and $\boldsymbol{T}^S$, as well as the textual instruction token $\boldsymbol{T}^T$, the LLM $\mathcal{F}_{\mathrm{LLM}}$ jointly processes these inputs to generate the output token embeddings $\boldsymbol{Y}_{txt}$.
To obtain the textual output, the token embeddings $\boldsymbol{Y}_{\text{txt}}$ are projected into the vocabulary space via a learnable linear layer. A special token, \texttt{<MASK>}, is introduced into the vocabulary, activating the LLM to switch from text generation to a segmentation-aware mode.
Subsequently, the output embedding corresponding to the \texttt{<MASK>} token, denoted as $\boldsymbol{Y}_{\text{txt}}[\texttt{<MASK>}]$ and the image embeddings ${\boldsymbol{H}}_{\text{img}}$ are passed to the downstream segmentation head. 

\subsection{Segmentation Head}
\label{sec:sfm}
To recover fine spatial details from low-resolution visual features, the model autoregressively predicts multiple <MASK> tokens, each corresponding to a position in the image. The hidden embeddings of the \texttt{<MASK>} tokens are used as queries over the processed image embeddings ${\boldsymbol{H}}_{\text{img}}$, and a scaled dot-product similarity map is computed to identify pixels semantically aligned with the target region. Positive responses in this similarity map are then converted into the predicted binary mask. The resulting similarity maps are concatenated and reshaped into a higher-resolution mask, enabling dynamic, image-conditioned upsampling.

\section{Experiments}
\label{sec:experiments}
\begin{table*}[!t]
\centering
\begin{minipage}[t]{0.56\textwidth}
\vspace{0pt}
\centering

\settablefont
\setlength{\tabcolsep}{1.5mm}

{%
\begin{tabular}{l|ccc|ccc|cc}

\toprule[1pt]

\multirow{2}{*}{Method}
& \multicolumn{3}{c|}{RefCOCO}
& \multicolumn{3}{c|}{RefCOCO+}
& \multicolumn{2}{c}{RefCOCOg}\\

& Val.
& TestA
& TestB
& Val.
& TestA
& TestB
& Val.
& Test\\

\hline
\hline

CRIS \cite{wang2022cris}
&70.5&73.2&66.1
&62.3&68.1&53.7
&59.9&60.4\\

ReLA \cite{liu2023gres}
&73.8&76.5&70.2
&66.0&71.0&57.7
&65.0&66.0\\

LISA \cite{lai2024lisa}
&74.1&76.5&71.1
&62.4&67.4&56.5
&66.4&68.5\\

LISA (ft) \cite{lai2024lisa}
&74.9&79.1&72.3
&65.1&70.8&58.1
&67.9&70.6\\

GSVA \cite{xia2024gsva}
&76.4&77.4&72.8
&64.5&67.7&58.6
&71.1&72.0\\

GSVA (ft) \cite{xia2024gsva}
&77.2&78.9&73.5
&65.9&69.6&59.8
&72.7&73.3\\

PixelLM \cite{ren2024pixellm}
&73.0&76.5&68.2
&66.3&71.7&58.3
&69.3&70.5\\

M$^2$SA \cite{jang2025mmr}
&74.0&76.8&69.7
&63.1&67.2&56.1
&67.0&68.3\\

SegAgent (ft) \cite{zhu2025segagent}
& {79.7}& {81.4}& {76.6}
&   {72.5}&75.8&66.9
&  {75.1}&  {75.2}\\

Seg-R1 \cite{you2025seg}
&74.3&78.7&67.6
&62.6&70.9&57.9
&71.0&71.4\\

POPEN \cite{zhu2025popen}
&    {78.5} & 79.9 & 73.0 & 70.3 & 74.4 & 62.4 & 73.8 &    {74.6}\\

SAM 3 Agent \cite{carion2025sam}       
& 75.5 & 77.6 & 71.0 & 67.3 & 71.1 & 63.4 & 73.4 & 74.0\\

READ \cite{qian2025reasoning}
&78.1 &   {80.2} &73.2 &68.4 &73.7 &60.4 &70.1 &71.4\\

WISE \cite{zhou2026efficient}
& - & 79.1 & - & - & 74.0 & - & - & 71.8\\

ConverSeg-Net \cite{sahoo2026conversational}
&78.2&  {80.3}&74.7&72.0&   {77.5}&66.3&   {74.1}&73.9\\

\midrule
\rowcolor[HTML]{DCDCDC}
\ournet-LLaVA1.5
&76.3&79.1&74.8&70.2&76.5&69.3&73.2&72.6\\
\rowcolor[HTML]{DCDCDC}
\ournet-LLaVA1.5 (ft)
&   {78.5}&79.6&   {75.3}&  {72.8}&  {78.0}&  {70.7}&   {74.1}&73.4\\
\rowcolor[HTML]{DCDCDC}
\ournet-InternVL2.5
&77.0&79.5&73.8
&   {72.5}&77.1&67.7
&73.3&73.7\\
\rowcolor[HTML]{DCDCDC}
\ournet-InternVL2.5 (ft)
&  \textbf{{80.9}}&\textbf{82.1}&  \textbf{77.2}& {\textbf{74.3}}& {\textbf{78.2}}& {\textbf{71.2}}& {\textbf{75.4}}& {\textbf{76.2}}\\
\bottomrule[1pt]

\end{tabular}
}


\captionof{table}{
Referring expression segmentation results on the RefCOCO, RefCOCO+, and RefCOCOg datasets, where “ft” denotes fine-tuning on training samples from the referring segmentation datasets.
}
\label{tb.res}

\end{minipage}
\hfill
%
\begin{minipage}[t]{0.425\textwidth}
\vspace{0pt}
\centering

\settablefont
\setlength{\tabcolsep}{2.5mm}

\begin{tabular}{l|cc|cc}

\toprule[1pt]

\multirow{2}{*}{Method}
&\multicolumn{2}{c|}{Val.}
&\multicolumn{2}{c}{Test}\\

&gIoU&cIoU&gIoU&cIoU\\

\hline
\hline

OVSeg \cite{liang2023open}
&28.5&18.6&26.1&20.8\\

ReLA \cite{liu2023gres}
&22.4&19.9&21.3&22.0\\

X-Decoder \cite{zou2023generalized}
&22.6&17.9&21.7&16.3\\

SEEM \cite{zou2023segment}
&25.5&21.2&24.3&18.7\\

Grounded-SAM \cite{liu2024grounding}
&26.0&14.5&21.3&16.4\\

LISA-7B \cite{lai2024lisa} 
&44.4&46.0&36.8&34.1\\

LISA-7B (ft) \cite{lai2024lisa}
&52.9&54.0&47.3&48.4\\

PixelLM \cite{ren2024pixellm}
&44.4&32.1&41.1&38.1\\

M$^2$SA-7B \cite{jang2025mmr}
&30.5&35.2&24.7&28.0\\

Seg-R1 \cite{you2025seg}
&58.6&41.2&56.7&53.7\\
SegAgent \cite{zhu2025segagent}
&34.4&28.7&34.5&31.2\\
OISA-1B \cite{ying2025towards} &- &61.2 &- &50.8\\
ConverSeg-Net \cite{sahoo2026conversational}
&{56.4}&-&52.2&-\\
Text4Seg++ \cite{lan2026text4seg++}&59.1 &49.5 &57.1 &52.1\\
UniPixel \cite{liu2026unipixel} &60.5 &58.7 &- &- \\
\midrule
\rowcolor[HTML]{DCDCDC}
\ournet-LLaVA1.5&53.9 &57.7 &54.5 &56.8 \\
\rowcolor[HTML]{DCDCDC}
\ournet-LLaVA1.5(ft)
&54.1&60.5&55.0&58.1\\
\rowcolor[HTML]{DCDCDC}
\ournet-InternVL2.5
&57.0&57.2&60.5&63.3\\
\rowcolor[HTML]{DCDCDC}
\ournet-InternVL2.5(ft)
&\textbf{62.5}&\textbf{66.8}&\textbf{65.2}&\textbf{69.4}\\

\bottomrule[1pt]

\end{tabular}


\captionof{table}{
Reasoning segmentation results, where “ft” denotes fine-tuning on 239 samples from the ReasonSeg training set.} 

\label{tb.rs}

\end{minipage}

\end{table*}

\subsection{\ourmethod Dataset and Implementation Details}
\label{sec:dataset}

We establish the \ourmethod dataset, with each sample containing an RGB image, a corresponding relative depth map, implicit reasoning instructions, and pixel-level masks. The entire dataset is split into three sets for training (832 pairs), validation (312 pairs), and testing (417 pairs). The images are selected from several real-world driving datasets \cite{cordts2016cityscapes, menze2015kitti, caesar2020nuscenes,wang2019apolloscape}.  Fig. \ref{fig.DSRS_main} presents a detailed analysis of the dataset statistics. As shown in Fig. \ref{fig.DSRS_main}(a), with 1,561 annotated samples, \ourmethod is larger than the widely used ReasonSeg \cite{lai2024lisa} and provides broad coverage of instruction complexities, mask scales, and driving-specific reasoning categories: spatial relations, potential risks, challenging conditions, and traffic rules. Fig. \ref{fig.DSRS_main}(b) shows that instructions range in length and complexity from straightforward logical reasoning (\eg, ``areas where parking is allowed'') to more complex multi-hop reasoning involving spatial relations (\eg, ``Considering taxi transportation as an alternative is practical. Which taxi sits nearest to us in the left lane?''). The frequency-weighted word cloud in Fig. \ref{fig.DSRS_main}(c) highlights a diverse and domain-specific vocabulary involved in driving scenes. As shown in Fig. \ref{fig.DSRS_main}(d), the substantial variation in mask scales, ranging from small traffic signs to large lanes, demonstrates that the proposed dataset requires models to perform robust reasoning across diverse semantic categories and object scales.

Our \ournet was trained for 120,000 iterations on an A800 GPU with a batch size of two, utilizing the DeepSpeed engine to enable efficient memory management. For optimization, we employed the AdamW optimizer with a learning rate of 0.0002. We quantify the segmentation performance of the model using the gIoU and cIoU metrics. 

\begin{figure}[t!]
\centering
\includegraphics[width=0.475\textwidth]{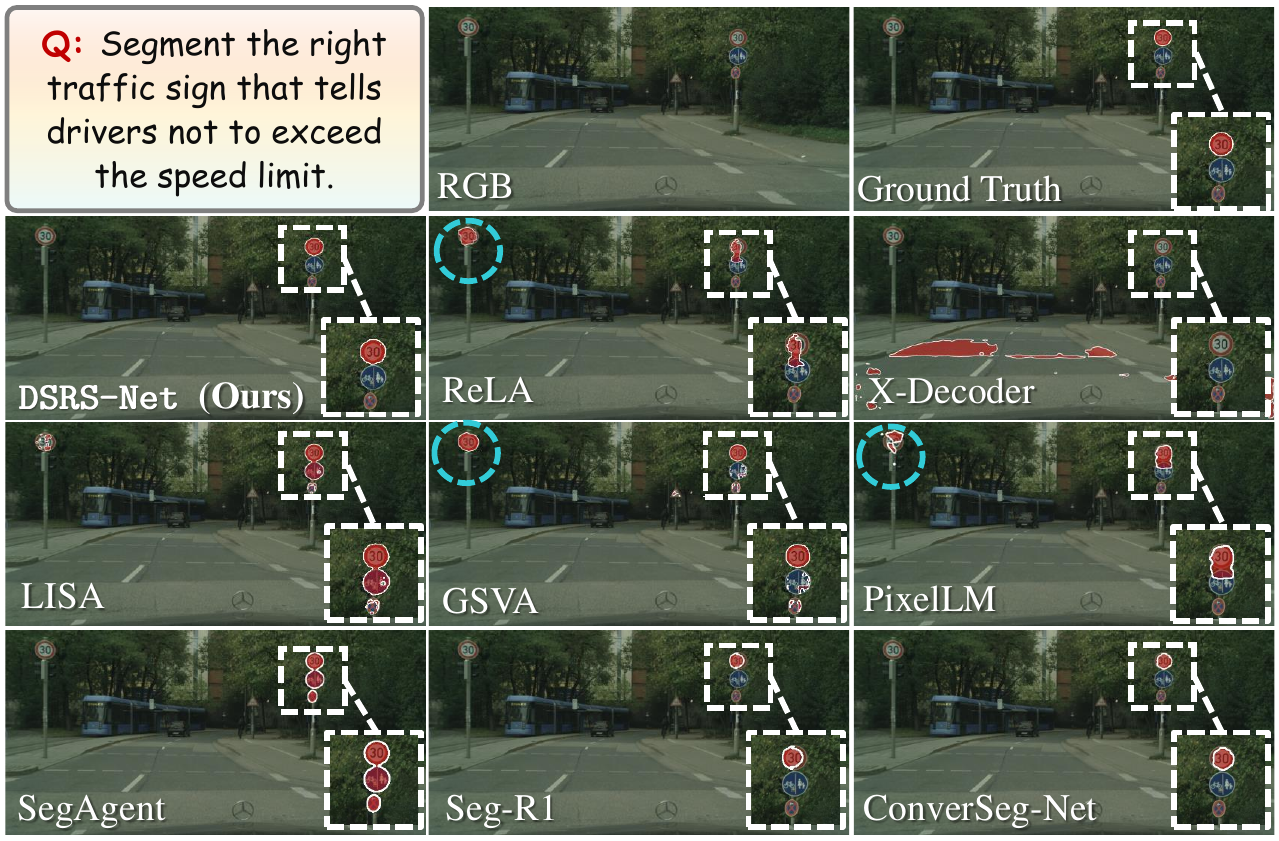}
\centering
\caption
{
Qualitative comparisons on the \ourmethod dataset, where ground-truth areas are highlighted with white dashed boxes and wrong areas are highlighted with blue dashed circles.
}
\vspace{-0.2em}
\label{fig.main_vis}
\end{figure}

\subsection{Driving Scene Segmentation Results}
\label{sec.exp_trans}
We compare our proposed \ournet with representative language-driven segmentation networks on the \ourmethod dataset, as presented in Table \ref{tb.dsrs}. To evaluate the generalizability of \ournet, we first conduct a zero-shot test using the pre-trained weights obtained from the mixed-source training data. Our proposed \ournet outperforms existing SoTA methods in zero-shot driving scene reasoning segmentation, demonstrating its exceptional generalizability. This is primarily due to the richer geometric prior knowledge embedded in our \ournet, so that familiar geometric features can also be captured effectively when encountering unseen scenes.
Moreover, it can be observed that finetuned \ournet consistently outperforms other language-driven segmentation networks in the \ourmethod task, demonstrating its superior reasoning capabilities in automated driving. As illustrated in Fig. \ref{fig.main_vis}, it is evident that our proposed \ournet significantly outperforms existing approaches in driving scene reasoning segmentation. In scenarios involving multiple visually similar candidates that necessitate deep reasoning and scene-specific knowledge, our method demonstrates superior inference capabilities and robustness.


\subsection{Language-Driven Segmentation Results}
\label{sec.exp_reason}

In this section, we present additional experiments to demonstrate the versatility of \ournet in handling other language-driven segmentation tasks. As presented in Table \ref{tb.res}, we first evaluate the methods on the validation and test sets of RefCOCO, RefCOCO+ \cite{kazemzadeh2014referitgame}, and RefCOCOg \cite{mao2016generation}. \ournet achieves superior performance across these referring expression segmentation benchmarks. As shown in Table \ref{tb.rs}, we further evaluate the effectiveness of \ournet on the ReasonSeg \cite{lai2024lisa} dataset. Specifically, \ournet-InternVL2.5 (ft) outperforms previous SoTA approaches, achieving an improvement of over 2.0\% in gIoU and 5.6\% in cIoU. These results highlight the robustness and generalizability of the \ournet framework in handling language-driven segmentation tasks with varying levels of instruction complexity.

\subsection{Ablation Studies}
\label{sec.abla1}

\begin{table}[!t]
\begin{center}
\setlength{\tabcolsep}{3.8mm}
{
\settablefont
{
\begin{tabular}{cccc|cc}
\toprule[1pt]
\makecell[l]{RGB \\Branch} 
& \makecell[l]{Depth\\CLIP \\Branch} 
& \makecell[l]{GPE}
& \makecell[l]{DHFI} 
& gIoU 
& cIoU  \\
\hline
\hline
$\checkmark$&  &  &  &53.00 &50.42\\
$\checkmark$&$\checkmark$&  &  
&  {55.25} 
&   {52.12}\\
$\checkmark$&  &$\checkmark$&  
&37.39 
&36.67\\
$\checkmark$&  &  &$\checkmark$
&   {53.28} 
&  {55.10}\\
\rowcolor[HTML]{DCDCDC}
$\checkmark$&  &$\checkmark$&$\checkmark$
& {55.40} 
& {56.97}\\
\bottomrule[1pt]
\end{tabular}
}
\caption{Ablation studies on the core designs of \ournet. } 

\label{tb.abla1}
}
\vspace{-0.7em}
\end{center}
\end{table}

\begin{table}[!t]
\begin{center}
\setlength{\tabcolsep}{2.8mm}
{
\settablefont
{
\begin{tabular}{cc|l|cc}
\toprule[1pt]
\makecell[l]{Contextual \\Features} 
& \makecell[l]{Geometric \\Features} 
& Fusion Strategies 
& gIoU 
& cIoU  \\
\hline
\hline
 &$\checkmark$&Self-Attention 
&44.56
&44.08  \\
$\checkmark$&$\checkmark$&\makecell[l]{Element-wise \\Summation} 
&50.47
&51.12\\
$\checkmark$&$\checkmark$&\makecell[l]{Channel-wise \\Concatenation} 
&   {51.09} 
&   {51.86} \\
$\checkmark$&$\checkmark$&Cross-Attention 
&  {52.78}
&  {54.05}\\
\rowcolor[HTML]{DCDCDC}
$\checkmark$&$\checkmark$&${DA}^2$ 
& {55.40}
& {56.97}\\
\bottomrule[1pt]
\end{tabular}
}
\caption{Ablation studies on the DHFI strategy. }
\label{tb.abla2}
\vspace{-0.7em}
}
\vspace{-0.5em}
\end{center}
\end{table}

In this section, we conduct comprehensive ablation experiments by training \ournet for 80k iterations to investigate the individual contributions of key components within \ournet. 
As presented in Table \ref{tb.abla1}, removing any of them leads to a performance deterioration. Directly feeding the depth map into the vision encoder brings only a limited improvement, suggesting that RGB-pretrained MLLMs cannot effectively leverage unfamiliar geometric modalities without explicit alignment. In contrast, incorporating both the GPE and DHFI modules substantially improves driving scene reasoning segmentation. The degradation observed when only the GPE module is used further shows the representational incompatibility between geometric and contextual features, which may introduce conflicts within the LLM. These results demonstrate that the performance gain primarily stems from the effective alignment and interaction of geometric and contextual features, rather than merely from the inclusion of an additional depth input.

Moreover, Table \ref{tb.abla2} shows that simple fusion strategies such as element-wise summation and channel-wise concatenation provide limited gains, whereas attention-based interaction is more effective. The proposed ${DA}^2$ module achieves superior performance, demonstrating that depth-aware offsets help the model focus on contextually and geometrically relevant regions during feature interaction.

\section{Additional Discussion and Experiments}
Further discussions, experiments, and video are provided in the supplementary material. We first distinguish \ourmethod from other language-driven segmentation tasks, and further compare the proposed benchmark with existing spatial-reasoning and autonomous-driving datasets. Additional dataset analyses describe its category definitions, instruction distributions, and the dataset construction pipeline. We also provide detailed descriptions of the model architecture, loss, training datasets, evaluation metrics, depth map generation pipeline, and competitor configurations. Moreover, we discuss the challenges of spatial reasoning in MLLMs, the extensibility of our contributions to other multimodal tasks, current limitations, and failure cases of \ournet. Finally, additional quantitative and qualitative experiments, including fair comparisons using the same LLM backbone, robustness experiments with perturbed depth maps, resource consumption and efficiency analyses, category-wise evaluations on the \ourmethod dataset, and further ablation studies, consistently demonstrate that effective extraction and contextual alignment of geometric priors are essential for reliable reasoning segmentation in complex driving scenes.

\section{Conclusion and Future Work}
\label{sec:conclusion}
This paper introduced driving scene reasoning segmentation, a novel multimodal perception task, and released a mixed-traffic benchmark dataset specifically designed for complex driving scenarios. Additionally, we proposed \ournet, a novel framework that effectively infuses geometric features for deeper scene reasoning. Extensive experiments and comprehensive analyses validate the effectiveness of our novel contributions and the superior performance of \ournet. We hope our work offers new insights into the integration of MLLMs with vision tasks, paving the way for future advancements in automated driving.

{
    \small
    \bibliographystyle{IEEEtran}
    \bibliography{main}

@inproceedings{xia2024gsva,
  title={{GSVA}: Generalized segmentation via multimodal large language models},
  author={Xia, Zhuofan and others},
  booktitle={Proceedings of the IEEE/CVF Conference on Computer Vision and Pattern Recognition (CVPR)},
  pages={3858--3869},
  year={2024}
}

@article{shen2025reasoning,
  title={Reasoning segmentation for images and videos: a survey},
  author={Shen, Yiqing and others},
  journal={Computing Research Repository {(CoRR)}},
  volume={abs/2505.18816},
  url={https://arxiv.org/abs/2505.18816}, 
  year={2025}
}

@inproceedings{lai2024lisa,
  title={{LISA}: reasoning segmentation via large language model},
  author={Lai, Xin and others},
  booktitle={Proceedings of the IEEE/CVF Conference on Computer Vision and Pattern Recognition (CVPR)},
  pages={9579--9589},
  year={2024}
}

@article{kabir2025logic,
  title={{Logic-RAG}: augmenting large multimodal models with visual-spatial knowledge for road scene understanding},
  author={Kabir, Imran and others},
  journal={Computing Research Repository {(CoRR)}},
  volume={abs/2503.12663},
  url={https://arxiv.org/abs/2503.12663}, 
  year={2025}
}

@article{yang2023improved,
  title={An improved baseline for reasoning segmentation with large language model},
  author={Yang, Senqiao and others},
  journal={CoRR},
  year={2023}
}

@inproceedings{wang2025traffic,
  title={{Traffic scenario logic}: a spatial-temporal logic for modeling and reasoning of urban traffic scenarios},
  author={Wang, Ruolin and others},
  booktitle={Proceedings of the AAAI Conference on Artificial Intelligence (AAAI)},
  volume={39},
  number={14},
  pages={15195--15202},
  year={2025}
}

@inproceedings{jia2023think,
  title={Think twice before driving: towards scalable decoders for End-to-End autonomous driving},
  author={Jia, Xiaosong and others},
  booktitle={Proceedings of the IEEE/CVF Conference on Computer Vision and Pattern Recognition (CVPR)},
  pages={21983--21994},
  year={2023}
}

@inproceedings{cheng2024spatialrgpt,
 author = {{Cheng, An-Chieh} and others},
 booktitle = {Advances in Neural Information Processing Systems (NeurIPS)},
 pages = {135062--135093},
 publisher = {Curran Associates, Inc.},
 title = {{SpatialRGPT}: grounded spatial reasoning in vision-language models},
 volume = {37},
 year = {2024}
}

@inproceedings{zhu2025segagent,
  title={{SegAgent}: exploring pixel understanding capabilities in {MLLMs} by imitating human annotator trajectories},
  author={Zhu, Muzhi and others},
  booktitle={Proceedings of the IEEE/CVF Conference on Computer Vision and Pattern Recognition (CVPR)},
  pages={3686--3696},
  year={2025}
}

@inproceedings{liu2023visual,
 author = {Liu, Haotian and others},
 booktitle = {Advances in Neural Information Processing Systems (NeurIPS)},
 pages = {34892--34916},
 publisher = {Curran Associates, Inc.},
 title = {Visual instruction tuning},
 url = {https://proceedings.neurips.cc/paper_files/paper/2023/file/6dcf277ea32ce3288914faf369fe6de0-Paper-Conference.pdf},
 volume = {36},
 year = {2023}
}

@inproceedings{kirillov2023segment,
  title={{Segment anything}},
  author={Kirillov, Alexander and others},
  booktitle={Proceedings of the IEEE/CVF international conference on computer vision (ICCV)},
  pages={4015--4026},
  year={2023}
}

@inproceedings{ranasinghe2024learning,
  title={Learning to localize objects improves spatial reasoning in {Visual-LLMs}},
  author={Ranasinghe, Kanchana and others},
  booktitle={Proceedings of the IEEE/CVF Conference on Computer Vision and Pattern Recognition (CVPR)},
  pages={12977--12987},
  year={2024}
}

@article{cai2024spatialbot,
  title={{SpatialBot}: precise spatial understanding with vision language models},
  author={Cai, Wenxiao and others},
  journal={Computing Research Repository {(CoRR)}},
  volume={abs/2406.13642},
  url={https://arxiv.org/abs/2406.13642},
  year={2024}
}

@inproceedings{nadeem2025maskadapt,
  title={{MaskAdapt}: unsupervised geometry-aware domain adaptation using multimodal contextual learning and {RGB-Depth} masking},
  author={Nadeem, Numair and others},
  booktitle={Proceedings of the IEEE/CVF Conference on Computer Vision and Pattern Recognition (CVPR)},
  pages={5422--5432},
  year={2025}
}

@article{yang2025spatialllm,
  title={How to enable {LLM} with {3D} capacity? A survey of spatial reasoning in {LLMs}},
  author={Zha, Jiahui and others},
  journal={Computing Research Repository {(CoRR)}},
  volume={abs/2504.05786},
  url={https://arxiv.org/abs/2504.05786},
  year={2025}
}

@inproceedings{wang2025spatialclip,
  title={{SpatialCLIP}: learning {3D}-aware image representations from spatially discriminative language},
  author={Wang, Zhipeng and others},
  booktitle={Proceedings of the IEEE/CVF Conference on Computer Vision and Pattern Recognition (CVPR)},
  year={2025}
}

@article{feng2025spatialintelligence,
  title={A survey of large language model-powered spatial intelligence across scales: advances in embodied agents, smart cities, and earth science},
  author={Feng, Jiaqi and others},
  journal={Computing Research Repository {(CoRR)}},
  volume={abs/2504.09848},
  url={https://arxiv.org/abs/2504.09848},
  year={2025}
}

@inproceedings{yuan2021incorporating,
  title={Incorporating convolution designs into visual {Transformers}},
  author={Yuan, Kun and others},
  booktitle={Proceedings of the IEEE/CVF International Conference on Computer Vision (ICCV)},
  pages={579--588},
  year={2021}
}

@inproceedings{cheng2022masked,
  title={Masked-attention mask {Transformer} for universal image segmentation},
  author={Cheng, Bowen and others},
  booktitle={Proceedings of the IEEE/CVF Conference on Computer Vision and Pattern Recognition (CVPR)},
  pages={1290--1299},
  year={2022}
}

@inproceedings{gu2023convformer,
  title={{ConvFormer}: combining {CNN} and {Transformer} for medical image segmentation},
  author={Gu, Pengfei and others},
  booktitle={2023 IEEE 20th International Symposium on Biomedical Imaging (ISBI)},
  pages={1--5},
  year={2023},
  organization={IEEE}
}

@inproceedings{hong20233d,
 author = {Hong, Yining and others},
 booktitle = {Advances in Neural Information Processing Systems (NeurIPS)},
 pages = {20482--20494},
 publisher = {Curran Associates, Inc.},
 title = {{3D-LLM}: injecting the {3D} world into large language models},
 url = {https://proceedings.neurips.cc/paper_files/paper/2023/file/413885e70482b95dcbeeddc1daf39177-Paper-Conference.pdf},
 volume = {36},
 year = {2023}
}

@article{wang2025generative,
  title={Generative {AI} for autonomous driving: frontiers and opportunities},
  author={Wang, Yuping and others},
  journal={Computing Research Repository {(CoRR)}},
  volume={abs/2505.08854},
  url={https://arxiv.org/abs/2505.08854},
  year={2025}
}

@article{han2025controlthinker,
  title={{ControlThinker}: unveiling latent semantics for controllable image generation through visual reasoning},
  author={Han, Feng and others},
  journal={Computing Research Repository {(CoRR)}},
  volume={abs/2506.03596},
  url={https://arxiv.org/abs/2506.03596},
  year={2025}
}

@inproceedings{xia2022vision,
  title={Vision {Transformer} with deformable attention},
  author={Xia, Zhuofan and others},
  booktitle={Proceedings of the IEEE/CVF Conference on Computer Vision and Pattern Recognition (CVPR)},
  pages={4794--4803},
  year={2022}
}

@article{lei2016layer,
  title={Layer normalization},
  author={Jimmy Lei Ba and others},
  journal={Computing Research Repository {(CoRR)}},
  volume={abs/1607.06450},
  url={https://arxiv.org/abs/1607.06450},
  year={2016}
}

@inproceedings{caesar2020nuscenes,
  title={{nuScenes}: a multimodal dataset for autonomous driving},
  author={Caesar, Holger and others},
  booktitle={Proceedings of the IEEE/CVF Conference on Computer Vision and Pattern Recognition (CVPR)},
  pages={11621--11631},
  year={2020}
}

@inproceedings{yu2020bdd100k,
  title={{BDD100k}: a diverse driving dataset for heterogeneous multitask learning},
  author={Yu, Fisher and others},
  booktitle={Proceedings of the IEEE/CVF Conference on Computer Vision and Pattern Recognition (CVPR)},
  pages={2636--2645},
  year={2020}
}

@inproceedings{cordts2016cityscapes,
  title={The {CityScapes} dataset for semantic urban scene understanding},
  author={Cordts, Marius and others},
  booktitle={Proceedings of the IEEE/CVF Conference on Computer Vision and Pattern Recognition (CVPR)},
  pages={3213--3223},
  year={2016}
}

@inproceedings{menze2015kitti,
  title={Object scene flow for autonomous vehicles},
  author={Menze, Moritz and others},
  booktitle={Proceedings of the IEEE/CVF Conference on Computer Vision and Pattern Recognition (CVPR)},
  pages={3061--3070},
  year={2015}
}

@article{wang2019apolloscape,
  title={The {ApolloScape} open dataset for autonomous driving and its application},
  author={Wang, Peng and others},
  journal={IEEE Transactions on Pattern Analysis and Machine Intelligence},
  volume={1},
  year={2019}
}

@article{jiang2025survey,
  title={A survey on vision-language-action models for autonomous driving},
  author={Jiang, Sicong and others},
  journal={Computing Research Repository {(CoRR)}},
  volume={abs/2506.24044},
  url={https://arxiv.org/abs/2506.24044},
  year={2025}
}

@article{liu2025refersam,
  title={{ReferSAM}: unleashing segment anything model for referring image segmentation},
  author={Liu, Sun-Ao and others},
  journal={IEEE Transactions on Circuits and Systems for Video Technology},
  year={2025},
  publisher={IEEE}
}

@inproceedings{hu2016segmentation,
  title={Segmentation from natural language expressions},
  author={Hu, Ronghang and others},
  booktitle={European Conference on Computer Vision (ECCV)},
  pages={108--124},
  year={2016},
  organization={Springer}
}

@article{wen2023dilu,
  title={{DiLu}: a knowledge-driven approach to autonomous driving with large language models},
  author={Wen, Licheng and others},
  journal={Computing Research Repository {(CoRR)}},
  volume={abs/2309.16292},
  url={https://arxiv.org/abs/2309.16292},
  year={2023}
}

@article{wang2023drivemlm,
  title={{DriveMLM}: aligning multi-modal large language models with behavioral planning states for autonomous driving},
  author={Wang, Wenhai and others},
  journal={Computing Research Repository {(CoRR)}},
  volume={abs/2312.09245},
  url={https://arxiv.org/abs/2312.09245},
  year={2023}
}

@inproceedings{shao2024lmdrive,
  title={{LMDrive}: closed-Loop end-to-end driving with large language models},
  author={Shao, Hao and others},
  booktitle={Proceedings of the IEEE/CVF Conference on Computer Vision and Pattern Recognition (CVPR)},
  pages={15120--15130},
  year={2024}
}

@inproceedings{sima2024drivelm,
  title={{DriveLM}: driving with graph visual question answering},
  author={Sima, Chonghao and others},
  booktitle={European Conference on Computer Vision (ECCV)},
  pages={256--274},
  year={2024},
  organization={Springer}
}

@article{chi2025impromptu,
  title={{Impromptu VLA}: open weights and open data for driving vision-language-action models},
  author={Chi, Haohan and others},
  journal={Computing Research Repository {(CoRR)}},
  volume={abs/2505.23757},
  url={https://arxiv.org/abs/2505.23757},
  year={2025}
}

@inproceedings{ding2023pla,
  title={{PLA}: language-driven open-vocabulary {3D} scene understanding},
  author={Ding, Runyu and others},
  booktitle={Proceedings of the IEEE/CVF Conference on Computer Vision and Pattern Recognition (CVPR)},
  pages={7010--7019},
  year={2023}
}

@inproceedings{zhou2017scene,
  title={Scene parsing through {ADE20K} dataset},
  author={Zhou, Bolei and others},
  booktitle={Proceedings of the IEEE/CVF Conference on Computer Vision and Pattern Recognition (CVPR)},
  pages={633--641},
  year={2017}
}

@inproceedings{mao2016generation,
  title={Generation and comprehension of unambiguous object descriptions},
  author={Mao, Junhua and others},
  booktitle={Proceedings of the IEEE/CVF Conference on Computer Vision and Pattern Recognition (CVPR)},
  pages={11--20},
  year={2016}
}

@inproceedings{kazemzadeh2014referitgame,
  title={{ReferItGame}: referring to objects in photographs of natural scenes},
  author={Kazemzadeh, Sahar and others},
  booktitle={Proceedings of the 2014 conference on empirical methods in natural language processing (EMNLP)},
  pages={787--798},
  year={2014}
}

@inproceedings{liu2023gres,
  title={{GRES}: generalized referring expression segmentation},
  author={Liu, Chang and others},
  booktitle={Proceedings of the IEEE/CVF Conference on Computer Vision and Pattern Recognition (CVPR)},
  pages={23592--23601},
  year={2023}
}

@inproceedings{hu2022lora,
  title={Lo{RA}: low-rank adaptation of large language models},
  author={Edward J Hu and others},
  booktitle={International Conference on Learning Representations (ICLR)},
  year={2022},
  url={https://openreview.net/forum?id=nZeVKeeFYf9}
}

@inproceedings{liang2023open,
  title={Open-vocabulary semantic segmentation with mask-adapted {CLIP}},
  author={Liang, Feng and others},
  booktitle={Proceedings of the IEEE/CVF Conference on Computer Vision and Pattern Recognition (CVPR)},
  pages={7061--7070},
  year={2023}
}

@inproceedings{liu2024grounding,
  title={Grounding {DINO}: marrying {DINO} with grounded pre-training for open-set object detection},
  author={Liu, Shilong and others},
  booktitle={European Conference on Computer Vision (ECCV)},
  pages={38--55},
  year={2024},
  organization={Springer}
}

@inproceedings{zou2023generalized,
  title={Generalized decoding for pixel, image, and language},
  author={Zou, Xueyan and others},
  booktitle={Proceedings of the IEEE/CVF Conference on Computer Vision and Pattern Recognition (CVPR)},
  pages={15116--15127},
  year={2023}
}

@inproceedings{zou2023segment,
  title={Segment everything everywhere all at once},
  author={{Zou, Xueyan} and others},
  booktitle={Advances in Neural Information Processing Systems (NeurIPS)},
  volume={36},
  pages={19769--19782},
  year={2023}
}

@inproceedings{wang2022cris,
  title={{CRIS}: CLIP-driven referring image segmentation},
  author={Wang, Zhaoqing and others},
  booktitle={Proceedings of the IEEE/CVF Conference on Computer Vision and Pattern Recognition (CVPR)},
  pages={11686--11695},
  year={2022}
}

@InProceedings{fan2020sne-roadseg,
author = {Fan, Rui and others},
title = {{SNE-RoadSeg}: incorporating surface normal information into semantic segmentation for accurate freespace detection},
booktitle = {Proceedings of the European Conference on Computer Vision (ECCV)},
year = {2020},
pages={340--356},
organization = {Springer},
}

@inproceedings{ma2022sqa3d,
  title={{SQA3D}: situated question answering in {3D} scenes},
  author={Ma, Xiaojian and others},
  booktitle={International Conference on Learning Representations (ICLR)},
  year={2023},
}

@inproceedings{linghu2024multi,
 author = {Linghu, Xiongkun and others},
 booktitle = {Advances in Neural Information Processing Systems (NeurIPS)},
 pages = {140903--140936},
 title = {Multi-modal situated reasoning in {3D} scenes},
 year = {2024}
}

@article{chen2024grounded3dllm,
      title={Grounded {3D-LLM}  with referent tokens}, 
      author={Chen, Yilun and others},
      journal={Computing Research Repository {(CoRR)}},
      volume={abs/2405.10370},
      url={https://arxiv.org/abs/2405.10370},
      year={2024},
}

@article{abdelreheem2022scanents,
    author = {Abdelreheem, Ahmed and others},
    title = {{ScanEnts3D}: exploiting phrase-to-{3D} object correspondences for improved visio-linguistic models in {3D} scenes},
    journal = {Computing Research Repository {(CoRR)}},
    volume = {abs/2212.06250},
    url = {https://arxiv.org/abs/2212.06250},
    year = {2022}
}

@inproceedings{embodiedscan,
    title={{EmbodiedScan}: a holistic multi-modal {3D} perception suite towards embodied {AI}},
    author={Wang, Tai and others},
    booktitle={Proceedings of the IEEE/CVF Conference on Computer Vision and Pattern Recognition (CVPR)},
    year={2024},
}

@article{huang2025surprise3d,
  title={{SURPRISE3D}: a dataset for spatial understanding and reasoning in complex {3D} scenes},
  author={Huang, Jiaxin and others},
  journal={Computing Research Repository {(CoRR)}},
  volume={abs/2507.07781},
  url={https://arxiv.org/abs/2507.07781},
  year={2025}
}

@article{zhang2024vla3ddataset3dsemantic,
  title={{VLA-3D}: a dataset for {3D} semantic scene understanding and navigation},
  author={Zhang, Haochen and others},
  journal={Computing Research Repository {(CoRR)}},
  volume={abs/2411.03540},
  url={https://arxiv.org/abs/2411.03540},
  year={2024}
}

@article{zhu2024scanreasonempowering3dvisual,
  title={{ScanReason}: empowering {3D} visual grounding with reasoning capabilities},
  author={Zhu, Chenming and others},
  journal={Computing Research Repository {(CoRR)}},
  volume={abs/2407.01525},
  url={https://arxiv.org/abs/2407.01525},
  year={2024}
}

@article{zhao2025extending,
  title={Extending large vision-language model for diverse interactive tasks in autonomous driving},
  author={Zhao, Zongchuang and others},
  journal={Computing Research Repository {(CoRR)}},
  volume={abs/2505.08725},
  url={https://arxiv.org/abs/2505.08725},
  year={2025}
}

@article{hao2025driveactionbenchmarkexploringhumanlike,
  title={{DriveAction}: a benchmark for exploring human-like driving decisions in {VLA} models},
  author={Hao, Yuhan and others},
  journal={Computing Research Repository {(CoRR)}},
  volume={abs/2506.05667},
  url={https://arxiv.org/abs/2506.05667},
  year={2025}
}

@misc{nie2024reason2driveinterpretablechainbasedreasoning,
  title={{Reason2Drive}: towards interpretable and chain-based reasoning for autonomous driving},
  author={Nie, Ming and others},
  journal={Computing Research Repository {(CoRR)}},
  volume={abs/2312.03661},
  url={https://arxiv.org/abs/2312.03661},
  year={2024}
}

@InProceedings{Plummer_2015_ICCV,
author = {Plummer, Bryan A. and others},
title = {{Flickr30k Entities}: collecting region-to-phrase correspondences for richer image-to-sentence models},
booktitle = {Proceedings of the IEEE International Conference on Computer Vision (ICCV)},
year = {2015},
pages={1576--1584}
}

@article{10.1162/tacl_a_00166,
    author = {Young, Peter and others},
    title = {From image descriptions to visual denotations: new similarity metrics for semantic inference over event descriptions},
    journal = {Transactions of the Association for Computational Linguistics},
    volume = {2},
    pages = {67-78},
    year = {2014},
    month = {02},
}

@article{krishna2017visual,
  title={{Visual Genome}: connecting language and vision using crowdsourced dense image annotations},
  author={Krishna, Ranjay and others},
  journal={International Journal of Computer Vision},
  volume={123},
  number={1},
  pages={32--73},
  year={2017},
}

@InProceedings{Lin_2016_CVPR,
author = {Lin, Tsung-Yu and others},
title = {Visualizing and understanding deep texture representations},
booktitle = {Proceedings of the IEEE Conference on Computer Vision and Pattern Recognition (CVPR)},
year = {2016},
}

@InProceedings{Kurita_2023_ICCV,
    author    = {Kurita, Shuhei and others},
    title     = {{RefEgo}: referring expression comprehension dataset from first-person perception of {Ego4D}},
    booktitle = {Proceedings of the IEEE/CVF International Conference on Computer Vision (ICCV)},
    year      = {2023},
    pages     = {15214-15224}
}

@inproceedings{SegPoint,
  title={{SegPoint}: segment any point cloud via large language model},
  author={He, Shuting and others},
  booktitle={Proceedings of the European Conference on Computer Vision (ECCV)},
  year={2024},
  pages={349--367},
}

@inproceedings{kim2018textual,
  title={Textual explanations for self-driving vehicles},
  author={Kim, Jinkyu and others},
  booktitle={Proceedings of the European Conference on Computer Vision (ECCV)},
  year={2018},
  pages = {577--593},
}

@inproceedings{reason3d,
title={{Reason3D}: searching and reasoning {3D} segmentation via large language model},
author={Huang, Kuan-Chih and others},
booktitle={International Conference on 3D Vision 2025 (3DV)},
year={2025},
url={https://openreview.net/forum?id=w1M87lE7D2}
}

@inproceedings{jia2024bench,
 author = {Jia, Xiaosong and others},
 booktitle = {Advances in Neural Information Processing Systems (NeurIPS)},
 pages = {819--844},
 title = {{Bench2Drive}: towards multi-ability benchmarking of closed-loop end-to-end autonomous driving},
 volume = {37},
 year = {2024}
}

@article{yang2024depth,
  title={Depth anything v2},
  author={Yang, Lihe and others},
  journal={Advances in Neural Information Processing Systems (NeurIPS)},
  volume={37},
  pages={21875--21911},
  year={2024}
}

@article{wu2024towards,
  title={Towards open vocabulary learning: A survey},
  author={Wu, Jianzong and others},
  journal={IEEE Transactions on Pattern Analysis and Machine Intelligence},
  volume={46},
  number={7},
  pages={5092--5113},
  year={2024},
  publisher={IEEE}
}

@ARTICLE{11098671,
  author={Feng, Yunjian and Li, Jun},
  journal={IEEE Transactions on Intelligent Transportation Systems}, 
  title={Auto-Prompting {SAM} for Container Detection and Localization in Container Yards}, 
  year={2025},
  volume={26},
  number={8},
  pages={11504-11514},
}

@ARTICLE{11048675,
  author={Yang, Shuo and others},
  journal={IEEE Transactions on Intelligent Transportation Systems}, 
  title={A Safe and Efficient Self-Evolving Algorithm for Decision-Making and Control of Autonomous Driving Systems}, 
  year={2025},
  volume={26},
  number={8},
  pages={11466-11478},
}

@ARTICLE{11193653,
  author={Pan, Weihang and others},
  journal={IEEE Transactions on Intelligent Transportation Systems}, 
  title={Cooperative Driving at Multiple Unsignalized Intersections in Fully Autonomous Driving Scenarios}, 
  year={2025},
  volume={26},
  number={12},
  pages={23298-23313},
  }

@inproceedings{caesar2018coco,
  title={Coco-stuff: Thing and stuff classes in context},
  author={Caesar, Holger and others},
  booktitle={Proceedings of the IEEE conference on computer vision and pattern recognition},
  pages={1209--1218},
  year={2018}
}

@inproceedings{jang2025mmr,
  title={{MMR}: A large-scale benchmark dataset for multi-target and multi-granularity reasoning segmentation},
  author={Jang, Donggon and others},
  booktitle={International Conference on Learning Representations (ICLR)},
  pages={67406--67431},
  year={2025}
}

@inproceedings{ren2024pixellm,
  title={{PixelLM}: Pixel reasoning with large multimodal model},
  author={Ren, Zhongwei and others},
  booktitle={Proceedings of the IEEE/CVF Conference on Computer Vision and Pattern Recognition (CVPR)},
  pages={26374--26383},
  year={2024}
}

@article{you2025seg,
  title={{Seg-R1}: Segmentation can be surprisingly simple with reinforcement learning},
  author={You, Zuyao and others},
  journal={Computing Research Repository (CoRR)},
  volume={abs/2506.22624},
  url={https://arxiv.org/abs/2506.22624},
  year={2025}
}

@inproceedings{sahoo2026conversational,
  title={Conversational image segmentation: Grounding abstract concepts with scalable supervision},
  author={Sahoo, Aadarsh and others},
  booktitle={Proceedings of the IEEE/CVF Conference on Computer Vision and Pattern Recognition (CVPR)},
  pages={39476--39485},
  year={2026}
}

@article{liu2026unipixel,
  title={{UniPixel}: Unified object referring and segmentation for pixel-level visual reasoning},
  author={Liu, Ye and others},
  journal={Advances in Neural Information Processing Systems (NeurIPS)},
  volume={38},
  pages={126078--126108},
  year={2026}
}

@article{yang2023lisa++,
  title={{LISA++}: An improved baseline for reasoning segmentation with large language model},
  author={Yang, Senqiao and others},
  journal={Computing Research Repository (CoRR)},
  volume={abs/2312.17240},
  url={https://arxiv.org/abs/2312.17240},
  year={2023}
}

@inproceedings{lu2025rsvp,
  title={{RSVP}: Reasoning segmentation via visual prompting and multi-modal chain-of-thought},
  author={Lu, Yi and others},
  booktitle={Proceedings of the Annual Meeting of the Association for Computational Linguistics (ACL)},
  pages={14699--14716},
  year={2025}
}

@inproceedings{zhu2025popen,
  title={{POPEN}: Preference-based optimization and ensemble for {LVLM}-based reasoning segmentation},
  author={Zhu, Lanyun and others},
  booktitle={Proceedings of the IEEE/CVF Conference on Computer Vision and Pattern Recognition (CVPR)},
  pages={30231--30240},
  year={2025}
}

@inproceedings{qian2025reasoning,
  title={Reasoning to attend: Try to understand how {<SEG>} token works},
  author={Qian, Rui and others},
  booktitle={Proceedings of the IEEE/CVF Conference on Computer Vision and Pattern Recognition (CVPR)},
  pages={24722--24731},
  year={2025}
}

@article{zhou2026efficient,
  title={Efficient reasoning via thought compression for language segmentation},
  author={Zhou, Qing and others},
  journal={Computing Research Repository (CoRR)},
  volume={abs/2604.02040},
  url={https://arxiv.org/abs/2604.02040},
  year={2026}
}

@article{carion2025sam,
  title={{SAM} 3: Segment anything with concepts},
  author={Carion, Nicolas and others},
  journal={Computing Research Repository (CoRR)},
  volume={abs/2511.16719},
  url={https://arxiv.org/abs/2511.16719},
  year={2025}
}

@article{lan2026text4seg++,
  title={{Text4seg++}: Advancing image segmentation via generative language modeling},
  author={Lan, Mengcheng and others},
  journal={IEEE Transactions on Pattern Analysis and Machine Intelligence},
  year={2026},
  publisher={IEEE}
}

@inproceedings{ying2025towards,
  title={Towards omnimodal expressions and reasoning in referring audio-visual segmentation},
  author={Ying, Kaining and others},
  booktitle={Proceedings of the IEEE/CVF International Conference on Computer Vision (ICCV)},
  pages={22575--22585},
  year={2025}
}

@InProceedings{Neuhold_2017_ICCV,
author = {Neuhold, Gerhard and others},
title = {The Mapillary Vistas Dataset for Semantic Understanding of Street Scenes},
booktitle = {Proceedings of the IEEE International Conference on Computer Vision (ICCV)},
month = {Oct},
year = {2017}
}

@article{li2024llava,
  title={Llava-onevision: Easy visual task transfer},
  author={Li, Bo and others},
  journal={arXiv preprint arXiv:2408.03326},
  year={2024}
}

@article{yun2026star,
  title={{StAR}: Segment Anything Reasoner},
  author={Yun, Seokju and others},
  journal={Computing Research Repository (CoRR)},
  volume={abs/2603.14382},
  url={https://arxiv.org/abs/2603.14382},
  year={2026}
}

@article{bai2024one,
  title={{One Token to Seg Them All: Language Instructed Reasoning Segmentation in Videos}},
  author={Bai, Zechen and others},
  journal={Computing Research Repository (CoRR)},
  volume={abs/2409.19603},
  url={https://arxiv.org/abs/2409.19603},
  year={2024}
}

@article{liang2026seg,
  title={{Seg-ReSearch}: Segmentation with Interleaved Reasoning and External Search},
  author={Liang, Tianming and others},
  journal={Computing Research Repository (CoRR)},
  volume={abs/2602.04454},
  url={https://arxiv.org/abs/2602.04454},
  year={2026}
}
}

\newpage
\newcommand{\appendixsectionentry}[3]{%
  \par\noindent\hangindent=1.8em\hangafter=1%
  \makebox[1.8em][l]{\textbf{#1}}\textbf{#2}%
  \nobreak\dotfill\nobreak\makebox[1.6em][r]{\textcolor{pageblue}{\pageref{#3}}}\par}
\newcommand{\appendixsubsectionentry}[3]{%
  \par\noindent\hangindent=4.6em\hangafter=1%
  \hspace*{1.8em}\makebox[2.8em][l]{#1}#2%
  \nobreak\dotfill\nobreak\makebox[1.6em][r]{\textcolor{pageblue}{\pageref{#3}}}\par}

\makeatletter
\renewcommand\section{\@startsection{section}{1}{\z@}%
  {-2.0ex plus -0.5ex minus -0.2ex}%
  {3pt plus 2pt minus 1pt}%
  {\large\bfseries\raggedright}}
\makeatother

\makeatletter
\gdef\copyright@on{}
\makeatother

\expandafter\let\csname ver@hyperref.sty\endcsname\savedhyperrefversion
\makeatother

\twocolumn[{
\renewcommand\twocolumn[1][]{#1}%
\maketitle
\begin{center}
\centering
\includegraphics[width=0.99\textwidth]{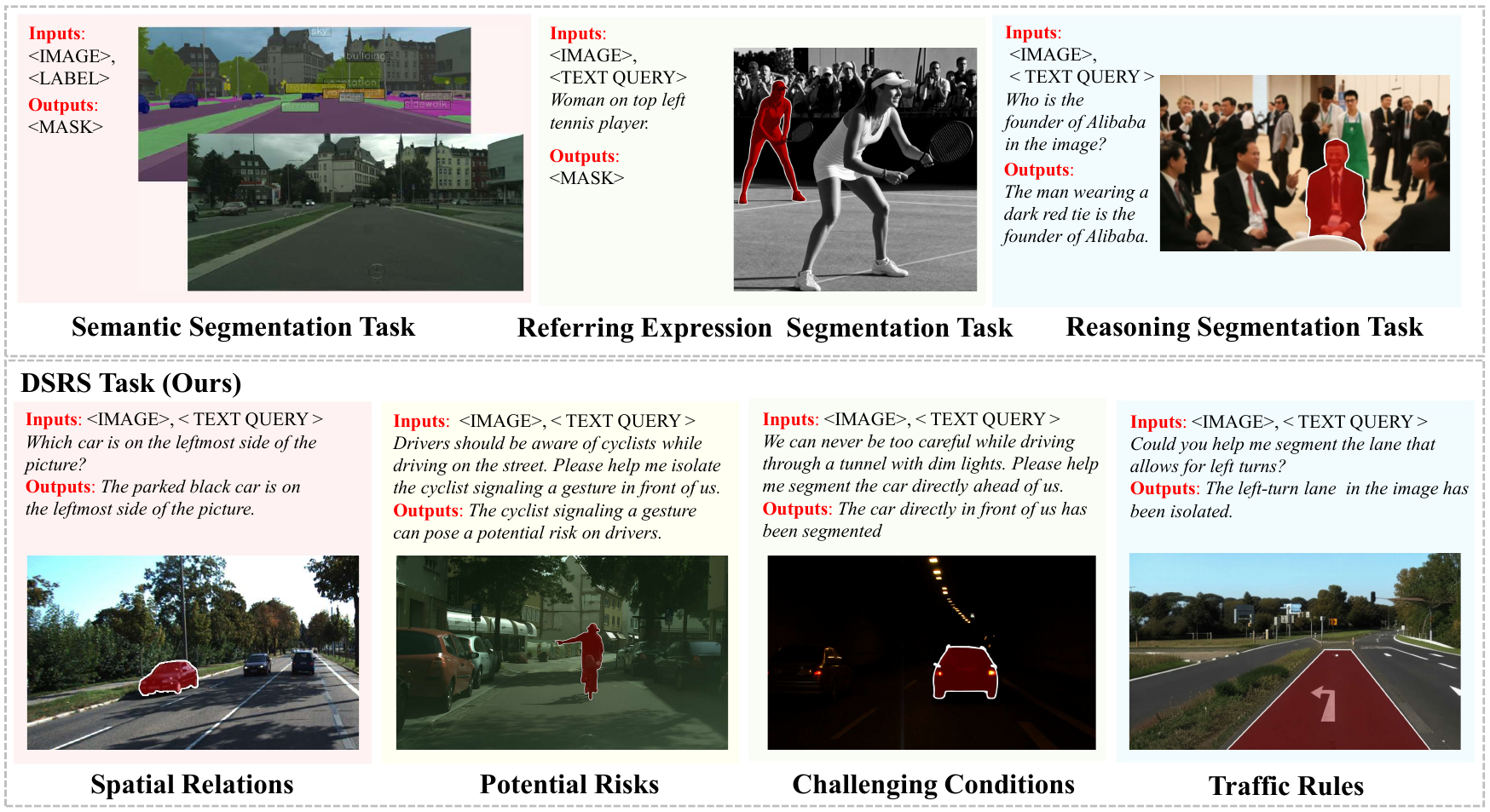}
\captionof{figure}{
  Comparisons between driving scene reasoning segmentation and other segmentation tasks.
}
\vspace{2em}
\label{fig.task}
\end{center}%
}]

\appendix
\begingroup
\small
\setlength{\parskip}{0pt}
\appendixsectionentry{A}{Ethics}{ethics}
\appendixsectionentry{B}{Additional Details on the \ourmethod Task and Dataset}{sec:task_dataset}
\appendixsubsectionentry{B.1}{Comparisons with Various Perception Tasks}{sec:task_comparison}
\appendixsubsectionentry{B.2}{Comparison with Existing Benchmark Datasets}{sec:benchmark_comparison}
\appendixsubsectionentry{B.3}{Comprehensive Analysis of \ourmethod Dataset}{sec:dataset_analysis}
\appendixsectionentry{C}{Additional Methodological Details}{sec:method_details}
\appendixsubsectionentry{C.1}{Implementation Details for \ournet Architecture}{sec:architecture_details}
\appendixsubsectionentry{C.2}{Depth-Aware Deformable Attention Blocks}{sec:architecture_da}
\appendixsubsectionentry{C.3}{Overall Loss}{sec:loss}
\appendixsectionentry{D}{Additional Discussion}{sec:discussion}
\appendixsubsectionentry{D.1}{Spatial Relation Inferring in MLLM-Based Models}{sec:mllm}
\appendixsubsectionentry{D.2}{Failure Cases}{sec:failure_cases}
\appendixsubsectionentry{D.3}{Extensibility and Broader Applications}{sec:broader_applications}
\appendixsubsectionentry{D.4}{Future Directions}{sec:future_directions}
\appendixsectionentry{E}{Technical Details}{sec:technical_details}
\appendixsubsectionentry{E.1}{Implementation Details}{sec:implementation_details}
\appendixsubsectionentry{E.2}{Training Datasets}{sec:training_datasets}
\appendixsubsectionentry{E.3}{Evaluation Metrics}{sec:evaluation_metrics}
\appendixsubsectionentry{E.4}{Competitors and Implementation Details}{sec:competitors}
\appendixsectionentry{F}{Additional Quantitative Experimental Results}{sec:quant_results}
\appendixsubsectionentry{F.1}{Depth Source, Quality, and Robustness Analysis}{sec:depth_source}
\appendixsubsectionentry{F.2}{Fair Comparison with Depth-Enhanced Baselines}{sec:depth_enhanced}
\appendixsubsectionentry{F.3}{Computational Efficiency and Deployment Analysis}{sec:efficiency}
\appendixsubsectionentry{F.4}{Additional Comparison on the ReasonSeg Benchmark}{sec:additional_reasonseg}
\appendixsubsectionentry{F.5}{Per-Category Results on the \ourmethod Dataset}{sec:per_category_results}
\appendixsubsectionentry{F.6}{Additional Ablation Studies}{sec:additional_ablation}
\appendixsectionentry{G}{Additional Qualitative Experimental Results}{sec:qual_results}
\endgroup
\par\medskip
\noindent\rule{\columnwidth}{0.4pt}
\medskip

\begin{table*}[!t]
\begin{center}
\setlength{\tabcolsep}{3.8mm}{
\settablefont
{
\begin{tabular}{l|cccccc}
\toprule
\makecell[l]{Dataset} & \makecell[c]{Input} & \makecell[c]{Output}
 & \makecell[c]{Lang.\\Source} & \makecell[c]{Scene} & \makecell[l]{Spatial\\Reasoning} &\makecell[l]{Traffic\\Rule}  \\
\hline
SQA3D \cite{ma2022sqa3d} & Lang, Image, Point cloud, Video & Lang & Human & IS & \cmark & \xmark \\
MSQA \cite{linghu2024multi} & Lang, Image, Point cloud & Lang & Human, LLM & IS & \cmark & \xmark \\
VLA‑3D \cite{zhang2024vla3ddataset3dsemantic} & Lang, Point cloud, Scene graph & Lang & Template & IS & \cmark & \xmark \\
Grounded 3D‑LLM \cite{chen2024grounded3dllm} & Lang, Point cloud & Mask & Template, LLM & IS & \cmark & \xmark \\
ScanEnts3D \cite{abdelreheem2022scanents}    & Lang, Point cloud & BBox & Human & IS & \cmark & \xmark \\
EmbodiedScan \cite{embodiedscan} & Lang, RGB‑D image & BBox & Template, LLM & IS & \cmark & \xmark \\
3D‑LLM \cite{hong20233d} & Lang, Point cloud & Lang, BBox & LLM &  IS & \cmark & \xmark \\
SegPoint \cite{SegPoint} & Lang, Point cloud & Mask       & LLM & IS & \cmark & \xmark \\
ScanReason \cite{zhu2024scanreasonempowering3dvisual} & Lang, RGB‑D image & BBox & LLM & IS & \cmark & \xmark \\
Reason3D \cite{reason3d} & Lang, Point cloud & Mask & LLM & IS & \cmark & \xmark \\
Surprise3D \cite{huang2025surprise3d} & Lang, Point cloud & Mask & Human, LLM & IS & \cmark & \xmark \\
\hline
nuScenes \cite{caesar2020nuscenes} & LiDAR, Image, Radar & BBox & - & Real DS & \xmark & \xmark \\
BDD100K  \cite{yu2020bdd100k}& Video & Mask, BBox & - & Real DS & \xmark & \xmark \\
BDD-X \cite{kim2018textual}& Lang, Video & Lang & Human & Real DS & \cmark & \cmark \\
DriveLM-Data \cite{sima2024drivelm} &Lang, Image & Lang & Template, Human & Real, Sim DS & \cmark & \cmark \\
Apollo \cite{wang2019apolloscape}& LiDAR, Image, Point cloud & Mask & - & Real DS & \xmark & \xmark \\
Bench2Drive \cite{jia2024bench} & LiDAR, Image & Action & - & Sim DS & \xmark & \xmark \\
Reason2Drive  \cite{nie2024reason2driveinterpretablechainbasedreasoning}& Video & Lang & Human,LLM & Real DS & \cmark & \xmark \\
ImpromptuVLA \cite{chi2025impromptu} & Video, State & Lang, Action & Human, LLM & Real DS & \cmark & \cmark \\
NuInteract \cite{zhao2025extending} & LiDAR, Image & Lang & Human, LLM & Real DS & \cmark & \xmark \\
DriveAction \cite{hao2025driveactionbenchmarkexploringhumanlike} & Lang, Video & Action & - & Real DS & \xmark & \xmark \\
\hline
RefCLEF \cite{kazemzadeh2014referitgame}& Lang, Image & BBox & Human & MS & \cmark & \xmark \\
RefCOCO \cite{kazemzadeh2014referitgame}& Lang, Image & Mask, BBox & Human & MS & \cmark & \xmark \\
RefCOCO+ \cite{kazemzadeh2014referitgame}& Lang, Image & Mask, BBox & Human & MS & \xmark & \xmark \\
gRefCOCO \cite{liu2023gres} & Lang, Image & Mask, BBox & Human & MS & \xmark & \xmark \\
SpatialQA \cite{cai2024spatialbot} & Lang, RGB-D image & Lang & Human & MS & \cmark & \xmark \\
Flickr30k \cite{10.1162/tacl_a_00166}& Image & Lang & Human & MS & \xmark & \xmark \\
Flickr 30k entities \cite{Plummer_2015_ICCV}& Lang, Image & BBox & Human & MS & \cmark & \xmark \\
ReasonSeg \cite{lai2024lisa}& Lang, Image & Lang, Mask & Human, LLM & MS & \xmark & \xmark \\
Visual Genome\cite{krishna2017visual} & Lang, Image & BBox & Human & MS & \cmark & \xmark \\
VGPhraseCut \cite{Lin_2016_CVPR}& Lang, Image & Mask & Human & MS & \xmark & \xmark \\
RefEgo \cite{Kurita_2023_ICCV}& Video, Lang & BBox & Human & MS & \xmark & \xmark \\

\hline
\rowcolor[HTML]{DCDCDC}
\textbf{\ourmethod (ours)} & Lang, RGB-D image & Lang, Mask & Human, LLM & Real DS & \cmark & \cmark \\
\bottomrule
\end{tabular}
}
\caption{Comparison with Other Vision Datasets. `Lang’ denotes language queries, `Lang Source’ denotes whether language queries are human-annotated or generated via template or LLM, `IS’ denotes indoor scenes, `DS’ denotes driving scenes, and `MS’ denotes multi-scene.} 
\label{tb.diff}
}\end{center}
\end{table*}

\section{Ethics}
\label{ethics}
In this research, we utilize ten public datasets for the development and evaluation of computer vision models: ADE20K \cite{zhou2017scene}, RefCOCO, RefCOCO+  \cite{kazemzadeh2014referitgame}, RefCOCOg \cite{mao2016generation}, SpatialQA \cite{cai2024spatialbot}, ReasonSeg \cite{lai2024lisa}, Cityscapes \cite{cordts2016cityscapes}, KITTI Semantics \cite{menze2015kitti}, nuScenes \cite{caesar2020nuscenes}, and ApolloScape \cite{wang2019apolloscape}. It is important to emphasize that we adhere strictly to the terms of usage for each dataset. We acknowledge that these datasets may contain images with visible faces and other personal data collected without consent. However, we assure that no biometric information has been processed. All images are used in accordance with the CC-BY license or in a manner compatible with the Data Analysis Permission.

\begin{table}[t]
{
\centering
\setlength{\tabcolsep}{0.6mm}
\settablefont
\begin{tabular}{@{}>{\raggedright\arraybackslash}p{0.30\linewidth}>{\raggedright\arraybackslash}p{0.50\linewidth}>{\raggedright\arraybackslash}p{\dimexpr0.20\linewidth-4\tabcolsep\relax}@{}}
\toprule
Dataset & Task type & Number of samples \\
\midrule
\rowcolor[HTML]{EFEFEF}
ReasonSeg \cite{lai2024lisa} & Open-domain reasoning segmentation & 1,218 \\
ReasonVOS \cite{bai2024one} & Video reasoning segmentation & 458 \\
\rowcolor[HTML]{EFEFEF}
FCoT \cite{you2025seg} & Foreground reasoning and segmentation-prompt generation & 1,500 \\
ConverSeg \cite{sahoo2026conversational} & Conversational image segmentation & 1,687 \\
\rowcolor[HTML]{EFEFEF}
ReasonSeg-X \cite{yun2026star} & Extended reasoning segmentation & 1,169 \\
OK-VOS \cite{liang2026seg} & Outside-knowledge video object segmentation & 1,000 \\
\midrule
\rowcolor[HTML]{EFEFEF}
\textbf{\ourmethod (ours)} & Driving scene reasoning segmentation & 1,561 \\
\bottomrule

\end{tabular}
\caption{Scale comparison with representative reasoning and instruction segmentation datasets.}
\label{tb.dataset_scale}
}
\end{table}

\section{Additional Details on the \ourmethod Task and Dataset}
\label{sec:task_dataset}

\subsection{Comparisons with Various Perception Tasks}
\label{sec:task_comparison}

Semantic segmentation offers pixel-level predictions, facilitating fine-grained scene understanding in applications such as medical imaging. However, its fundamental limitation lies in its closed-set training paradigm, which restricts the model to a fixed vocabulary of predefined classes \cite{wu2024towards}. This inherent inflexibility prevents it from accommodating user-specific requirements or interpreting ambiguous, open-world instructions that lie outside its training distribution.

To overcome these limitations, referring expression segmentation \cite{hu2016segmentation} has been introduced, which leverages natural language expressions to guide the segmentation of a specific object within an image. It enables more interactive and context-aware segmentation that aligns better with human intent. For instance, as illustrated in Fig. \ref{fig.task} in the supplementary material, instead of segmenting all instances of ``women'' in an image, a referring expression segmentation model can isolate ``woman on top left tennis player''.

Existing referring expression segmentation models are optimized for relatively simple expressions and often struggle with complex descriptions. This has motivated the research on reasoning segmentation, which requires reasoning capabilities to identify regions that satisfy abstract or indirect requirements specified in text queries. 

Nevertheless, the reasoning segmentation task has largely focused on broad representations of world knowledge and exhibits limited scene-specific awareness in driving scenarios \cite{kabir2025logic}. This limitation motivates us to introduce the driving scene reasoning segmentation, a new multimodal perception task specifically designed to address the scene-specific challenges in automated driving contexts, as illustrated in Fig. \ref{fig.task} in the supplementary material.

\begin{figure*}[t]
    \centering
    \includegraphics[width=0.8\textwidth]{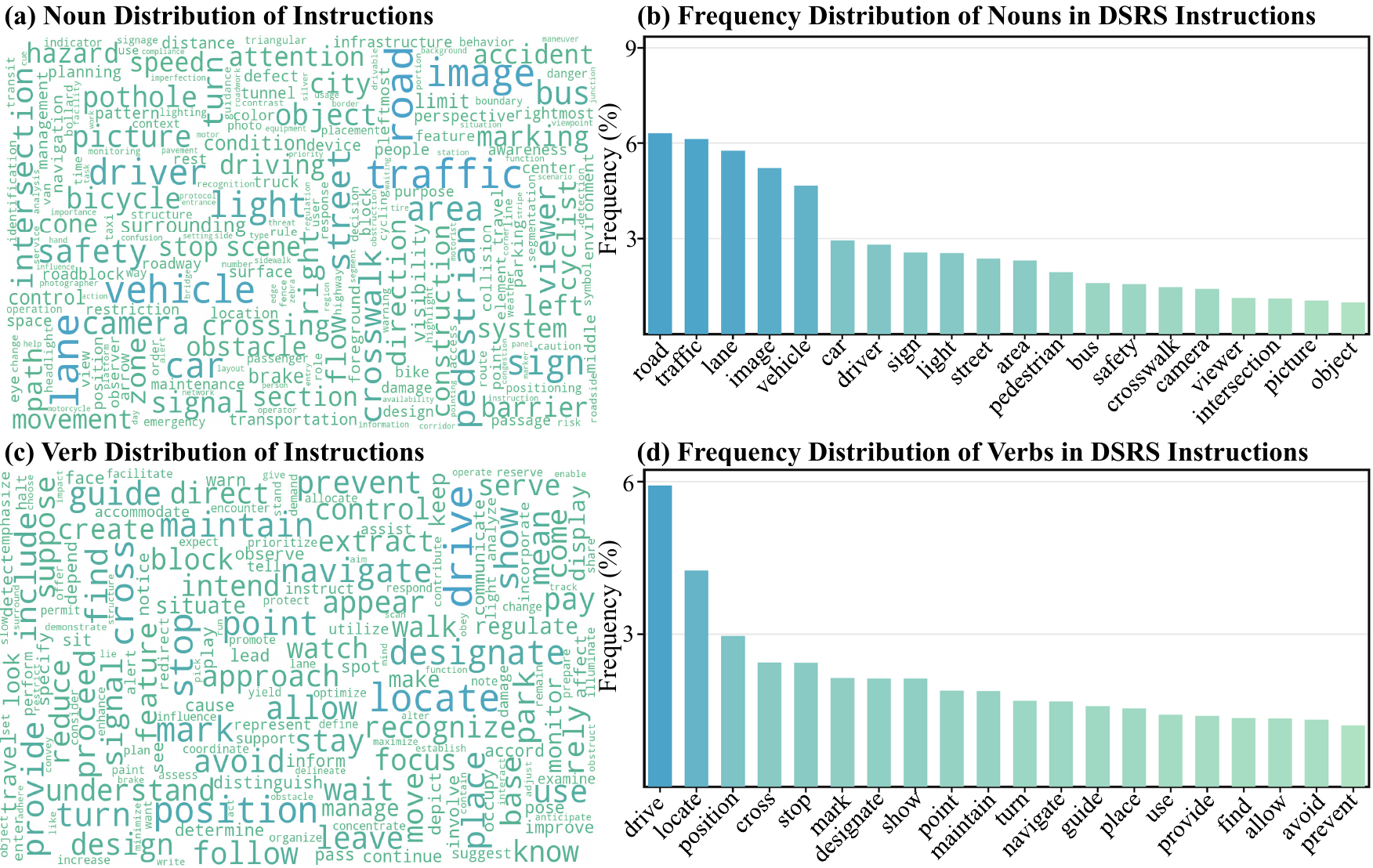}
    \vspace{-0.6em}
    \caption{Word distribution of instructions in the \ourmethod dataset. (a) and (c) illustrate word clouds of nouns and verbs appearing in the instructions of the \ourmethod dataset, respectively. (b) and (d) show the frequency distributions of representative nouns and verbs, respectively.}
    \label{fig:verb_noun_frequency}
\end{figure*}

\subsection{Comparison with Existing Benchmark Datasets}
\label{sec:benchmark_comparison}

As illustrated in Table \ref{tb.diff} in the supplementary material, most existing multimodal datasets that support spatial reasoning are primarily focused on indoor environments, where geometric priors are typically obtained from point clouds or bird's eye view (BEV) images, such as in SQA3D \cite{ma2022sqa3d} and MSQA \cite{linghu2024multi}. Although these datasets have contributed valuable insights into 3D spatial reasoning, they are limited in terms of object diversity and scene complexity, making them insufficient for modeling the rich, dynamic environments of the real world. For instance, in driving scenes, spatial relation inference is particularly critical, as it underpins essential tasks such as estimating object distances, interpreting road structures, and predicting dynamic behaviors.

In scene parsing tasks, pixel-level mask prediction serves as a critical foundation for precise object localization. For instance, given a text query such as ``Avoid the maintenance roads with irregular shapes'', the absence of pixel-level output makes it challenging for the system to precisely localize the referenced areas within the scene. While existing language-driven datasets, \eg, RefCOCO \cite{kazemzadeh2014referitgame} and VGPhraseCut \cite{Lin_2016_CVPR}, offer segmentation masks, such resources remain scarce in the context of automated driving. Datasets such as DriveLM-Data \cite{sima2024drivelm} and BDD-X \cite{kim2018textual} incorporate language prompts but only support coarse-grained textual outputs. Conversely, datasets like nuScenes \cite{caesar2020nuscenes} and BDD100K \cite{yu2020bdd100k} provide segmentation masks but lack alignment with language inputs. 
To date, there is no multimodal dataset that simultaneously supports natural language input and pixel-level mask output in real-world mixed traffic scenarios. This remains a critical bottleneck in advancing automated driving systems.

To fill the gap in existing multimodal datasets, we introduce the \ourmethod dataset. To the best of our knowledge, it is the first dataset in real-world mixed traffic scenes to jointly provide both text queries and segmentation masks, thereby filling a critical gap in the advancement of multimodal automated driving systems.

Dataset scale should be interpreted in relation to annotation granularity and intended usage. \ourmethod dataset is not intended as the sole corpus for training a 7B--8B MLLM from scratch, rather, it is a high-quality, domain-specific benchmark for evaluating and adapting pretrained MLLMs. Unlike large-scale pretraining corpora, constructing a pixel-level reasoning segmentation benchmark requires every sample to combine an unambiguous instruction, a reasoning-grounded target, and an accurate pixel-level mask, making annotation and multi-stage manual verification substantially more expensive. This is especially true for driving scenes, where annotations must also preserve traffic semantics, spatial relations, and safety-critical context. As shown in Table~\ref{tb.dataset_scale}, several representative image-based and video-based reasoning or instruction segmentation benchmarks are constructed within a comparable 1K-scale range, with some being substantially smaller. Following this high-quality benchmark setting, \ourmethod provides 1,561 carefully curated samples for evaluating pretrained MLLMs and supporting their parameter-efficient adaptation to driving-scene reasoning segmentation.

\begin{figure*}[t]
    \centering
    \includegraphics[width=0.99\linewidth]{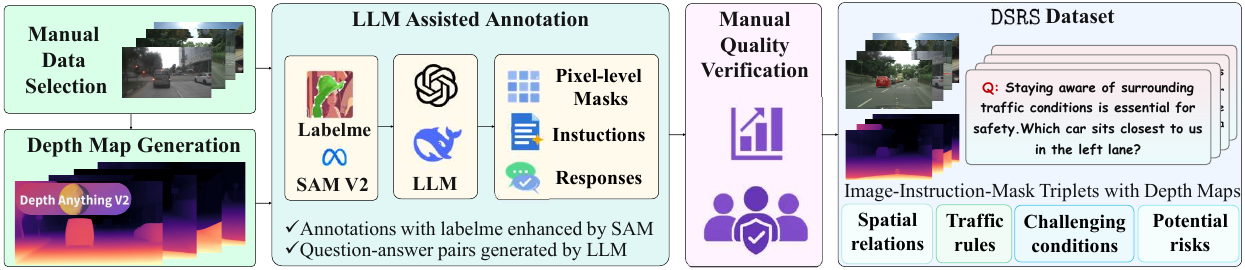}
    \vspace{-0.8em}
    \caption{Construction and Annotation Pipeline of Our \ourmethod Dataset.}
    \label{fig:DSRS_Pipeline}
\end{figure*}

\subsection{Comprehensive Analysis of \ourmethod Dataset}
\label{sec:dataset_analysis}

\subsubsection{Fine-Grained Category Definitions}
To comprehensively represent diverse road conditions encountered in complex driving scenarios, the dataset is categorized into the following categories:
\begin{itemize}
    \item \textit{Spatial relations}. This category includes samples that involve diverse spatial relations among objects within complex driving environments.
    \item \textit{Potential risks}. This category comprises samples involving traffic accidents and various potential hazards on the road, including both positive and negative obstacles, \eg, obstacles and road surface defects.
    \item \textit{Challenging conditions}. This category includes samples captured under adverse environmental conditions, such as extreme lighting, severe weather, and complex terrain.
    \item \textit{Traffic rules}. This category includes samples that reflect traffic rule enforcement via regulatory infrastructure elements, including dedicated lanes, crosswalks, traffic lights, and traffic signs.
\end{itemize}
The proposed \ourmethod dataset offers a comprehensive benchmark for evaluating and advancing the reasoning capabilities of models in automated driving scenarios.

\subsubsection{Word Distribution of Instructions}
\label{sec:word_distribution}

To further analyze the language features of the \ourmethod dataset, Fig. \ref{fig:verb_noun_frequency} presents the word distribution of nouns and verbs in the instructions of the \ourmethod dataset. As shown in Figs. \ref{fig:verb_noun_frequency}(a) and (b), the noun distribution reveals that instructions primarily involve road structures and road users. In particular, ``road'' and ``traffic'' account for approximately 6.3\% and 6.1\% of all nouns in the instructions, respectively. Other frequent nouns, including ``lane'', ``vehicle'', ``pedestrian'', ``sign'', and ``crosswalk'', indicate that instructions comprehensively reflect the complex road conditions in real-world driving scenes. In parallel, the verb distributions illustrated in Figs. \ref{fig:verb_noun_frequency}(c) and (d) show that these instructions are strongly associated with driving behaviors, operational requirements, and safety-critical scenarios. ``Drive'' accounts for 5.9\% of all verbs in the instructions, followed by ``locate'' at 4.3\% and ``position'' at 3.0\%. Verbs such as ``turn'', ``navigate'', and ``guide'' further demonstrate that models are required to interpret implicit instructions, identify the referred regions, and understand corresponding driving scenes. Overall, the word distribution depicted in Fig. \ref{fig:verb_noun_frequency} demonstrates that instructions of the \ourmethod dataset extend beyond explicit object descriptions and capture contextual details, which makes the \ourmethod dataset well-suited for evaluating models in automated driving scenarios. The dataset contains both concise queries and long compositional queries. Concise queries assess direct grounding, while long queries require multiple reasoning steps over spatial relations, traffic rules, and potential risks. For example, the prompt, ``In conditions of reduced visibility, what element above in the picture helps regulate traffic flow and ensure safety?'', requires the model to identify candidate objects in the image, infer which object serves to regulate traffic flow and enhance road safety, and accurately segment the corresponding region.
This variation helps distinguish genuine reasoning from superficial matching between words and visible objects.

\subsubsection{\ourmethod Dataset Construction and Annotation Pipeline}
\label{sec:dataset_construction}

As illustrated in Fig. \ref{fig:DSRS_Pipeline}, we present the construction pipeline of the \ourmethod dataset, which consists of five stages: image selection and depth map generation, pixel-level annotation, instruction and response generation, manual quality verification, and sample categorization. We exclude adjacent frames from different splits and assign every continuous scene to only one of the training, validation, and test sets. This scene-level separation prevents visual overlap and eliminates leakage caused by temporally neighboring observations.
To comprehensively cover diverse scenes and complex conditions in real-world driving scenarios, we select RGB images from multiple autonomous driving datasets and employ Depth Anything V2 \cite{yang2024depth} to obtain corresponding relative depth maps. To reduce dependence on country-specific road conventions, we retain samples whose answers are supported by explicit visual indicators such as lane markings, traffic lights, traffic signs, and clearly visible road structures. Samples that could produce ambiguous masks under different regional conventions are excluded during manual verification.
We then use LabelMe, which is assisted by SAM 2, to manually annotate the target regions in the selected images, producing pixel-level masks with accurate target boundaries.
After completing pixel-level annotations, we utilize large language models such as GPT-4o and DeepSeek-V3 to generate multiple implicit reasoning instructions for each target region. Subsequently, we generate corresponding responses for these instructions and organize them into question-answer pairs for reasoning chains.
To ensure the accuracy of the outputs produced in the preceding stages, we further conduct multiple rounds of manual quality verification on pixel-level masks, natural language instructions, and corresponding responses. Specifically, we check whether masks completely cover the target regions and whether the target boundaries are accurate. Meanwhile, we also verify whether the instructions are semantically consistent with the annotated regions while checking for repetition, ambiguity, and factual errors. For samples that do not pass verification, inaccurate masks are reannotated, while instructions or responses are revised accordingly.
Once all samples have passed manual quality verification, we categorize them into four high-level categories: \textit{Spatial Relations}, \textit{Potential Risks}, \textit{Challenging Conditions}, and \textit{Traffic Rules}. These categories comprise 13 fine-grained categories, which comprehensively represent diverse road conditions in real-world driving scenes.
By completing the five stages above, we obtain the \ourmethod dataset, which contains more than 1,500 image-instruction-mask samples, together with corresponding relative depth maps. Unlike existing datasets, our proposed \ourmethod dataset integrates implicit reasoning instructions, relative depth maps, and pixel-level annotations, thereby enabling scene-specific reasoning and accurate segmentation in complex driving scenarios.

\subsubsection{Mask Scale Distribution}
\label{sec:mask_scale_distribution}

\begin{figure}[t!]
\centering
\includegraphics[width=0.49\textwidth]{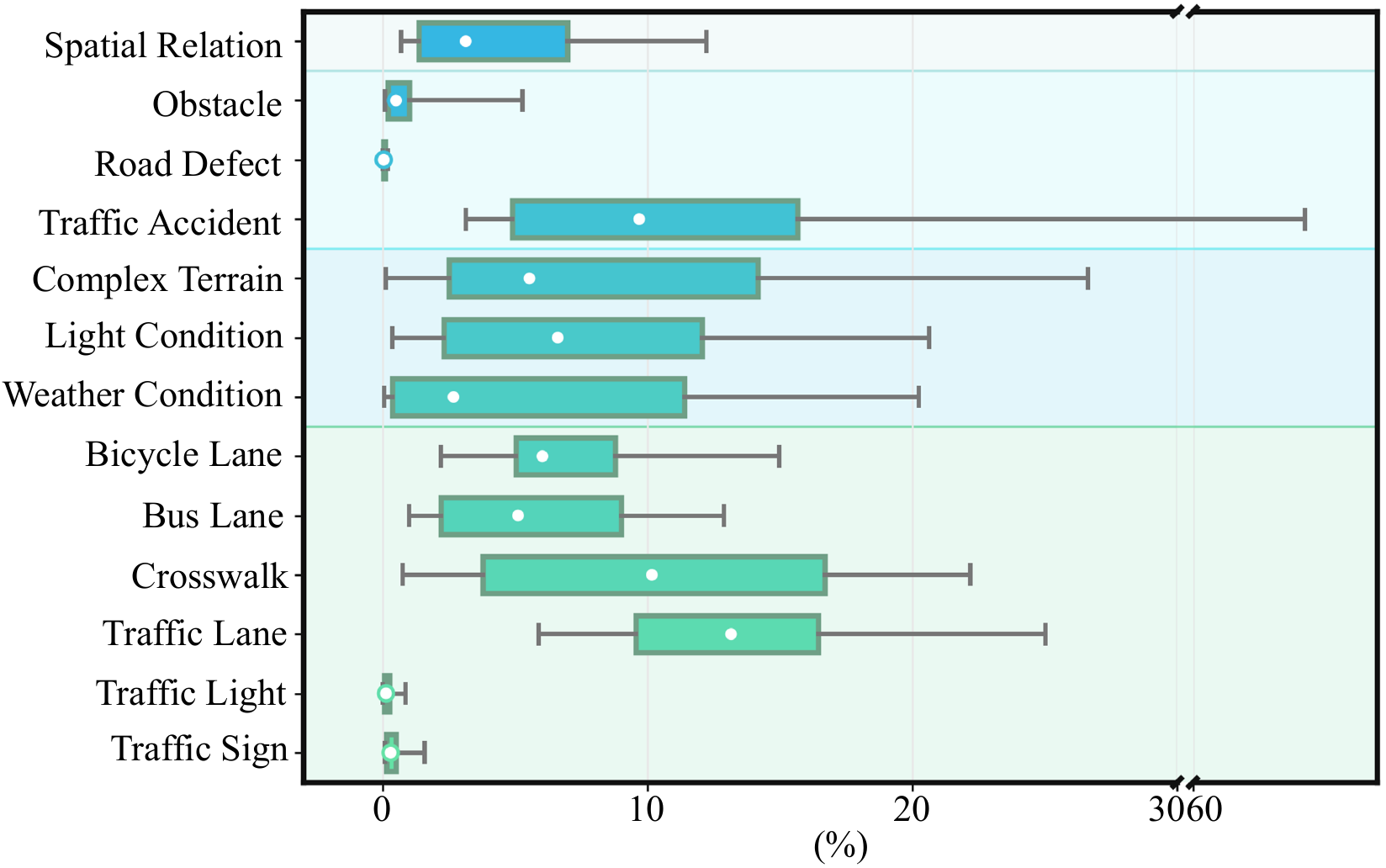}
\caption
{
Mask scale distribution across 13 fine-grained categories of our proposed \ourmethod dataset. White circles denote the median mask scale for each category. The lower extreme and the upper extreme represent the 5th and 95th percentiles, respectively. The horizontal axis is truncated between 30\% and 60\% for clarity.
}
\vspace{-0.5em}
\label{fig.mask_scale_distribution}
\end{figure}

\begin{table*}[t]
\centering
\setlength{\tabcolsep}{2.0mm}
\settablefont
\renewcommand{\arraystretch}{1.15}
\begin{tabular}{l|%
                >{\centering\arraybackslash}m{0.15\textwidth}%
                >{\centering\arraybackslash}m{0.15\textwidth}%
                >{\centering\arraybackslash}m{0.18\textwidth}|%
                >{\centering\arraybackslash}m{0.11\textwidth}|%
                >{\centering\arraybackslash}m{0.12\textwidth}}
\toprule[1pt]
\multirow{2}{*}{Model}
& \multicolumn{3}{c|}{Multimodal Encoder}
& \multirow{2}{*}{LLM}
& \multirow{2}{*}{Segmentation Head} \\
\cline{2-4}
& Vision Encoder
& Text Encoder
& Projector
& 
&  \\
\hline
\hline

LISA \cite{lai2024lisa}
& \makecell[c]{CLIP ViT-L/14,\\ SAM ViT-H}
& LLaMA Tokenizer
& MLP
& Vicuna 7B
& SAM decoder\\

\rowcolor[HTML]{EFEFEF}
GSVA \cite{xia2024gsva}
& \makecell[c]{CLIP ViT-L/14,\\ SAM ViT-H} 
& LLaMA Tokenizer
& MLP
& Vicuna 7B
& SAM decoder\\

PixelLM \cite{ren2024pixellm}
& CLIP ViT-L/14
& LLaMA Tokenizer
& MLP
& Vicuna 7B
& PixelLM decoder \\

\rowcolor[HTML]{EFEFEF}
M$^2$SA \cite{jang2025mmr}
& \makecell[c]{CLIP ViT-L/14,\\ SAM ViT-H}
& LLaMA Tokenizer
& MLP
& Vicuna 7B
& SAM decoder \\

SegAgent \cite{zhu2025segagent}
& \makecell[c]{Qwen-VL,\\ SClick ViT-L}
& Qwen-VL Tokenizer
& \makecell[c]{Qwen-VL adapter,\\ SClick prompt embedding}
& Qwen-VL 7B
& SClick decoder \\

\rowcolor[HTML]{EFEFEF}
Seg-R1 \cite{you2025seg}
& \makecell[c]{Qwen2.5-VL, \\ SAM2 Hiera-L}
& Qwen2.5-VL Tokenizer
& \makecell[c]{Qwen2.5-VL visual merger}
& Qwen2.5-VL 7B
& SAM2 decoder \\

ConverSeg-Net \cite{sahoo2026conversational}
& \makecell[c]{Qwen2.5-VL, \\ SAM2 Hiera-L}
& Qwen2.5-VL Tokenizer
& \shortstack[c]{MLP}
& Qwen2.5-VL 3B
& SAM2 decoder \\

\midrule

\rowcolor[HTML]{EFEFEF}
\textbf{\ournet-LLaVA1.5}
& CLIP ViT-L/14, GPE, DHFI
& LLaMA Tokenizer
& MLP
& Vicuna1.5 7B
& \ournet decoder \\

\rowcolor[HTML]{EFEFEF}
\textbf{\ournet-InternVL2.5}
& InternViT, GPE, DHFI
& InternVL2.5 Tokenizer
& MLP
& InternLM2.5 7B
& \ournet decoder \\

\bottomrule[1pt]
\end{tabular}
\caption{
Architectural comparison of representative language-guided segmentation
models. 
}
\label{tb.architecture}
\end{table*}

Fig. \ref{fig.mask_scale_distribution} presents the distribution of relative mask scales across 13 fine-grained categories of the \ourmethod dataset, where the relative mask scale is defined as the ratio of the annotated region to the entire image. The mask scale distribution demonstrates significant differences among categories. Most categories occupy less than 15\% of the entire image, which indicates that \ourmethod dataset contains many small targets and requires models to perform fine-grained scene inference and precise boundary modeling. Specifically, the mask scales of ``Road defect'', ``Traffic light'', ``Traffic sign'', and ``Obstacle'' range from approximately 0.04\% to 0.51\% of the entire image, placing greater demands on localizing targets precisely and extracting boundaries accurately. In contrast, the pixel-level annotated masks of categories such as ``Traffic lane'', ``Crosswalk'', ``Traffic accident'' cover larger scales of entire images, requiring models to understand traffic rules in more complex real-world road conditions. Notably, ``Traffic accident'' demonstrates the widest range of mask scales, ranging from 2.02\% to 72.06\% of the entire image. Moreover, ``Complex Terrain'', ``Light Condition'', and ``Weather Condition'' also exhibit broad mask scale distributions, reflecting pronounced differences in the mask scales of targets under challenging conditions. Overall, we form a challenging benchmark with diverse mask scales for evaluating scene-specific reasoning and accurate segmentation tasks in real-world automated driving scenarios.

\section{Additional Methodological Details}
\label{sec:method_details}

\subsection{Implementation Details for \ournet Architecture}
\label{sec:architecture_details}
As shown in Table \ref{tb.architecture}, our \ournet architecture comprises three key components: (1) a multimodal encoder, (2) an LLM, and (3) a segmentation head. The multimodal encoder consists of a vision encoder and a text encoder. The vision encoder extracts complementary information from the RGB-D images, which consists of an MLLM vision encoder, a geometric prior extractor (GPE), a depth-guided hierarchical feature interaction (DHFI) module, and projectors for modality alignment. 
Unlike conventional depth-guided segmentation methods that fuse RGB and depth features without an LLM, our design must preserve the visual-language balance already learned by the MLLM. GPE extracts structured geometric cues, while DHFI aligns them with hierarchical visual semantics before they enter the language space. This process enables the MLLM to interpret relative depth information such as spatial ordering, object boundaries, and scene layout without disrupting its original image-text alignment.

\subsection{Depth-Aware Deformable Attention Blocks}
\label{sec:architecture_da}

For reproducibility, given $N$ patch tokens, we concatenate the normalized pairwise depth differences and 2D relative positions to form candidate descriptors of shape $B\times N\times N\times3$, such that each query has $N$ candidates. A learned linear layer with $0.1$ dropout regresses the flattened descriptors into offsets of shape $B\times N\times M\times L\times K\times2$. We use $M=8$ heads, $L=1$ feature level, and $K=4$ sampling points, yielding 32 two-dimensional offsets per query. Although the offsets are not directly clipped, the depth differences are bounded by $\tanh$, and the final sampling coordinates are clamped to $[0,1]$. Bilinear interpolation is applied to value features projected from 1024 to 512 channels (64 channels per head), after which the output is projected back to 1024 channels. Four $\text{DA}^2$ blocks are stacked in total.

\subsection{Overall Loss}
\label{sec:loss}
In summary, the overall loss function can be formulated as a weighted combination of the text generation loss $\mathcal{L}_{\text{txt}}$ and the segmentation mask loss $\mathcal{L}_{\text{mask}}$, as follows:
\begin{equation}
\mathcal{L}
=
\lambda_{\text{txt}}\mathcal{L}_{\text{txt}}
+
\lambda_{\text{mask}}\mathcal{L}_{\text{mask}},
\label{eq:loss}
\end{equation}
where $\lambda_{\text{txt}}$ and $\lambda_{\text{mask}}$ balance the contributions of the text generation and segmentation losses, respectively. Specifically, we adopt the standard token-level cross-entropy loss $\mathcal{L}_{\text{txt}}$ for autoregressive text generation. For the segmentation branch, we define $\mathcal{L}_{\text{mask}}$ as a weighted sum of the sigmoid focal loss and the Dice loss, with weights $\lambda_{\text{focal}}$ and $\lambda_{\text{dice}}$. Given the ground-truth text tokens $\boldsymbol{Y}_{\text{txt}}$ and segmentation mask $\boldsymbol{M}$, the losses are formulated as follows:
\begin{equation}
\begin{gathered}
\mathcal{L}_{\text{txt}}
=
\operatorname{CE}
\left(
\widetilde{\boldsymbol{Y}}_{\text{txt}},
\boldsymbol{Y}_{\text{txt}}
\right),\\
\mathcal{L}_{\text{mask}}
=
\lambda_{\text{focal}}
\operatorname{Focal}
\left(
\widetilde{\boldsymbol{M}},\boldsymbol{M}
\right)
+
\lambda_{\text{dice}}
\operatorname{Dice}
\left(
\widetilde{\boldsymbol{M}},\boldsymbol{M}
\right).
\end{gathered}
\end{equation}
Here, the sigmoid focal loss is implemented using binary cross-entropy with logits and a focal modulation term. By minimizing Eq.~(\ref{eq:loss}), our \ournet is jointly optimized for autoregressive text generation and accurate segmentation mask prediction.

\section{Additional Discussion}
\label{sec:discussion}
\subsection{Spatial Relation Inferring in MLLM-Based Models}
\label{sec:mllm}
Although recent advancements in MLLMs have demonstrated impressive performance in language-driven tasks, their abilities to infer spatial relations remain a critical challenge.
Early efforts \cite{ding2023pla,hong20233d} first reconstruct 3D scenes from multi-view inputs like point clouds or neural fields, then feed these volumetric representations or scene graphs into language models to support spatial understanding. However, such pipelines are typically resource-intensive and require dense multi-view annotations, which are often unavailable in real-world scenarios. To this end, other approaches \cite{cai2024spatialbot,cheng2024spatialrgpt} incorporate depth maps alongside RGB images within standard vision-language pipelines, thereby bypassing explicit 3D reconstruction and enabling coarse-grained spatial relations inference. These methods are adequate for high-level textual reasoning but are inadequate for fine-grained vision tasks like pixel-level segmentation \cite{shen2025reasoning}. This limitation arises primarily because they often overlook the unique geometric properties required for accurate spatial reasoning and introduce modalities that remain unfamiliar to LLMs. To address these limitations, we propose \ournet, which leverages the DHFI module to structurally integrate hierarchical visual cues and geometric priors, thereby enhancing the model's capacity for scene cognitive segmentation in complex driving environments.

\begin{figure}[t!]
\centering
\includegraphics[width=0.485\textwidth]{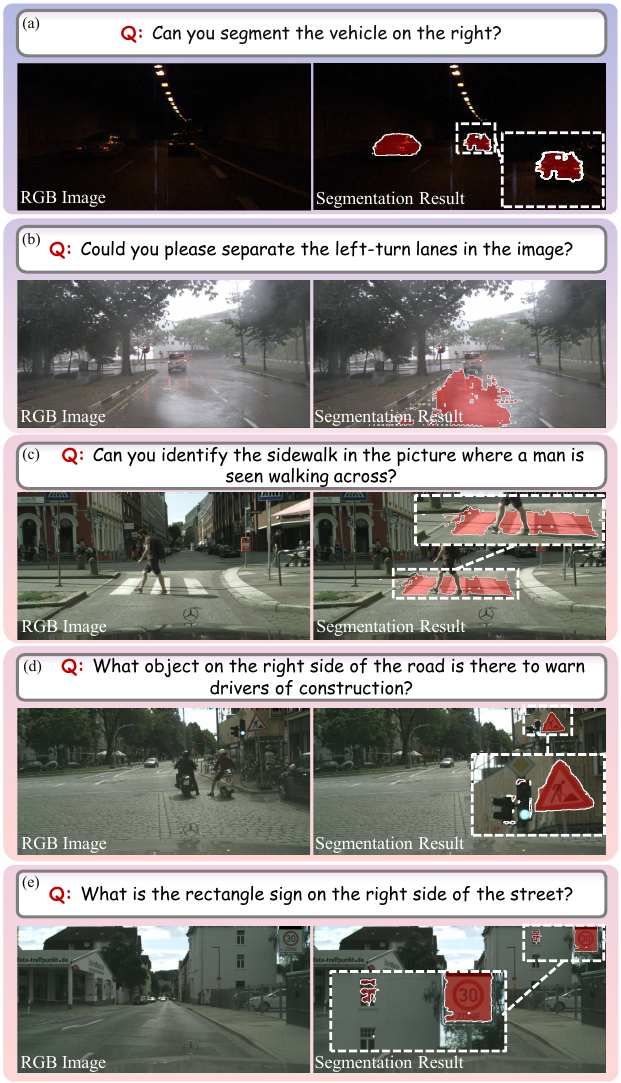}
\centering
\caption{Unsatisfactory results on the \ourmethod dataset: (a) areas with insufficient illumination; (b) blurred scene under rainy conditions; (c) areas with indistinct boundaries; (d) additional areas of erroneous segmentation; (e) contextual confusions. Challenging areas are highlighted with white dashed boxes.}
\label{fig.challenge}
\end{figure} 

\subsection{Failure Cases}
\label{sec:failure_cases}
While \ournet demonstrates significant progress, it still faces challenges in several scenarios. As illustrated in Fig. \ref{fig.challenge}(a), segmentation performance degrades in low-light regions, where objects cannot be clearly distinguished. In challenging weather conditions, such as the blurred rainy scene in Fig. \ref{fig.challenge}(b), the network struggles to effectively integrate geometric features for robust scene reasoning. Furthermore, when targets lack well-defined boundaries, such as the sidewalk in Fig. \ref{fig.challenge}(c), the model can localize the region but fails to produce accurate boundaries. A similar issue arises in Fig. \ref{fig.challenge}(d), where although the traffic sign is correctly segmented, additional areas of erroneous segmentation appear around it. In addition to segmentation quality, the network’s contextual reasoning also requires improvement. For instance, in Fig. \ref{fig.challenge}(e), \ournet mistakenly segments a window together with the traffic sign, likely due to confusion caused by the textual cue ``rectangle''. Addressing these limitations in both segmentation precision and contextual understanding presents promising directions for future research and optimization.

\subsection{Extensibility and Broader Applications}
\label{sec:broader_applications}
Moreover, the modular nature of the DHFI module allows it to be readily extended beyond \ournet and applied to other related tasks. Serving as an auxiliary branch, it effectively incorporates spatial information and can be seamlessly embedded into various backbone architectures. For instance, it can be incorporated into LLM-based frameworks to strengthen multimodal feature alignment. Additionally, its $\text{DA}^2$ submodule can serve as a depth-aware alternative to conventional cross-attention mechanisms, thereby strengthening the interaction of spatial features within transformer-based backbones. Furthermore, the integration of spatial cues benefits a wide range of vision tasks, including 3D scene reasoning, embodied visual question answering, and vision-language navigation, where explicitly modeling spatial dependencies is critical for precise perception and reliable reasoning. 

\subsection{Future Directions}
\label{sec:future_directions}
Looking ahead, advancements in LLMs present new opportunities for robust scene reasoning and decision-making in automated driving. Drawing inspiration from DriveLM's \cite{sima2024drivelm} graph-based reasoning for sequential decision-making, a promising future direction for \ournet lies in an evolution from a static reasoning framework to a dynamic system capable of localization, prediction, and action planning. Such an evolution would not only enhance \ournet's adaptability and responsiveness to dynamic real-time challenges but also pave the way for more intelligent and interactive automated driving systems.

\section{Technical Details}
\label{sec:technical_details}
\subsection{Implementation Details}
\label{sec:implementation_details}
For instruction tuning, we optimize \ournet using AdamW with an initial learning rate of $2\times10^{-4}$ and a weight decay of $0.01$. We adopt a cosine learning-rate schedule with 1K warm-up iterations and clip the gradient norm to $1.0$. The model is trained with a batch size of $2$ and $16$ gradient-accumulation steps. We employ LoRA to efficiently fine-tune the LLMs. All experiments are conducted using BF16 precision on an NVIDIA A800 GPU. 

\subsection{Training Datasets}
\label{sec:training_datasets}
 Our training data comprises four primary dataset types: (1) For the semantic segmentation dataset, we use ADE20K \cite{zhou2017scene}, Mapillary \cite{Neuhold_2017_ICCV} and COCO-Stuff \cite{caesar2018coco}; (2) For the referring expression segmentation dataset, we use RefCOCO, RefCOCO+  \cite{kazemzadeh2014referitgame}, and RefCOCOg \cite{mao2016generation}; (3) Visual question answering (VQA) datasets \cite{liu2023visual, li2024llava}; (4) For the reasoning segmentation dataset, we use the ReasonSeg \cite{lai2024lisa}. Notably, to ensure a zero-shot evaluation shown in Table 1 in the main text, the above training data excludes all samples from the \ourmethod dataset and is used exclusively for fine-tuning. 

\subsection{Evaluation Metrics}
\label{sec:evaluation_metrics}
We quantify the \ourmethod performance of models using the variations of intersection over union (IoU) metric, specifically cumulative intersection over union (cIoU) and generalized intersection over union (gIoU) metrics. 
\subsubsection{cIoU}
\label{sec:ciou}
The cIoU metric aims to evaluate segmentation performance at the dataset level. For a given prediction mask $\boldsymbol{P}$ and its corresponding ground truth mask $\boldsymbol{G}$, the IoU metric is calculated as the ratio of their intersection to the area of their union as follows:
\begin{equation}
\mathrm{IoU}=\frac{|\boldsymbol{P} \cap \boldsymbol{G}|}{|\boldsymbol{P} \cup \boldsymbol{G}|} .
\end{equation}
The cIoU extends IoU by aggregating intersection and union values across all samples before computing their ratio expressed as follows:
\begin{equation}
\mathrm{cIoU}=\frac{\sum_{i=1}^N\left|\boldsymbol{P}_i \cap \boldsymbol{G}_i\right|}{\sum_{i=1}^N\left|\boldsymbol{P}_i \cup \boldsymbol{G}_i\right|},
\end{equation}
where $N$ represents the total number of samples in the dataset, and $\boldsymbol{P}_i$ and $\boldsymbol{G}_i$ denote the prediction and ground truth masks for the $i$-th sample, respectively. This cumulative computation can prevent individual samples with small regions from disproportionately affecting the overall performance metric.

\subsubsection{gIoU}
\label{sec:giou}
The gIoU metric offers a complementary perspective to cIoU by evaluating segmentation performance at the instance level. Unlike cIoU, which provides a dataset-level assessment and favors larger objects, gIoU computes the mean IoU across all samples, treating each instance equally regardless of its size, formulated as follows:
\begin{equation}
\mathrm{gIoU}=\frac{1}{N}\sum_{i=1}^N\frac{\left|\boldsymbol{P}_i \cap \boldsymbol{G}_i\right|}{\left|\boldsymbol{P}_i \cup \boldsymbol{G}_i\right|},
\end{equation}
This metric can, therefore, evaluate the model's consistency across different instances and scales, as it gives equal weight to each prediction regardless of the object's size. In cases where the ground truth mask is empty, and the model correctly predicts no object ( $\left|\boldsymbol{P}_i \cup \boldsymbol{G}_i\right|=$ 0 ), the instance is assigned a perfect score of $\frac{\left|\boldsymbol{P}_i \cap \boldsymbol{G}_i\right|}{\left|\boldsymbol{P}_i \cup \boldsymbol{G}_i\right|}=1.0$ to avoid penalizing the model for a correct negative prediction.

\subsection{Competitors and Implementation Details}
\label{sec:competitors}
In this section, we provide a detailed summary of the models evaluated in Table 1 of the main text.

\setlength{\parskip}{0.45em}
\noindent\textbf{ReLA} \cite{liu2023gres} proposes a region-based baseline for generalized referring expression segmentation, where one expression may refer to single, multiple, or no target objects. It adaptively divides an image into regions with sub-instance clues and explicitly models both region-region and region-language dependencies to handle complex relational expressions. For implementation, we follow the official repository with the Swin-Base backbone and default inference pipeline. We obtain the zero-shot results using the released model weights, while the fine-tuned results are obtained by supervised fine-tuning on our \ourmethod dataset under its default training configuration.

\noindent\textbf{X-Decoder} \cite{zou2023generalized} presents a generalized decoding model that jointly predicts pixel-level segmentation masks and language tokens in a unified semantic space. It uses generic non-semantic queries and text-induced semantic queries to support diverse segmentation and vision-language tasks, enabling transferable pixel-level visual-semantic understanding across zero-shot and fine-tuning settings. For implementation, we follow the official repository with the Focal-L backbone and default evaluation pipeline. In the zero-shot setting, we directly evaluate the released X-Decoder checkpoint; for the fine-tuned setting, we initialize from this checkpoint and adapt the model on our \ourmethod dataset using its standard training setup.

\noindent\textbf{SEEM} \cite{zou2023segment} presents a promptable and interactive segmentation model that aims to segment everything in an image through a unified multimodal prompt interface. It introduces a decoding mechanism that supports diverse prompts, including points, boxes, scribbles, masks, text, and their combinations, while learning a joint visual-semantic space and using memory prompts to preserve interaction history for multi-round segmentation. For implementation, we follow the official repository and use SEEM\_v1 with the SAM-ViT-L backbone. The zero-shot setting evaluates the released SEEM checkpoint, and the fine-tuned setting applies supervised fine-tuning on our \ourmethod dataset under its default training configuration.

\noindent\textbf{Grounded-SAM} \cite{liu2024grounding} assembles Grounding DINO and Segment Anything Model into an open-world detection-and-segmentation pipeline, where arbitrary text inputs are first grounded as object boxes and then converted into pixel-level masks by SAM. This modular design enables text-guided segmentation of open-set regions and can be further connected with other vision models for automatic annotation, image editing, and multimodal visual analysis. For implementation, we follow the official repository and use Grounding DINO with the Swin-T backbone and SAM with the ViT-H backbone as the detector-segmenter pipeline. For zero-shot evaluation, we feed each instruction as the text prompt and follow the default inference pipeline for box prediction and mask generation. In the fine-tuned setting, we adapt the detector-segmenter pipeline to our \ourmethod dataset with the official training setup.

\noindent\textbf{LISA} \cite{lai2024lisa} proposes a large language instructed segmentation assistant for reasoning segmentation, which produces segmentation masks from complex and implicit queries requiring visual reasoning and world knowledge. Architecturally, LISA consists of a vision branch and an MLLM branch. In the vision branch, a SAM vision backbone extracts dense, high-resolution visual features for the mask decoder. In the MLLM branch, LLaVA jointly processes the input image and text query using a visual encoder, a linear projector for feature alignment, and a language model for reasoning and response generation. To connect the two branches, LISA adds a \texttt{[SEG]} token and projects its last-layer MLLM embedding into the SAM prompt space via an MLP, following an embedding-as-mask paradigm. The SAM mask decoder combines this prompt with the SAM visual features to predict the final mask. For implementation, we follow the official repository and use the LISA-7B checkpoint. Its MLLM is LLaVA-Lightning-7B-v1-1, which uses CLIP ViT-L/14 as its visual encoder and Vicuna 7B as its language model, while SAM ViT-H serves as the segmentation backbone. The zero-shot setting evaluates this checkpoint through the default evaluation pipeline, while the fine-tuned setting performs supervised fine-tuning on our \ourmethod dataset following the official training recipe.

\noindent\textbf{GSVA} \cite{xia2024gsva} proposes a generalized segmentation vision assistant for generalized referring expression segmentation, targeting cases where one expression may refer to multiple objects or no valid target in the image. Architecturally, GSVA follows LISA and couples an MLLM branch with a Segmentation Foundation Model (SFM) branch. In the MLLM branch, LLaVA processes the input image and text using a vision encoder, a linear projector for feature alignment, and a decoder-based language model. In the SFM branch, a frozen high-resolution vision encoder extracts dense features, and a trainable mask decoder predicts masks from segmentation queries. GSVA extends the embedding-as-mask mechanism by generating multiple weight-sharing \texttt{[SEG]} tokens, each guided by its preceding referent phrase, and \texttt{[REJ]} tokens for absent targets. An MLP projects the embeddings of all \texttt{[SEG]} tokens into the SFM prompt space for mask decoding, whereas \texttt{[REJ]} tokens are discarded and assigned all-zero masks. For implementation, we follow the official repository and use GSVA-7B with LLaVA-Lightning-7B-v1-1 as the prompt encoder and SAM ViT-H as the segmentation model. LLaVA-Lightning-7B-v1-1 uses CLIP ViT-L/14 as its visual encoder and Vicuna 7B as its language model. The zero-shot setting evaluates the released pretrained GSVA-7B weights with the default evaluation pipeline, and the fine-tuned setting further applies supervised fine-tuning on our \ourmethod dataset under its default configuration.

\noindent\textbf{PixelLM} \cite{ren2024pixellm} proposes an efficient large multimodal model for pixel-level reasoning and understanding. Architecturally, it comprises a CLIP-ViT vision encoder, a language model, a segmentation codebook, and a lightweight pixel decoder. The final-layer visual features are aligned with the language space through a vision-to-language projector, while multi-scale features are mapped to the pixel decoder through a separate vision-to-decoder projector. The hidden embeddings of the generated codebook tokens encode target-specific information at different visual scales, and the pixel decoder combines them with the corresponding image features to progressively refine and fuse scale-wise masks. This design avoids an additional segmentation model such as SAM, while a target refinement loss further improves multi-target discrimination and mask quality. For implementation, we follow the official PixelLM repository and use PixelLM-7B initialized from LLaVA-Lightning-7B-v1-1. Its MLLM uses CLIP ViT-L as the vision encoder, an MLP projector for feature alignment, and Vicuna 7B as the language model, while the segmentation codebook and PixelLM decoder are used for mask generation. The zero-shot setting evaluates the released PixelLM-7B checkpoint, while the fine-tuned setting applies supervised fine-tuning on our \ourmethod dataset using the default training configuration.

\noindent\textbf{M$^2$SA} \cite{jang2025mmr} extends LISA for multi-target and multi-granularity reasoning segmentation with early local feature fusion and multiple \texttt{[SEG]} tokens. Architecturally, it consists of a vision branch and an MLLM branch. In the vision branch, the SAM vision encoder extracts both early-layer local features and last-layer global features. Two convolutional layers align the local features with the global features, which are then fused to provide fine-grained visual representations for mask decoding. In the MLLM branch, LLaVA encodes the image using a CLIP vision encoder, maps the visual features into the language embedding space through a vision-to-text projector, and jointly processes them with the text query using a language model. The MLLM generates a separate \texttt{[SEG]} token for each target, and an MLP prompt projector maps their hidden states into the SAM prompt space. The SAM mask decoder then combines these prompt embeddings with the fused SAM visual features to predict the corresponding segmentation masks. For implementation, we follow the official repository and use the M$^2$SA-7B checkpoint. Its MLLM is LLaVA-Lightning-7B-v1-1, which uses CLIP ViT-L/14 as its visual encoder and Vicuna 7B as its language model, while SAM ViT-H serves as the segmentation backbone. For zero-shot evaluation, we evaluate the released M$^2$SA-7B checkpoint with the default validation pipeline, and the fine-tuned setting further adapts the model on our \ourmethod dataset under its default training configuration.

\noindent\textbf{SegAgent} \cite{zhu2025segagent} introduces the Human-Like Mask Annotation Task (HLMAT), which formulates segmentation as a multi-step decision-making process that imitates human annotation trajectories. Architecturally, SegAgent couples an MLLM policy with an interactive segmentation model without introducing implicit mask tokens or modifying the MLLM output space. At each step, the current mask is overlaid on the original image and encoded by the frozen Qwen-VL ViT. A Qwen-VL vision-language adapter maps the resulting visual tokens into the language space, where Qwen jointly reasons over the visual state and target description to generate the next positive or negative click as text coordinates. SClick then uses its ViT-L visual backbone to update the segmentation mask from the accumulated clicks and previous mask, and the resulting mask is fed back to the MLLM for the next decision. For implementation, we follow the official two-stage pipeline and use SegAgent-Qwen initialized from Qwen-VL-Chat together with the pretrained SClick model. For the fine-tuned setting, since our \ourmethod dataset is not annotated in the HLMAT trajectory format, we freeze SegAgent-Qwen to retain its prompt-conditioned point selection ability while adapting the segmentation module.

\noindent\textbf{Seg-R1} \cite{you2025seg} uses reinforcement learning to enhance pixel-level understanding in large multimodal models by reducing dense mask prediction to sparse prompt generation. Architecturally, it couples a Qwen2.5-VL policy model with a SAM2 segmentation branch. In the MLLM branch, the Qwen2.5-VL vision encoder extracts image features, and its built-in visual merger maps them into the language space. Qwen then autoregressively generates a reasoning trace together with bounding boxes, foreground and background points, and their labels as ordinary text tokens. In the segmentation branch, the SAM2 image encoder extracts dense visual features, while its prompt encoder converts the generated boxes and points into prompt embeddings. The SAM2 mask decoder combines these embeddings with the image features to predict the final mask. Seg-R1 trains this prompt-generation policy with Group Relative Policy Optimization using format and segmentation rewards. For implementation, we follow the official repository and use the Seg-R1-7B checkpoint with Qwen2.5-VL as the policy model and SAM2 Hiera-L as the segmentation model. For zero-shot evaluation, we test the released checkpoint under the official Reasoning Segmentation protocol. For the fine-tuned setting, we start from Seg-R1-7B and conduct supervised fine-tuning on our \ourmethod dataset.

\noindent\textbf{ConverSeg-Net} \cite{sahoo2026conversational} targets conversational image segmentation by grounding intent-driven concepts involving functional and physical reasoning into pixel-accurate masks. Architecturally, it couples a SAM2 vision branch with a compact Qwen2.5-VL prompt encoder. In the vision branch, the frozen SAM2 image encoder processes the image once to produce a spatial embedding. In the vision-language branch, Qwen2.5-VL jointly encodes the image and conversational prompt and extracts final-layer hidden states for the text tokens and the end-of-sequence (EOS) token. A linear projector maps the text-token states to sparse prompt embeddings, while a two-layer MLP with a SiLU activation maps the EOS state to a dense prompt embedding that captures global image-text context. The SAM2 mask decoder uses bidirectional cross-attention to fuse these language-conditioned prompts with the spatial image embedding and predicts the final mask. For implementation, we follow the official repository and use the open-source ConverSeg-Net-3B checkpoint with Qwen2.5-VL as the prompt encoder. The zero-shot inference is performed with ConverSeg-Net-3B, and the fine-tuned setting initializes from this checkpoint and applies supervised fine-tuning on our \ourmethod dataset.

\section{Additional Quantitative Experimental Results}
\label{sec:quant_results}

\subsection{Depth Source, Quality, and Robustness Analysis}
\label{sec:depth_source}

We generate all depth maps offline using the Large variant of Depth Anything V2 \cite{yang2024depth}. The generated maps represent monocular relative depth rather than metric depth and are directly fed into our \ournet. Additionally, Table 4 of the main paper demonstrates that relative depth provides effective geometric cues. Compared with the RGB-only baseline, incorporating relative depth through GPE and DHFI improves gIoU from $53.0$ to $55.4$. Moreover, commonly used segmentation datasets, including ADE20K \cite{zhou2017scene}, COCO-Stuff \cite{caesar2018coco}, referring-segmentation benchmarks \cite{kazemzadeh2014referitgame, mao2016generation}, and ReasonSeg \cite{lai2024lisa}, do not provide ground-truth depth. Offline relative depth is therefore a more realistic and broadly applicable choice than assuming access to metric-depth annotations.

\begin{table}[t]
\centering
\setlength{\tabcolsep}{5.3mm}
\settablefont
\begin{tabular}{lcc}
\toprule[1pt]
Depth input & Perturbation & gIoU  \\
\midrule
None (RGB-only) & -- & 53.0  \\
Depth Anything V2-Small & None & 54.2\\
Depth Anything V2-Base & None & 54.9\\
Depth Anything V2-Large & None & 55.4  \\
\midrule
Depth Anything V2-Large & Gaussian ($\sigma= 0.2$) &  53.5 \\
Depth Anything V2-Large & Gaussian ($\sigma= 0.5$) &  48.8 \\
Depth Anything V2-Large & Gaussian ($\sigma= 0.8$) &  40.2 \\
\bottomrule[1pt]
\end{tabular}
\caption{Influence of depth sources and Gaussian perturbations on the \ourmethod dataset. Gaussian noise is applied to relative-depth maps normalized to $[0,1]$.}
\label{tb:depth_robustness}
\end{table}

\begin{figure}[t!]
\centering
\includegraphics[width=0.465\textwidth]{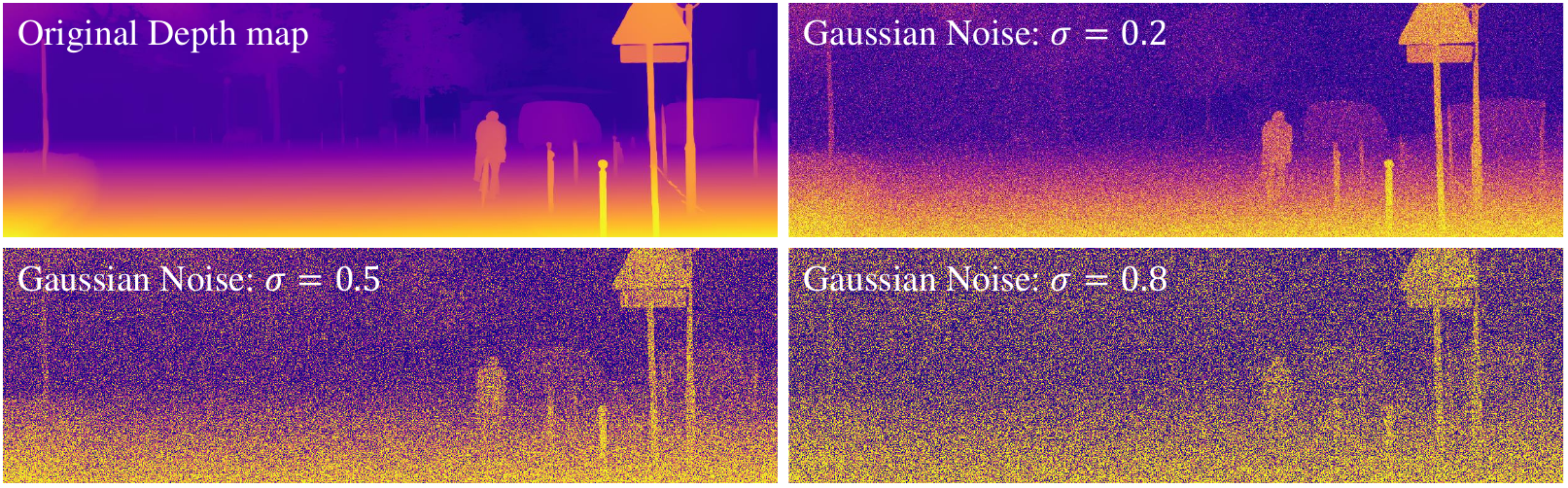}
\centering
\caption{Visualizations of noisy depth maps.}
\label{fig.noisy}
\end{figure} 

\begin{figure}[t!]
\centering
\includegraphics[width=0.485\textwidth]{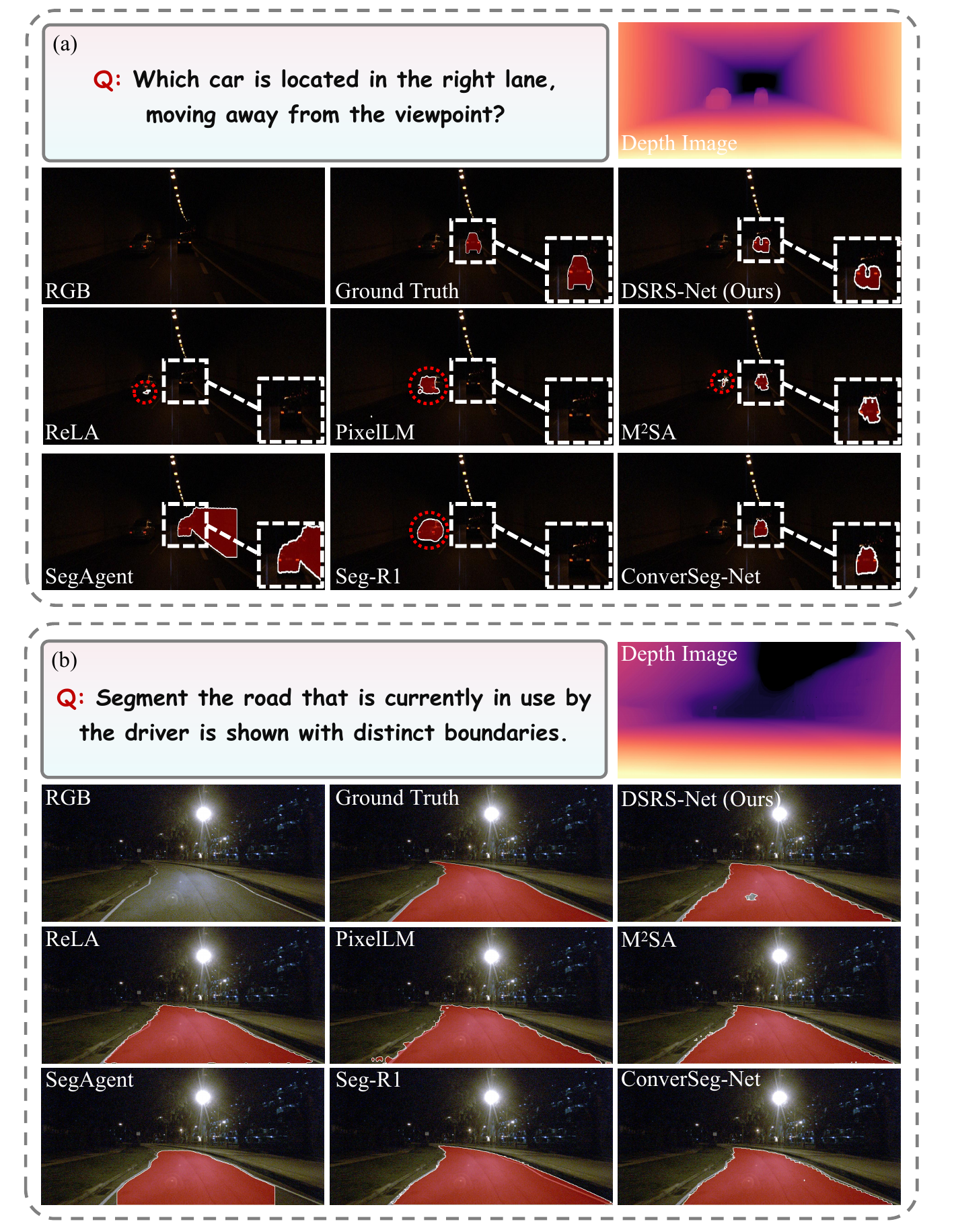}
\caption{Qualitative results of depth quality analysis under low illumination.}
\label{fig.baddepth1}
\end{figure}

\begin{figure}[t!]
\centering
\includegraphics[width=0.485\textwidth]{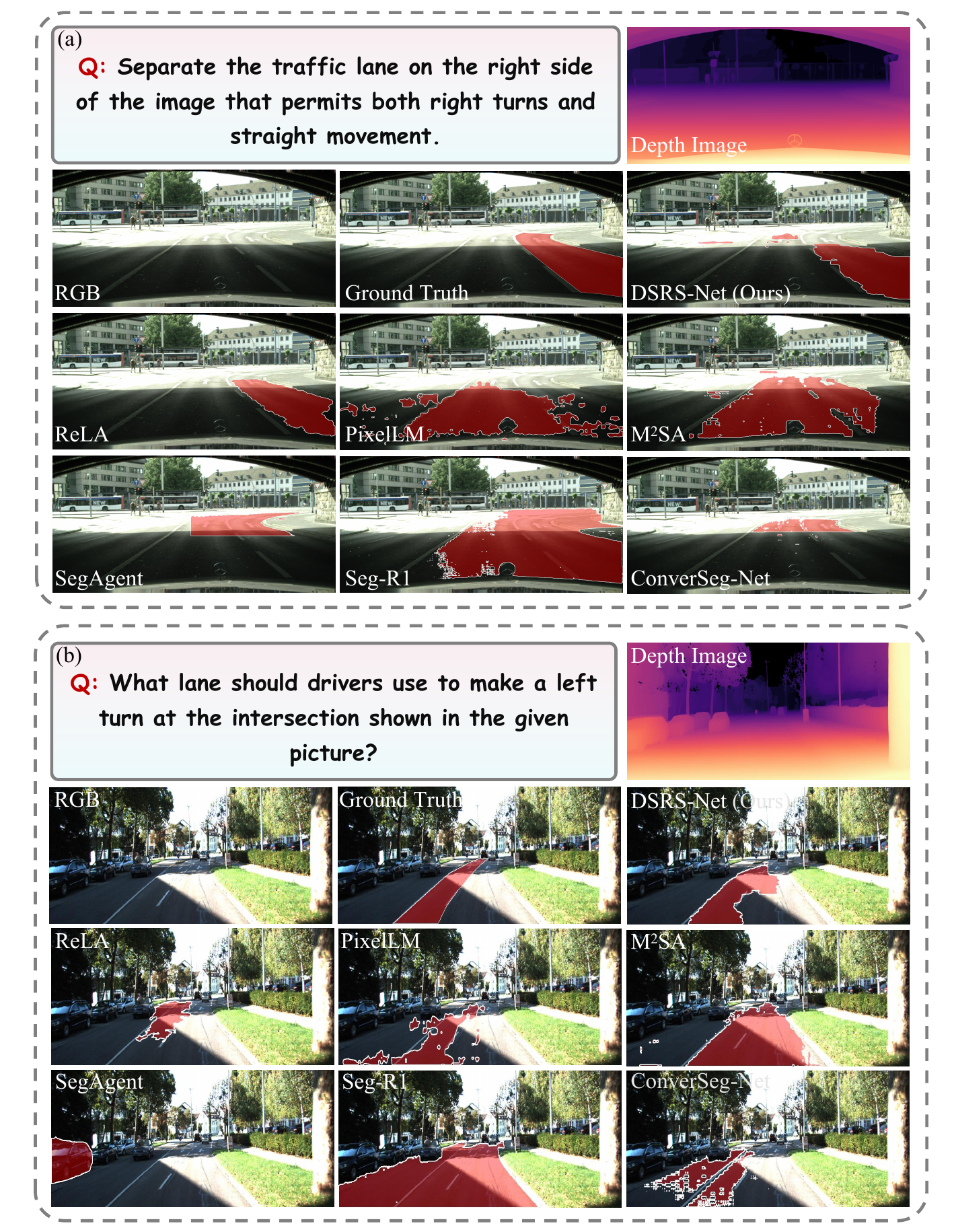}
\caption{Qualitative results of depth quality analysis under strong illumination.}
\label{fig.baddepth2}
\vspace{-0.5em}
\end{figure}

\begin{table}[t]
\centering
\setlength{\tabcolsep}{1.3mm}
\settablefont
\resizebox{\columnwidth}{!}{
\begin{tabular}{l|cc|cc|cc|cc}
\toprule[1pt]
\multirow{3}{*}{Method}
& \multicolumn{4}{c|}{Zero-shot (zs)}
& \multicolumn{4}{c}{Fine-tuned (ft)} \\
\cline{2-9}
& \multicolumn{2}{c|}{Val} & \multicolumn{2}{c|}{Test}
& \multicolumn{2}{c|}{Val} & \multicolumn{2}{c}{Test} \\
& gIoU & cIoU & gIoU & cIoU & gIoU & cIoU & gIoU & cIoU \\
\midrule
LISA-LLaVAv1 \cite{lai2024lisa}
& 22.7 & 20.3 & 30.5 & 28.3 & 55.7 & 52.5 & 56.8 & 55.1 \\
GSVA-LLaVAv1 \cite{xia2024gsva}
& 26.1 & 19.4 & 31.1 & 26.3 & 59.7 & 60.8 & 59.9 & 61.0 \\
\rowcolor[HTML]{DCDCDC}
\textbf{\ournet-LLaVAv1}
& \textbf{27.3} & \textbf{22.7} & \textbf{34.2} & \textbf{29.1}
& \textbf{61.6} & \textbf{64.5} & \textbf{61.0} & \textbf{65.0} \\
\bottomrule[1pt]
\end{tabular}
}
\caption{Same-backbone comparison on the \ourmethod dataset. \ournet-LLaVAv1 adopts the LISA backbone augmented with our GPE and DHFI components.}
\label{tb:same_backbone}
\end{table}

\begin{figure}[t!]
\centering
\includegraphics[width=0.485\textwidth]{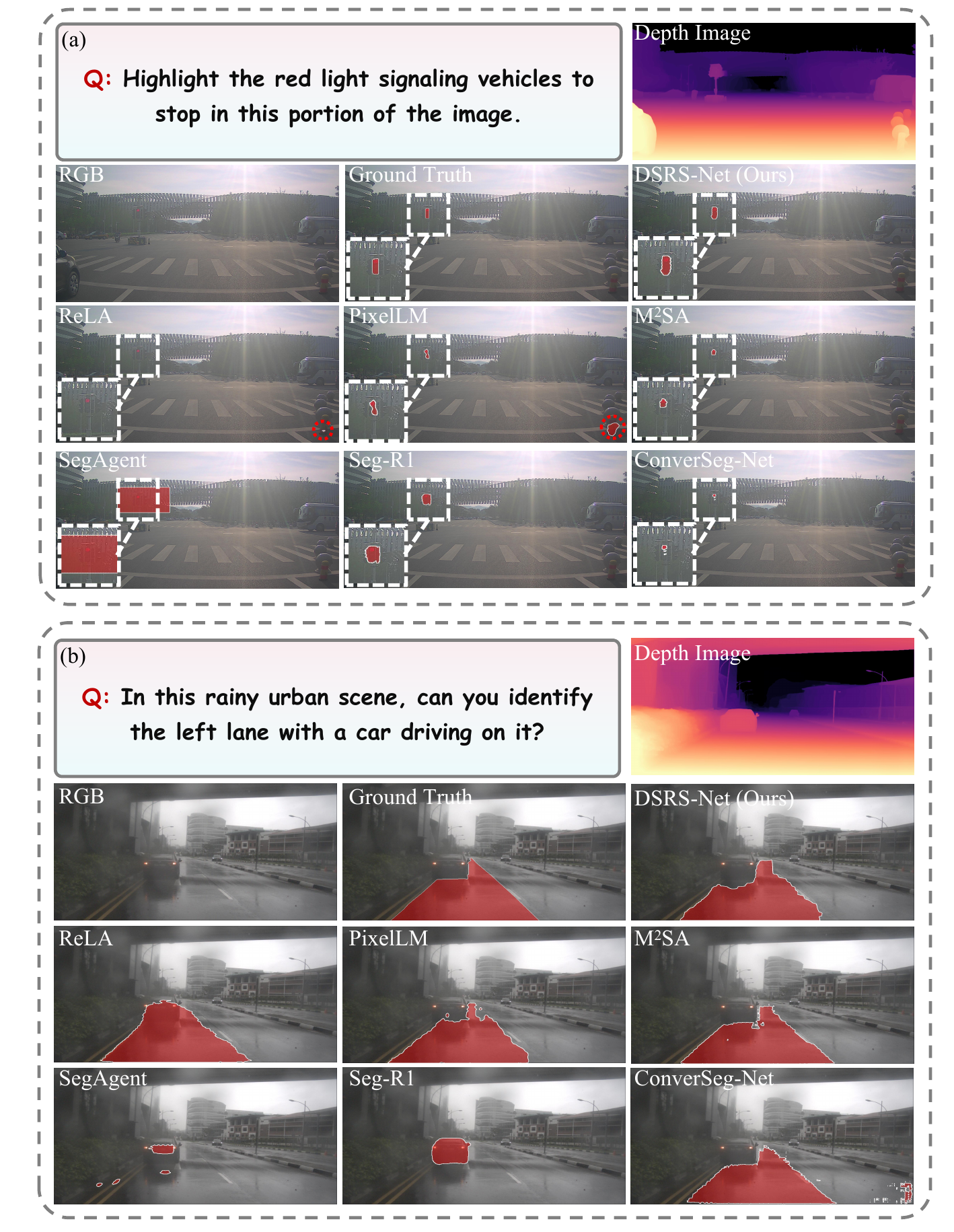}
\caption{Qualitative results of depth quality analysis under visual blur.}
\label{fig.baddepth3}
\vspace{-0.5em}
\end{figure}

\begin{table}[t]
\centering
\setlength{\tabcolsep}{2.7mm}
\settablefont
\begin{tabular}{lccc}
\toprule[1pt]
Metric & \makecell{\ournet-\\LLaVA1.5} & \makecell{\ournet-\\InternVL2.5} & \makecell{Depth\\Generation} \\
\midrule
Input resolution &$336 \times 336$   & $896 \times 896$ & $518 \times 518$\\
Total parameters (B) &7.12   & 8.14 & 0.34\\
Trainable parameters (M) &318.46   & 82.43 & N/A \\
FLOPs (T/image) &17.31   &36.06 & 1.30\\
Latency (s/image) &2.07   & 3.81 & 0.12 \\
Throughput (images/s) &0.48   & 0.26 & 8.65 \\
Peak GPU memory (GB) &28.43   & 34.68 & 1.75 \\
Test GPU & A800 & A800 & RTX 3090 \\
Execution mode & Online & Online & Offline \\
\bottomrule[1pt]
\end{tabular}
\caption{Computational efficiency of \ournet and the offline depth generation pipeline. Both \ournet variants are evaluated on an A800 GPU with batch size 1. Relative depth maps are generated in advance with Depth Anything V2-Large on an RTX 3090 GPU.}
\label{tb:computational_efficiency}
\end{table}

\begin{figure}[t!]
\centering
\includegraphics[width=0.485\textwidth]{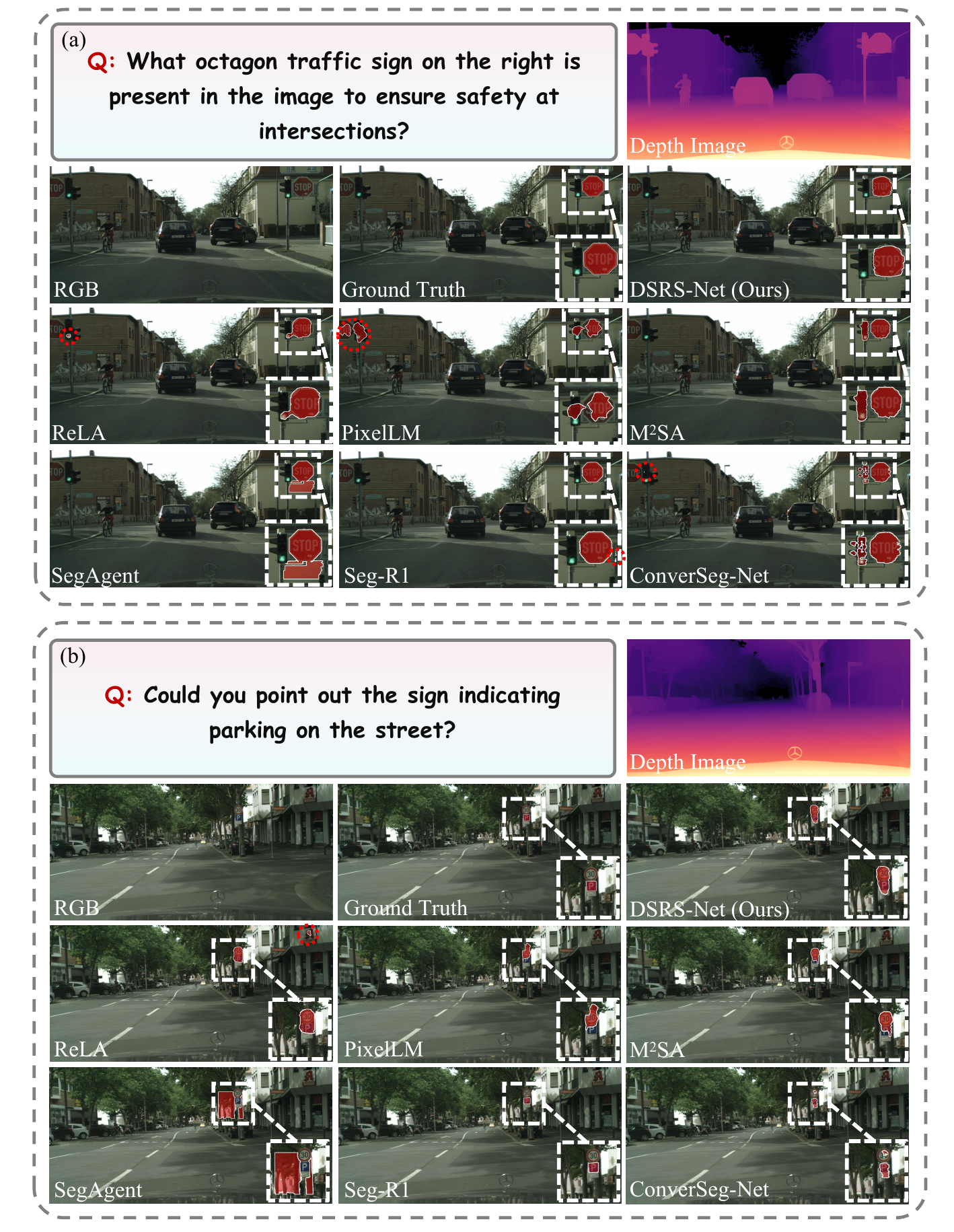}
\caption{ {Qualitative results of depth quality analysis under background clutter.}}
\label{fig.baddepth4}
\end{figure}

\begin{figure}[t!]
\centering
\includegraphics[width=0.485\textwidth]{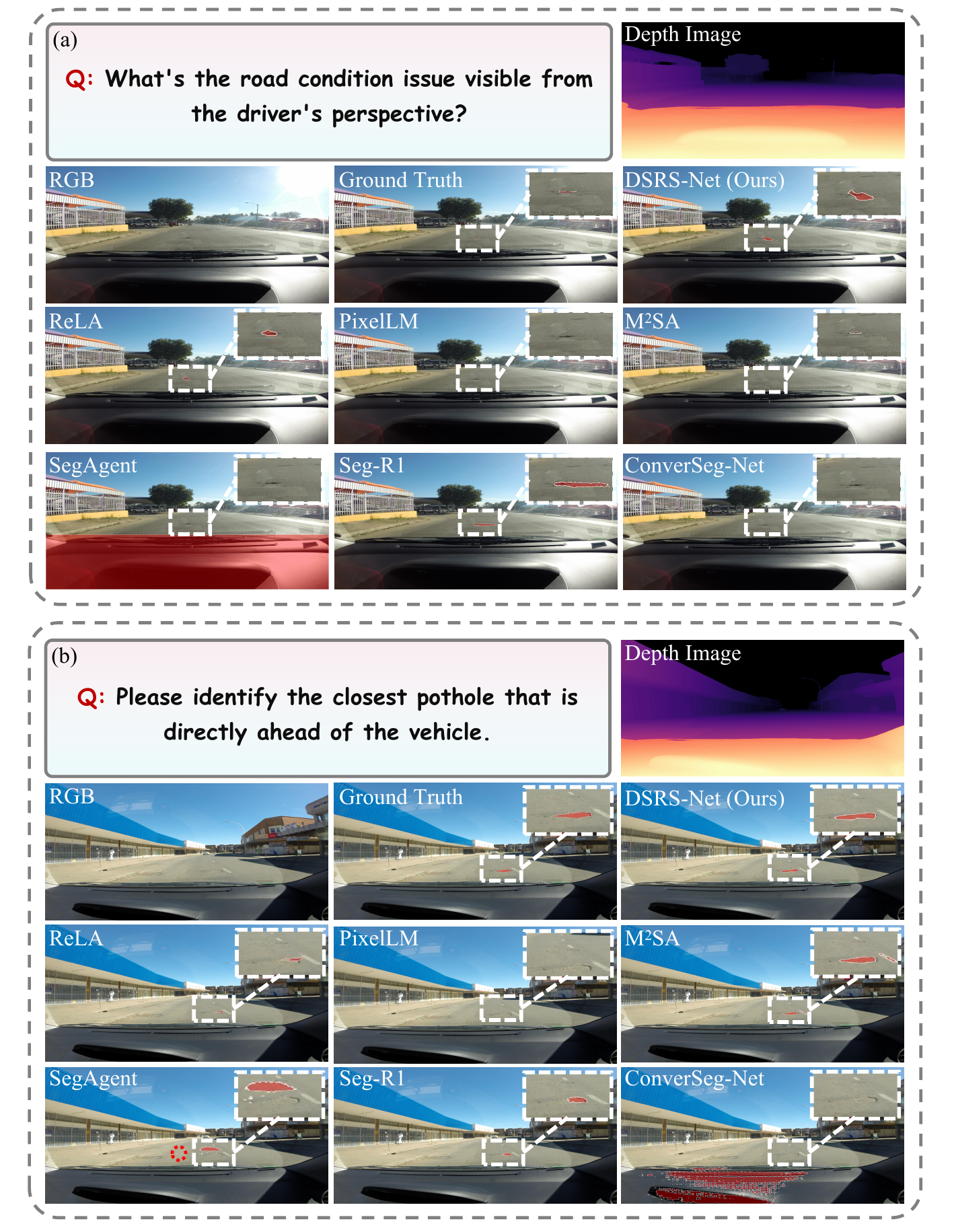}
\caption{ {Qualitative results of depth quality analysis for targets with small spatial extent.}}
\label{fig.baddepth5}
\end{figure}

We further evaluate \ournet using relative-depth maps generated by the Small, Base, and Large variants of Depth Anything V2 and perturb the Large-model depth maps with zero-mean Gaussian noise at different intensities. As shown in Fig. \ref{fig.noisy} and Table \ref{tb:depth_robustness}, all three depth variants yield positive results over RGB-only inference, indicating that the benefit is not restricted to the highest-capacity depth estimator. The model also remains robust under small perturbations because GPE and DHFI can tolerate moderate local depth errors. Under severe corruption, however, performance falls below the RGB-only baseline. Heavily distorted geometric relationships conflict with the appearance features and disrupt the original balance learned by the LLM and interaction modules. This analysis shows that \ournet is robust to variations in depth-estimation quality and moderate noise, but is not invariant to severely corrupted geometric inputs.

To examine the performance of existing models on the \ourmethod task when estimated depth is degraded, we qualitatively evaluate these models under five representative challenging conditions: low illumination, strong illumination, visual blur, background clutter, and targets with small spatial extent. The qualitative results under these conditions correspond to Figs. \ref{fig.baddepth1}--\ref{fig.baddepth5}, respectively.

Figs. \ref{fig.baddepth1}--\ref{fig.baddepth3} show that extreme illumination and adverse weather simultaneously degrade RGB visibility and monocular depth quality. In dark tunnels and nighttime scenes, low appearance contrast and imprecise depth boundaries around roads and pedestrians often result in localization errors and incomplete segmentation masks. Conversely, overexposed road regions distort local depth and interfere with lane localization. Fog and rain further obscure small structures and reduce depth contrast among vehicles, roads, and roadside objects, resulting in imprecise contours.

Background clutter and small target scale introduce similar difficulties. As shown in Fig. \ref{fig.baddepth4}, nearby windows, walls, and vegetation blur the depth boundaries of traffic signs. In Fig. \ref{fig.baddepth5}, potholes occupy only a few pixels and exhibit weak depth differences from the surrounding road, increasing both missed detections and false positives. These cases indicate that unreliable depth is harmful when the RGB appearance is already ambiguous or the target geometry is too fine-grained to be preserved.

Importantly, these failures cannot be attributed solely to the geometric branch. The conditions that produce poor depth maps are themselves highly challenging for RGB-only reasoning segmentation. For example, in Fig. \ref{fig.baddepth1}(b), competing RGB-based models without explicit geometric priors also fail to identify the correct drivable road because illumination and lane-boundary cues are severely degraded. Thus, depth degradation and RGB ambiguity jointly limit performance, while corrupted geometry may further disturb the learned balance between appearance and geometric features. In future work, we will introduce depth-confidence estimation and uncertainty-aware adaptive fusion, allowing the model to suppress unreliable geometric cues and fall back to RGB features when necessary. We will also incorporate depth-corruption augmentation during training to improve robustness under low light, adverse weather, clutter, and small-target scenarios.

\subsection{Fair Comparison with Depth-Enhanced Baselines}
\label{sec:depth_enhanced}

\begin{table*}[t]
\centering
\setlength{\tabcolsep}{4.4mm}
\settablefont
\begin{tabular}{ll|cc|cc|cc|cc}
\toprule[1pt]
\multirow{3}{*}{Method} & \multirow{3}{*}{MLLM} & \multicolumn{2}{c|}{Val} & \multicolumn{6}{c}{Test} \\
\cline{3-10}
& & \multicolumn{2}{c|}{Overall} & \multicolumn{2}{c|}{Short Query} & \multicolumn{2}{c|}{Long Query} & \multicolumn{2}{c}{Overall} \\
& & gIoU & cIoU & gIoU & cIoU & gIoU & cIoU & gIoU & cIoU \\
\hline
\hline
OVSeg \cite{liang2023open} & - & 28.5 & 18.6 & 18.0 & 15.5 & 28.7 & 22.5 & 26.1 & 20.8 \\
ReLA \cite{liu2023gres} & - & 22.4 & 19.9 & 17.6 & 15.0 & 22.6 & 23.8 & 21.3 & 22.0 \\  
X-Decoder \cite{zou2023generalized} & - & 22.6 & 17.9 & 20.4 & 11.6 & 22.2 & 17.5 & 21.7 & 16.3 \\
SEEM \cite{zou2023segment}& - & 25.5 & 21.2 & 20.1 & 11.5 & 25.6 & 20.8 & 24.3 & 18.7 \\
Grounded-SAM \cite{liu2024grounding} & - & 26.0 & 14.5 & 17.8 & 10.8 & 22.4 & 18.6 & 21.3 & 16.4 \\
LLaVA-7B + OVSeg & LLaVA-NeXT-7B & 38.2 & 23.5 & 24.2 & 18.7 & 44.6 & 37.1 & 39.7 & 31.8 \\
LLaVA-7B + CoT + OVSeg & LLaVA-NeXT-7B & 42.1 & 26.6 & 27.5 & 19.6 & 44.6 & 40.2 & 40.4 & 34.0 \\
LISA-7B \cite{lai2024lisa} & LLaVA-v1 7B & 44.4 & 46.0 & 37.6 & 34.4 & 36.6 & 34.7 & 36.8 & 34.1 \\
LISA-7B (ft) \cite{lai2024lisa} & LLaVA-v1 7B & 52.9 & 54.0 & 40.6 & 40.6 & 49.4 & 51.0 & 47.3 & 48.4 \\
LISA-7B \cite{lai2024lisa} & LLaVA-1.5 7B & 53.6 & 52.3 & 47.1 & 48.5 & 49.2 & 48.9 & 48.7 & 48.8 \\
LISA-7B (ft) \cite{lai2024lisa} & LLaVA-1.5 7B & 61.3 & 62.9 & 48.3 & 46.3 & 57.9 & 59.7 & 55.6 & 56.9 \\
LISA++-7B (ft) \cite{yang2023lisa++} & LLaVA-1.5 7B & 64.2 & 68.1 & 49.6 & 51.1 & 59.3 & 61.7 & 57.0 & 59.5 \\
PixelLM \cite{ren2024pixellm} & LLaVA-v1 7B & 44.4 & 32.1 & - & - & - & - & 41.1 & 38.1 \\
M$^2$SA-7B \cite{jang2025mmr} & LLaVA-v1 7B & 30.5 & 35.2 & - & - & - & - & 24.7 & 28.0 \\
Seg-R1 \cite{you2025seg} & Qwen2.5-VL 7B & 58.6 & 41.2 & - & - & - & - & 56.7 & 53.7 \\
SegAgent \cite{zhu2025segagent} & Qwen-VL 7B & 34.4 & 28.7 & - & - & - & - & 34.5 & 31.2 \\
RSVP-LLaVA \cite{lu2025rsvp} & LLaVA-NeXT-7B & 59.2 & 56.7 & 47.9 & 42.0 & 58.4 & 53.0 & 55.9 & 50.7 \\
RSVP-Qwen \cite{lu2025rsvp} & Qwen2-VL 7B & 58.6 & 48.5 & 48.5 & 44.3 & 57.1 & 53.8 & 56.6 & 51.6 \\
RSVP-Gemini \cite{lu2025rsvp} & Gemini-Flash & 56.9 & 49.2 & 47.3 & 40.2 & 60.2 & 65.6 & 57.1 & 59.2 \\
RSVP-GPT \cite{lu2025rsvp} & GPT-4o & 64.7 & 63.1 & 55.4 & 50.4 & 61.9 & 62.5 & 60.3 & 60.0 \\
ConverSeg-Net \cite{sahoo2026conversational} & Qwen2.5-VL 3B & 56.4 & - & 53.8 & - & 51.7 & - & 52.2 & - \\
\midrule
\rowcolor[HTML]{DCDCDC}
\textbf{\ournet}& LLaVA1.5 7B &53.9 &57.7 &40.6  &34.8  &57.0  &63.8 &54.5 &56.8     \\
\rowcolor[HTML]{DCDCDC}
\textbf{\ournet (ft)} & LLaVA1.5 7B &54.1  &60.5  &43.2  &40.2  &58.7 & 64.0 &55.0 &58.1   \\
\rowcolor[HTML]{DCDCDC}
\textbf{\ournet }&InternVL2.5 8B  &57.0  &57.2  &51.6  &46.1  &63.4 &68.8  &60.5  &63.3  \\
\rowcolor[HTML]{DCDCDC}
\textbf{\ournet (ft)} &InternVL2.5 8B &62.5  &66.8  &54.6  &51.1  &68.6 &74.9  &65.2 &69.4 \\
\bottomrule[1pt]
\end{tabular}
\caption{Supplementary comparison on the ReasonSeg benchmark under different query lengths.}
\label{tb.reasonseg_query}
\end{table*}

As shown in Table \ref{tb:same_backbone}, to isolate the contribution of our architectural innovations from that of the MLLM backbone, we integrate GPE and DHFI into LISA while retaining the same LLaVA-v1 7B backbone. The controlled gains are smaller than the headline improvements in the main comparison because the backbone advantage is removed. Even under this stricter setting, \ournet raises test gIoU from $30.5$ to $34.2$ in zero-shot evaluation and from $56.8$ to $61.0$ after fine-tuning. These positive results show that depth-aware geometric modeling and hierarchical feature interaction remain effective independently of backbone strength. The fine-tuned test cIoU also rises from $55.1$ to $65.0$, providing further evidence that our method improves mask quality under a matched backbone.

\begin{table*}[t]
  \centering
  \settablefont

  {

  \setlength{\tabcolsep}{3.0mm}
  \begin{tabular}{l|c|ccc|ccc|cccccc|c}
    \toprule[1pt]
    \multirow{2}{*}{\textbf{Methods}} &
    \multirow{2}{*}{
      \raisebox{0pt}[4.5em][4.5em]{%
      \rotatebox[origin=c]{90}{\textbf{Spatial Relations}}
      }
      } &
    \multicolumn{3}{c|}{\textbf{Potential Risks}} &
    \multicolumn{3}{c|}{\textbf{Challenging Conditions}} &
    \multicolumn{6}{c|}{\textbf{Traffic Rules}} &
    
    \multirow{2}{*}{
      \raisebox{0pt}[4.5em][4.5em]{%
      \rotatebox[origin=c]{90}{\textbf{Overall}}}} \\
    \cline{3-14}
    & &
    \rule{0pt}{5em}\rotatebox[origin=c]{90}{Obstacle} &
    \rotatebox[origin=c]{90}{Road Defect} &
    \rotatebox[origin=c]{90}{Traffic Accident} &
    \rotatebox[origin=c]{90}{Complex Terrain} &
    \rotatebox[origin=c]{90}{Light Condition} &
    \rotatebox[origin=c]{90}{Weather Condition} &
    \rotatebox[origin=c]{90}{Bicycle Lane} &
    \rotatebox[origin=c]{90}{Bus Lane} &
    \rotatebox[origin=c]{90}{Crosswalk} &
    \rotatebox[origin=c]{90}{Traffic Lane} &
    \rotatebox[origin=c]{90}{Traffic Light} &
    \rotatebox[origin=c]{90}{Traffic Sign} &
    \\
    \hline
    \hline
    PixelLM \cite{ren2024pixellm}
    & 53.0                           
    & 27.9 & 10.4 & 65.4            
    & 55.5 & 53.0 & 35.2            
    & 55.3 & 50.3 & 47.4 & 64.0 & 19.2 & 25.4  
    & 40.1                          
    \\

    M$^2$SA \cite{jang2025mmr}
    & 82.5                           
    & 46.7 & 52.1 & 81.0            
    & 62.1 & 70.7 & 64.1            
    & 59.6 & 42.9 & 53.6 & 66.5 & 43.9 & 42.8  
    & 57.5                            
    \\

    SegAgent \cite{zhu2025segagent}
    & 80.5                           
    & 17.4 & 0.4 & 80.9             
    & 47.1 & 58.7 & 33.7            
    & 22.8 & 6.4 & 20.4 & 32.9 & 12.6 & 16.7  
    & 30.3                          
    \\


    ConverSeg-Net \cite{sahoo2026conversational}
    & 83.1                           
    & 45.2 & 2.0 & 86.5             
    & 74.0 & 66.1 & 62.7            
    & 25.0 & 21.2 & 32.1 & 49.9 & 48.4 & 46.3  
    & 47.9                             
    \\

    \midrule
    
    \rowcolor[HTML]{DCDCDC}
    \ournet~(Ours)
    & 85.0                             
    & 58.8 & 41.5 & 86.5                  
    & 77.8 & 77.9 & 66.5                  
    & 80.9 & 70.4 & 72.6 & 74.5 & 53.6 & 57.3  
    & \textbf{71.2}                   
    \\

    \bottomrule[1pt]
  \end{tabular}%
  }

  \caption{Per-category results on the \ourmethod dataset.}
  \label{tab:per-class}
\end{table*}

\subsection{Computational Efficiency and Deployment Analysis}
\label{sec:efficiency}

We evaluate the computational efficiency of \ournet with both LLaVA1.5-7B and InternVL2.5-8B backbones. Both model variants are tested on an A800 GPU with batch size one. The relative depth maps are generated and tested in advance using Depth Anything V2-Large on an RTX 3090 GPU. We report the total and trainable parameters, floating-point operations (FLOPs), end-to-end inference latency, throughput, and peak GPU memory. Latency is averaged over 100 inference runs after 10 warm-up runs, and throughput is measured without data-loading time. Since the relative-depth maps are generated offline by Depth Anything V2, depth estimation is excluded from the online latency and memory measurements. The reported overhead therefore reflects the deployable reasoning-segmentation network itself.

As shown in Table \ref{tb:computational_efficiency}, the LLaVA1.5 configuration provides a lighter deployment option, whereas the InternVL2.5 configuration targets stronger reasoning performance with a larger computational budget. The comparison enables practitioners to select a suitable configuration according to available memory and latency constraints. Moreover, offline depth generation avoids introducing an additional depth-estimation network into the online inference path, making \ournet applicable to systems in which relative-depth maps can be precomputed or supplied by an independent perception module.

\subsection{Additional Comparison on the ReasonSeg Benchmark}
\label{sec:additional_reasonseg}

\begin{table*}[!t]
\centering
\settablefont
\setlength{\tabcolsep}{10.0pt}
\begin{tabular}{cccc|cccc|cccc}
\toprule[1pt]
\multirow{2}{*}{\makecell{RGB\\Branch}}
& \multirow{2}{*}{\makecell{Depth CLIP\\Branch}}
& \multirow{2}{*}{GPE}
& \multirow{2}{*}{DHFI}
& \multicolumn{4}{c|}{gIoU}
& \multicolumn{4}{c}{cIoU} \\
\cline{5-12}
& & & & Run 1 & Run 2 & Run 3
& Mean$_{\color{violet}\pm\mathrm{Std}}$
& Run 1 & Run 2 & Run 3
& Mean$_{\color{violet}\pm\mathrm{Std}}$ \\
\midrule

$\checkmark$ & & & &
52.82 & 53.18 & 53.00 &
$53.00_{\color{violet}\pm0.18}$ &
50.66 & 50.18 & 50.42 &
$50.42_{\color{violet}\pm0.24}$ \\

$\checkmark$ & $\checkmark$ & & &
55.04 & 55.46 & 55.25 &
$55.25_{\color{violet}\pm0.21}$ &
52.12 & 51.95 & 52.29 &
$52.12_{\color{violet}\pm0.17}$ \\

$\checkmark$ & & $\checkmark$ & &
37.66 & 37.12 & 37.39 &
$37.39_{\color{violet}\pm0.27}$ &
36.48 & 36.67 & 36.86 &
$36.67_{\color{violet}\pm0.19}$ \\

$\checkmark$ & & & $\checkmark$ &
53.42 & 53.28 & 53.14 &
$53.28_{\color{violet}\pm0.14}$ &
54.88 & 55.32 & 55.10 &
$55.10_{\color{violet}\pm0.22}$ \\

\rowcolor[HTML]{EFEFEF}
$\checkmark$ & & $\checkmark$ & $\checkmark$ &
55.24 & 55.40 & 55.56 &
$\textbf{55.40}_{\color{violet}\pm0.16}$ &
57.10 & 56.84 & 56.97 &
$\textbf{56.97}_{\color{violet}\pm0.13}$ \\

\bottomrule[1pt]
\end{tabular}
\caption{Component-wise ablation results over three independent runs
on the \ourmethod dataset. \textcolor{violet}{$\pm$Std} denotes the
sample standard deviation across the three runs.}
\label{tb.sup_abla1}
\end{table*}

\begin{table}[!t]
\begin{center}
\setlength{\tabcolsep}{3.4mm}
{
\settablefont
{
\begin{tabular}{ll|l|cc}
\toprule[1pt]
\makecell[l]{Contextual \\Features} 
& \makecell[l]{Geometric \\Features} 
& Fusion Strategies 
& gIoU 
& cIoU  \\
\hline
\hline
 &$\checkmark$&Self-Attention 
&44.56
&44.08  \\
$\checkmark$&$\checkmark$&\makecell[l]{Element-wise \\Summation} 
&50.47
&51.12\\
$\checkmark$&$\checkmark$&\makecell[l]{Channel-wise \\Concatenation} 
&\thirdcell{51.09} 
&\thirdcell{51.86} \\
$\checkmark$&$\checkmark$&Cross-Attention 
&\secondcell{52.78}
&\secondcell{54.05}\\
$\checkmark$&$\checkmark$&${DA}^2$ 
&\bestcell{55.40}
&\bestcell{56.97}\\
\bottomrule[1pt]
\end{tabular}
}
\caption{Ablation studies on the DHFI strategy. }
\label{tb.sup_abla2}
}
\vspace{-0.5em}
\end{center}
\end{table}

Table \ref{tb.reasonseg_query} further analyzes performance with respect to query length. For short queries, \ournet-InternVL2.5 (ft) achieves $54.6$ gIoU and $51.1$ cIoU, improving the gIoU of LISA++ by $5.0$ points while matching its cIoU, and remaining competitive with the strongest RSVP-GPT result. The advantage becomes substantially larger for long queries. \ournet-InternVL2.5 (ft) reaches $68.6$ gIoU and $74.9$ cIoU, outperforming LISA++ by $9.3$ and $13.2$ points and RSVP-GPT by $6.7$ and $12.4$ points, respectively. This pattern indicates that \ournet is particularly effective for long, compositional instructions that require multi-step semantic reasoning, spatial-relation understanding, and precise grounding of the inferred target. We attribute this strength to the explicit interaction between contextual appearance features and geometric priors, which helps preserve target-relevant spatial evidence throughout complex reasoning. Consequently, \ournet obtains the best overall test performance of $65.2$ gIoU and $69.4$ cIoU, demonstrating its advantage in resolving complex queries without sacrificing performance on concise instructions.

\subsection{Per-Category Results on the \ourmethod Dataset}
\label{sec:per_category_results}

Table \ref{tab:per-class} shows that \ournet achieves the best overall gIoU of $71.2$, demonstrating broad gains across different driving-scene reasoning categories. The model is particularly effective at spatial relations, traffic accidents, complex terrain, and adverse illumination, where correct predictions require jointly reasoning about scene context and geometric layout. It also performs strongly on traffic-rule categories involving spatially structured regions, including bicycle lanes, bus lanes, crosswalks, and traffic lanes, indicating that geometric priors help associate implicit functional descriptions with the correct road areas. Nevertheless, substantial room for improvement remains for road defects, traffic lights, traffic signs, and obstacles. Road defects often exhibit weak depth and appearance contrast with the surrounding road, while traffic lights and signs occupy very small regions and require fine-grained recognition of state, shape, and semantics. Weather conditions also remain challenging because rain and fog simultaneously degrade visual details and depth quality. Future work should therefore strengthen small-object representation, boundary refinement, uncertainty-aware geometric fusion, and category-specific reasoning for ambiguous or visually degraded targets.

\subsection{Additional Ablation Studies}
\label{sec:additional_ablation}

\begin{figure*}[t!]
\centering
\includegraphics[height=0.95\textheight,keepaspectratio]{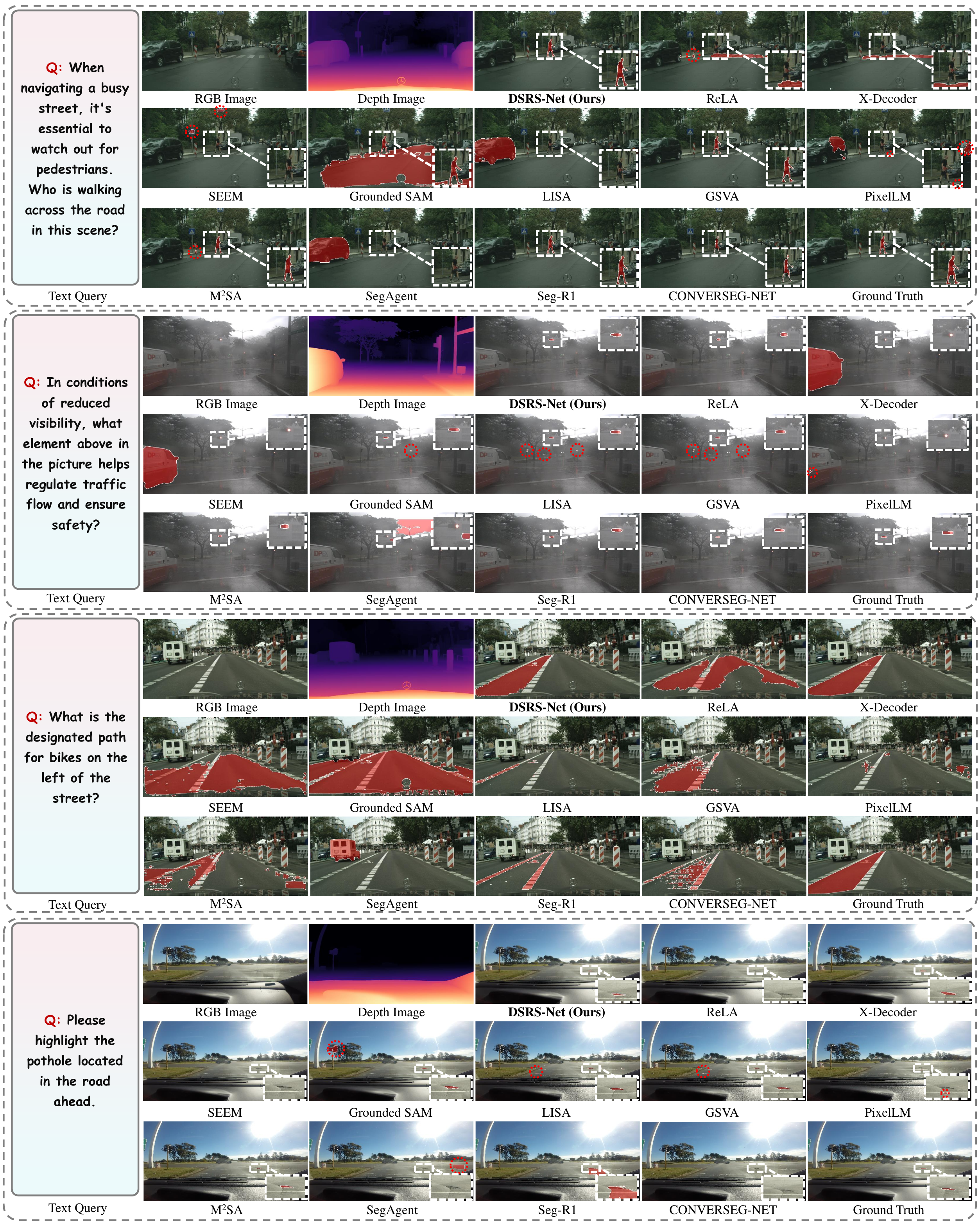}
\caption{Qualitative comparisons between \ournet and existing language-driven segmentation networks on the \ourmethod dataset, where ground-truth areas are highlighted with white dashed boxes and wrong areas are highlighted with red dashed circles.}
\label{fig.sup_pred1}
\end{figure*}

\begin{figure*}[t!]
\centering
\includegraphics[height=0.95\textheight,keepaspectratio]{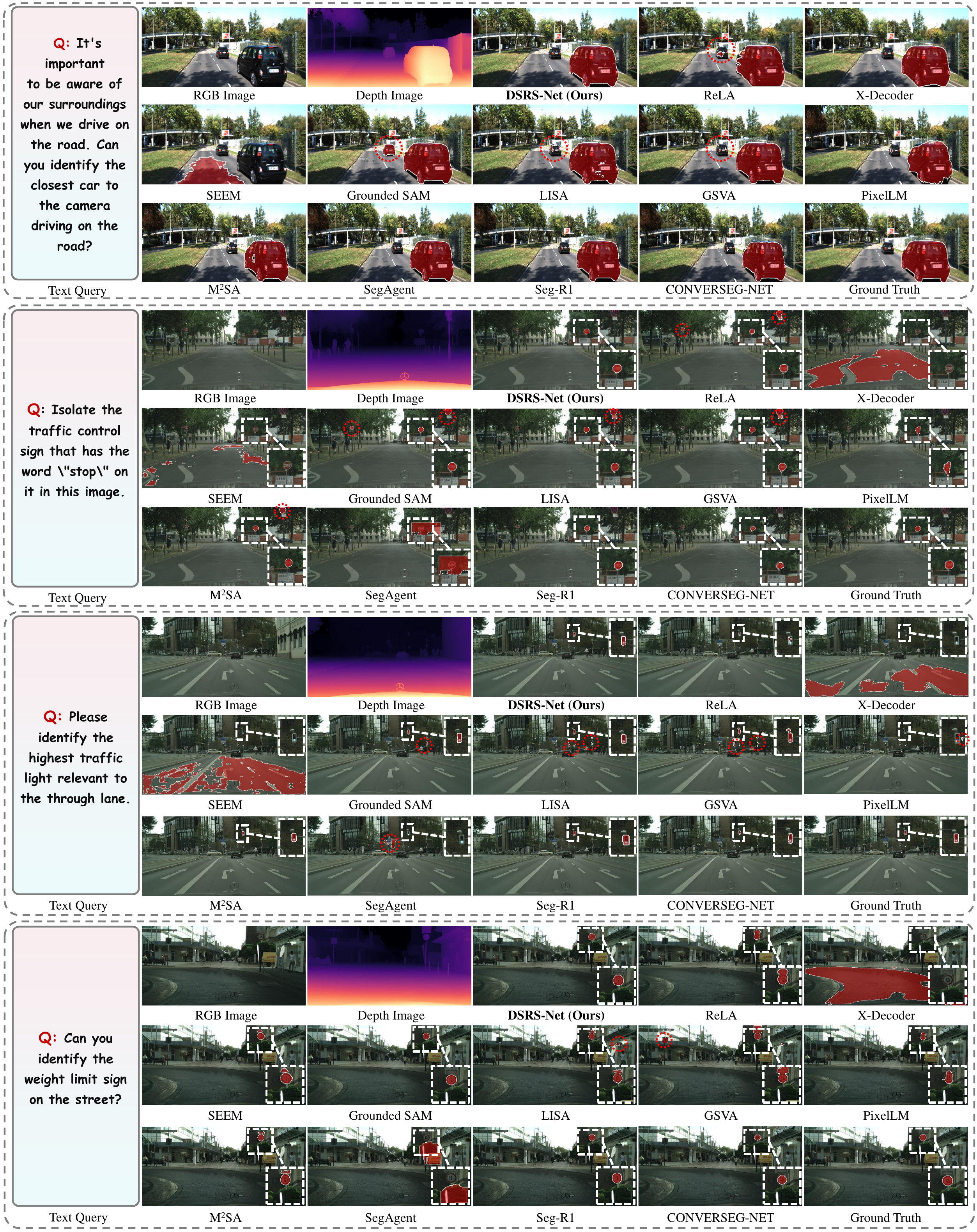}
\caption{Qualitative comparisons between \ournet and existing language-driven segmentation networks on the \ourmethod dataset.}
\label{fig.sup_pred2}
\end{figure*}

As a further analysis of the ablation results reported in the main text, we examine how geometric priors and feature-interaction strategies contribute to \ournet.
As shown in Table \ref{tb.sup_abla1}, we first feed the depth map into the CLIP-based vision encoder alongside the RGB image, resulting in a limited increase of 1.70\% in cIoU compared to that of the RGB-only baseline. We attribute this phenomenon to the RGB-only pretraining mechanism of the MLLM, which renders it incapable of leveraging prior knowledge to effectively process geometric features. As discussed earlier, scene inference requires a specific mechanism to align geometric features with their underlying contextual representations. By incorporating the proposed GPE and DHFI modules, we observe a significant positive performance improvement in driving scene reasoning segmentation. Notably, gIoU and cIoU increase by 1.4\% and 6.6\%, respectively. These findings lend further support to our hypothesis stated earlier. 

As observed, the improvement cannot be achieved without the contribution of the GPE module, which leads to an increase of approximately 2.1\% in gIoU. However, when the GPE module is applied without the support of an interaction mechanism, performance significantly degrades to 36.7\% in cIoU. This performance degradation likely arises from the representational incompatibility between the geometric features and the contextual features, thereby causing integration conflicts within the LLM. The effectiveness of the convolutional GPE module is significantly enhanced only when it follows the DHFI module. These experimental results underscore that the successful utilization of geometric features depends critically on both their effective extraction and proper alignment with contextual information. 

As shown in Table \ref{tb.sup_abla2}, to investigate the most effective structure for the DHFI strategy, we substitute the basic ${DA}^2$ blocks with alternatives. We first construct baseline architectures utilizing self-attention blocks exclusively on geometric features. As expected, such a baseline architecture fails to achieve performance comparable to other fusion strategies. We then investigate the effectiveness of aligning geometric features with contextual vision features. It is evident that simple fusion strategies such as element-wise summation and channel-wise concatenation yield limited performance improvements, while attention-based strategies significantly improve the framework's reasoning performance. Specifically, our proposed ${DA}^2$ module achieves an increase of 2.6\% in gIoU and 2.9\% in cIoU compared to the deformable cross-attention strategy. These results demonstrate the efficacy of depth-aware offsets in guiding the model to focus more on contextually and geometrically relevant regions during the feature interaction process.

\section{Additional Qualitative Experimental Results}
\label{sec:qual_results}

We compare the proposed \ournet with existing language-driven segmentation networks on the \ourmethod dataset. Qualitative results are provided in Figs. \ref{fig.sup_pred1} and \ref{fig.sup_pred2} of the supplementary material. The evaluated scenarios are visually complex and involve multiple aspects, including spatial relations, potential risks, challenging conditions, and traffic rules. Moreover, the referred segmentation targets are highly diverse, encompassing a wide range of objects commonly encountered in driving scenes.

As illustrated in Figs. \ref{fig.sup_pred1} and \ref{fig.sup_pred2}, it is evident that our proposed \ourmethod significantly outperforms existing approaches in driving scene reasoning segmentation. When the text query refers to the function of an object rather than its explicit appearance, most competing approaches struggle to associate it with the correct visual target. Furthermore, the performance gap becomes even more pronounced under challenging environmental conditions. For instance, as shown in the second example of Fig. \ref{fig.sup_pred1} in the supplementary material, although scene visibility is reduced due to rain, \ourmethod accurately localizes the traffic light and yields accurate segmentation results. In contrast, other methods misidentify small twinkling elements as traffic lights and struggle to extract accurate object boundaries.
Moreover, in scenarios involving multiple visually similar candidates that necessitate deep reasoning, as illustrated in the third example of Fig. \ref{fig.sup_pred2} in the supplementary material, our method demonstrates superior inference capabilities and robustness.

\vfill

\end{document}